\documentclass[journal]{IEEEtran}
\ifCLASSINFOpdf
  \usepackage[pdftex]{graphicx}  
\else
\fi
\usepackage[colorlinks=true]{hyperref}
\usepackage{amssymb}
\usepackage{algorithm}
\usepackage{amsmath}

\usepackage{amsthm}
\usepackage{graphicx}
\usepackage{subfigure}
\usepackage{tabularx}
\usepackage{booktabs}
\usepackage{url}
\usepackage{bbm}
\usepackage[noend]{algpseudocode}
\usepackage{array}
\usepackage{balance}
\usepackage{multirow}
\usepackage{multicol}
\usepackage{threeparttable}
\usepackage{xcolor}
\usepackage{subfigure}
\allowdisplaybreaks[4]

\newcommand{\diag}{{\rm diag}}

\newcommand{\Ecal}{{\mathcal{E}}}

\newcommand{\Ncal}{{\mathcal{N}}}

\renewcommand{\a}{{\bf a}}
\renewcommand{\b}{{\bf b}}

\newcommand{\g}{{\bf g}}

\renewcommand{\r}{{\bf r}}

\renewcommand{\v}{{\bf v}}

\newcommand{\x}{{\bf x}}
\newcommand{\y}{{\bf y}}
\newcommand{\z}{{\bf z}}

\newcommand{\A}{{\bf A}}

\newcommand{\Dcal}{\mathcal{D}}

\newcommand{\G}{{\bf G}}
\newcommand{\Gcal}{{\mathcal{G}}}

\newcommand{\I}{{\bf I}}

\newcommand{\M}{{\bf M}}

\newcommand{\N}{{\bf N}}

\renewcommand{\P}{{\bf P}}
\newcommand{\Q}{{\bf Q}}

\newcommand{\U}{{\bf U}}

\newcommand{\W}{{\bf W}}

\newcommand{\X}{{\bf X}}

\newcommand{\Z}{{\bf Z}}

\newcommand{\bLambda}{\mathbf{\Lambda}}

\newcommand{\bSigma}{\boldsymbol{\Sigma}}

\newcommand{\bomega}{\boldsymbol{\omega}}
\newcommand{\bOmega}{\boldsymbol{\Omega}}

\newcommand{\1}{{\bf 1}}
\newcommand{\0}{{\bf 0}}

\newcommand{\argmin}{\operatornamewithlimits{\tiny argmin}}

\newcommand{\Prox}{\textbf{Prox}}

\newcommand{\lrincir}[1]{\left( #1 \right)}

\newcommand{\lrnorm}[1]{\left\lVert#1\right\rVert}
\newcommand{\lrangle}[1]{\left\langle#1 \right\rangle}

\newcommand{\RR}{\mathbb{R}}

\begin{document}

\title{Medical Federated Model with Mixture of Personalized and Sharing Components}

\author{Yawei Zhao, Qinghe Liu$^{\dagger}$, Xinwang Liu, Kunlun He$^{\ddagger}$
\thanks{Yawei Zhao, Qinghe Liu, and Kunlun He are with the Medical Big Data Research Center, Chinese PLA General Hospital, Beijing, 100835, China. Xinwang Liu is with the School of Computer, National University of Defense Technology, Changsha, Hunan, China. E-mail: \texttt{csyawei.zhao@gmail.com}, \texttt{liuqinghe9638@163.com}, \texttt{xinwangliu@nudt.edu.cn}, \texttt{kunlunhe@plagh.org}. ${\dagger}$ means equal contribution. ${\ddagger}$ represents corresponding author. 
}
}

\maketitle

\begin{abstract}
Although data-driven methods usually have noticeable performance on disease diagnosis and treatment, they are suspected of leakage of privacy  due to collecting data for model training. Recently, federated learning provides a secure and trustable alternative to collaboratively train model without any exchange of medical data among multiple institutes. Therefore, it has draw much attention due to its natural merit on privacy protection.  However, when heterogenous medical data exists between different hospitals, federated learning usually has to face with degradation of performance. In the paper, we propose a new personalized framework of federated learning to handle the problem. It successfully yields personalized models based on awareness of similarity between local data, and achieves better tradeoff between generalization and personalization than existing methods. After that, we further design a differentially sparse regularizer to improve communication efficiency during procedure of model training. Additionally, we propose an effective method to reduce the computational cost, which improves computation efficiency significantly. Furthermore, we collect $5$ real  medical datasets, including $2$ public medical image datasets and $3$ private multi-center clinical diagnosis datasets, and evaluate its performance by conducting nodule classification, tumor segmentation, and clinical risk prediction tasks. Comparing with $13$ existing related methods, the proposed method successfully achieves the best model performance, and meanwhile  up to $60\%$ improvement of communication efficiency.  Source code is public, and can be accessed at: \url{https://github.com/ApplicationTechnologyOfMedicalBigData/pFedNet-code}.
\end{abstract}


\IEEEpeerreviewmaketitle
%\balance

\section{Introduction}
\label{sect_introduction}
With proliferation of data, decision models generated by data-driven paradigm have shown remarkable performance on clinical diagnosis and treatment  \cite{PMID:37156936,PMID:36864252,PMID:36702948,PMID:35314822,gehrung2021triage}. Those medical models are usually trained by using multiple institutes' data, which may lead to leakage of privacy due to centralization of medical data. Recently, federated learning has shown significant advantages on alleviating such concerns, since it does not require exchange of medical data between hospitals\footnote{Federated learning usually consists of one \textit{server} and multiple \textit{clients}. Either server or client may represent a hospital.}  \cite{brauneck2023federated,warnat2021swarm,chen2021communication,wu2022federated,wu2022communication,froelicher2021truly}. More and more federated models have been developed for clinical diagnosis and treatment \cite{triple-negative,pati2022federated,bercea2022federated,Dayan2021,Bai2021}.  

Although federated learning has drawn much attention due to its superiority in privacy protection, its performance may face with degradation due to heterogenous data of different medical institutes \cite{kairouz2021advances}. For example, as one of data heterogeneity,  label unbalance widely exists between comprehensive hospital and specialized hospital, e.g. tumor hospital, which may highly impair model's performance \cite{chen2022personalized}. To mitigate such drawback of federated learning, personalized models are extensively investigated \cite{tan2022towards}, and extensive personalized methods such as \textit{FedAMP} \cite{huang2021personalized}, \textit{FedRoD} \cite{Chen2021OnBG}, \textit{APFL} \cite{deng2020adaptive}, \textit{FPFC} \cite{yu2022clustered}, \textit{IFCA} \cite{ghosh2022efficient}, \textit{pFedMe} \cite{dinh2020}, \textit{SuPerFed} \cite{hahn2022}, \textit{FedRep} \cite{collins2021exploiting} have been proposed. Although personalized models yielded by those methods have shown adaption to heterogenous data, they usually have three major limitations  in medical scenario, including \textit{sub-optimal performance}, \textit{requirement of prior assumption}, and \textit{limited flexibility}. Specifically, in terms of \textit{sub-optimal performance}, those methods usually work well in some general datasets such as MNIST\footnote{\url{http://yann.lecun.com/exdb/mnist/}}, and CIFAR\footnote{\url{https://www.cs.toronto.edu/~kriz/cifar.html}}, but  have unsatisfied performance in real medical scenario due to high complication of medicine \cite{deng2020adaptive,ghosh2022efficient,dinh2020}. Additionally, in terms of \textit{requirement of prior assumption}, existing methods may assume either clustering structure among clients \cite{ghosh2022efficient} or client's computing resources \cite{collins2021exploiting}, which may be either hard to know or not satisfied in real medical scenario. Moreover, in terms of \textit{limited flexibility}, some existing methods develop personalized models based on similarity network for clients' local data, but limit to few special topologies of network such as complete graph \cite{yu2022clustered,huang2021personalized}, star graph \cite{hanzely2020federated,hahn2022,Chen2021OnBG}, and may not achieve optimum for general medical scenarios directly. 

Moreover, another drawback of those existing personalized models is their limited usability for medical application. One of major reasons is that they are not able to handle both \textit{sharing} and \textit{personalized} components of medical data discriminately. Those components widely exist in clinical diagnosis and treatment \cite{chan2011personalized}. 
\begin{itemize}
\item \textbf{Sharing component.} Generally, clinical diagnosis and treatment are conducted by referring to international clinical guidelines such as  the Clinical Practice Guidelines (CPG)\footnote{\url{https://www.nccih.nih.gov/health/providers/clinicalpractice}}, which usually provide general clinical treatment for the whole population. Those clinical guidelines provide some necessary operations for some symptom, no matter who he/she is. For example,  when inflammation appears, patients have to conduct blood routine examination before clinical diagnosis. It is widely applied for all patients who appear the corresponding symptom. 
\item \textbf{Personalized component.} Every patient has his/her family genetic history, allergen, medication records etc. Such patient's information is unique, and should be considered carefully before offering clinical diagnosis and treatment. For example, when inflammation appears, the patient who has history of penicillin allergy cannot be treated by using penicillin. 
\end{itemize} Therefore, it is necessary for personalized models to capture above common and personalized characteristics of medical data. Unfortunately, few existing methods are designed to handle the case. Although Collins et al. develop a local model for every client with shared representation among clients \cite{collins2021exploiting}, it ignores the underlying similarity between clients who own similar data, and cannot allow to adjust personalization as need for medical scenario.

To mitigate limitations of those existing methods, we propose a new formulation of personalized federated learning, namely \textit{pFedNet}, and develop a flexible framework to obtain good adaption to heterogenous medical data. The personalized model\footnote{In the paper, we do not distinguish difference between the proposed \textit{model} and \textit{formulation}, and denote them by using \textit{pFedNet} indiscriminately.} consists of the sharing component and the personalized component, and is designed to capture both common and personalized characteristics of medical data. Note that \textit{pFedNet} builds personalized models based on \textit{similarity network} of clients' data, which is able to find underlying relation between personalized models. Additionally, it does not rely on any extra assumption on clients' clustering structure,  and any special topology of similarity network, and thus is more suitable to real medical scenario. 

Furthermore, we propose a new communication efficient regularizer to reduce workload of communication between clients and server, which can encourage elements of local update to own clustering structure, and thus improve communication efficiency. After that, we propose a new framework to optimize and obtain personalized models, which successfully reduces computational cost significantly. Finally, we collect $5$ real medical datasets, which includes $2$ public datasets of medical image and $3$ private datasets of medical records. $3$ classic medical tasks, including nodule classification, tumor segmentation, and clinical risk prediction, are conducted to evaluate the proposed method. Numerical results show that the proposed method successfully outperforms existing methods on performance, and meanwhile achieves up to $60\%$ promotion of communication efficiency. In summary, contributions of the paper are summarized as follows.
\begin{itemize}
\item We propose a new personalized federated model for the medical scenario, which is built based on awareness of similarity of between medical institutes' data, and  successfully captures both sharing and personalized characteristics of patients' data. 
\item We develop a new communication efficient regularizer to reduce workload of communication during learning of personalized model, and a new optimization framework to reduce the computational cost.
\item Extensive empirical studies have been conducted to evaluate the effectiveness of the proposed model and the optimization framework.
\end{itemize}

The paper is organized as follows. Section \ref{section_related_work} reviews related literatures. Section \ref{section_formulation} presents the proposed formulation, and explains its application. Section \ref{sect_cer} presents a communication efficient regularizer, which is able to decrease communication's workload during learning of models. Section \ref{sect_computation_efficient} presents an efficient method to reduce computation cost during federated learning. Section \ref{sect_empirical_studies} presents extensive empirical studies, and Section \ref{section_conclusion} concludes the paper.

\section{Related Work}
\label{section_related_work}
In the section, we review related literatures on methodology of personalized federated learning, and medical applications of federated learning. 
\subsection{Personalized Federated Learning}
Personalized federated learning combines benefits of personalized model and federated learning, while taking into account the unique characteristics and preferences of each client \cite{tan2022towards}. Its methodology usually has five branches including \textit{parameter decoupling}, \textit{knowledge distillation}, \textit{multi-task learning}, \textit{model interpolation}, and \textit{clustering}. Specifically, the branch of \textit{parameter decoupling} classifies parameters of model into two categories: base parameters and personalized parameters, where base parameters are shared between client and server, and personalized parameters are stored at client privately \cite{arivazhagan2019federated,bui2019federated,liang2020think}. The branch of \textit{knowledge distillation} transfers the knowledge from teacher's model to student's model, which can significantly enhance the performance of local models \cite{li2019fedmd,zhu2021data,lin2020ensemble,he2020group,bistritz2020distributed}. The branch of \textit{multi-task learning} views client's model as a task, and abstracts the learning procedure of personalized federated models as a multi-task learning task \cite{smith2017federated,corinzia2019variational,huang2021personalized,shoham2019overcoming}. The branch of \textit{model interpolation} simultaneously learns a global model for all clients, and a local model for every client. It usually makes tradeoff between the global model and local models to achieves the optimum of personalization \cite{hanzely2020federated,deng2020adaptive,diao2020heterofl}. The branch of \textit{clustering} aims to generating similar personalized model for clients who own similar data distribution \cite{sattler2020clustered,briggs2020federated,ghosh2022efficient,duan2021fedgroup}. The proposed method of personalized federated learning, namely \textit{pFedNet}, belongs to the branch of \textit{clustering}, but meanwhile allows to decouple parameters flexibly. Those existing methods in the branch of \textit{clustering} either conduct model learning and client clustering separately \cite{sattler2020clustered,briggs2020federated}, or conduct model learning based on prior assumption on clustering, e.g. selection of the number of clusters and specific clustering method \cite{ghosh2020efficient,huang2019patient}. Comparing with them, \textit{pFedNet} focuses on learning of personalized models, and meanwhile finds clustering structure among clients implicitly. It does not rely on prior assumption on clustering, and thus usually obtain better performance benefiting from good adaption of local data.     

Among those existing methods, most related methods include two groups: \textit{personalized model based on similarity} and \textit{personalized model with mixture of components}. Specifically, the first group consists of methods such as \textit{FPFC} \cite{yu2022clustered}, \textit{FedAMP} \cite{huang2021personalized}, \textit{L2GD} \cite{hanzely2020federated}, \textit{FedRoD} \cite{Chen2021OnBG}, \textit{SuPerFed} \cite{hahn2022} etc. These existing methods yield personalized model based on some special topologies of similarity network of clients' local data, e.g. complete graph and star graph,  limiting their applications in medical scenarios. For example,  both \textit{FPFC} and \textit{FedAMP} generate personalized model based on the complete graph. \textit{L2GD}, \textit{FedRoD} , and \textit{SuPerFed} produce personalized model based on the star graph. Comparing with those methods, the proposed method, namely \textit{pFedNet}, does not have this limitation, and can work on any topology. Moreover, the second group consists of methods such as \textit{FedRep} \cite{collins2021exploiting}, who produce personalized model with mixture components. However,  \textit{FedRep} does not consider the similarity between client's data, and assumes every client has sufficient computing resources to update personalized component. Comparing with \textit{FedRep}, the proposed method supports flexible combination of personalized and sharing components, and meanwhile achieves better performance based on awareness of pairwise similarity between clients' local data. 

\subsection{Federated Learning in Medical Applications}
In recent years, several studies have explored the use of federated learning in medicine, and present promising results \cite{pfitzner2021federated,kaissis2021end}. One of the main medical applications is the development of predictive models for disease diagnosis and treatment \cite{Bai2021,Dayan2021}. For example, Bai et al. propose an open source framework for medical artificial intelligence, and offer diagnosis of COVID-19 by using federated learning method \cite{Bai2021}. Dayan et al. develop federated learning method to predict clinical outcomes in patients with COVID-19.  Additionally, another area where federated learning has shown promising is analysis of medical images \cite{medicalimaging,zhu2022federated}. For instance, Kaissis et al. review recent emerging methods on privacy preservation of medical images analysis, and discusses drawbacks and limitations of those existing methods \cite{medicalimaging}. Moreover, a general federated learning framework, namely PriMIA \cite{kaissis2021end} is developed, and its advantages on privacy protection, securely aggregation, and encrypted inference have been evaluated by conducting classification of paediatric chest X-rays images. Similarly, a federated learning method for predicting histological response to neoadjuvant chemotherapy in triple-negative breast cancer is recently proposed \cite{triple-negative}.  An automatic tumor boundary detector for the rare disease of glioblastoma has been proposed by using federated learning \cite{pati2022federated}, which presents impressive performance. Similar to those studies, the paper focuses on medical scenario, but provides a general and flexible learning framework for personalized models. The proposed formulation is inspired by the real procedure of clinical treatment, has wide applications for disease diagnosis and medical data analysis, and is not limited to a specific disease like those existing methods. 

\section{Formulation}
\label{section_formulation}
In the section, we first present similarity network for representing heterogenous client's data, and then develop a new formulation of personalized federated learning. Finally, we present the framework of alternative optimization to solve the proposed formulation.

\subsection{Personalized Representation based on Similarity Network}
\begin{figure}
\setlength{\abovecaptionskip}{0pt}
\setlength{\belowcaptionskip}{0pt}
\centering 
\includegraphics[width=0.8\columnwidth]{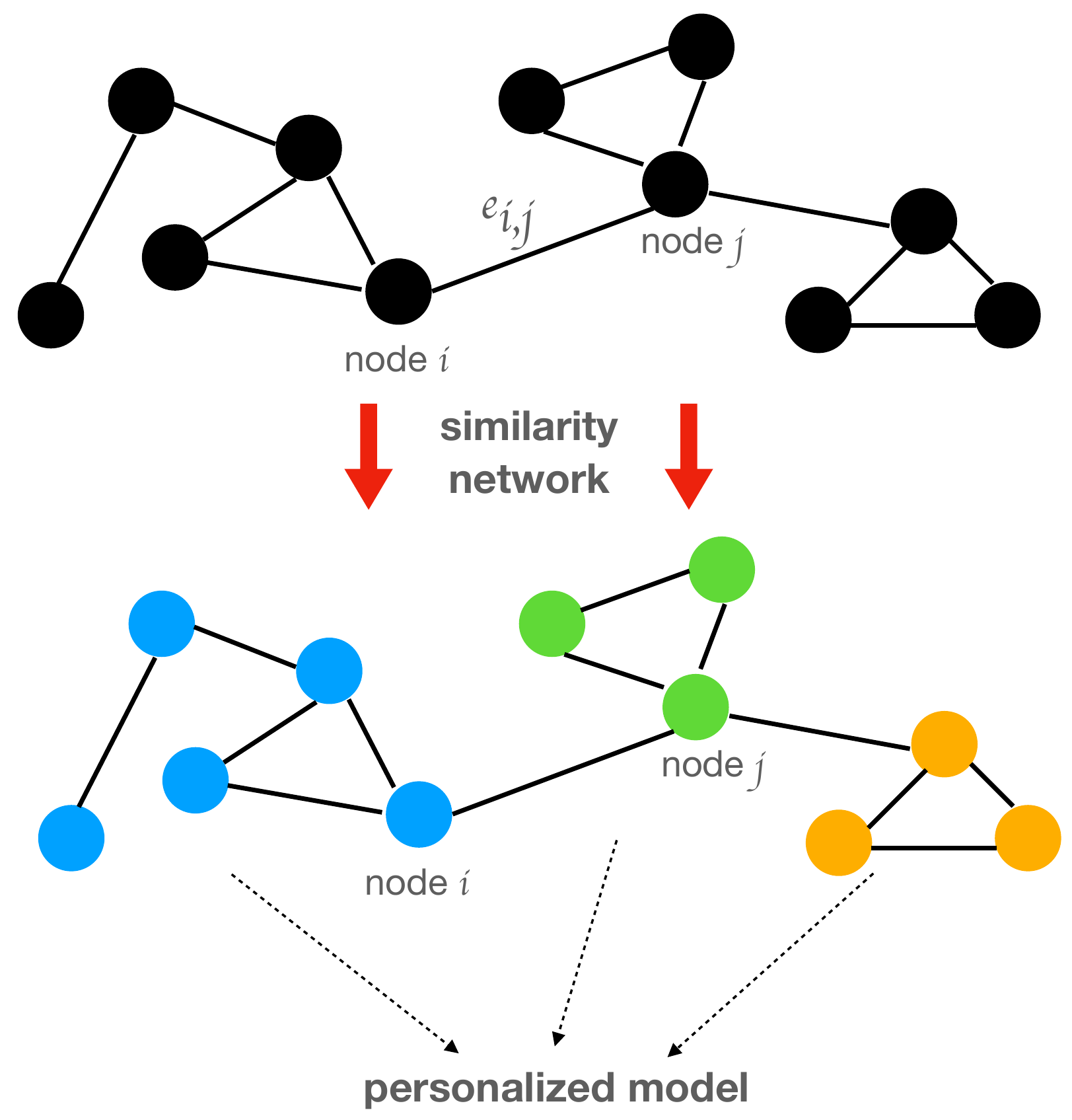}
\caption{Personalized representation based on similarity network}
\label{figure_personalized_representation_similarity_network}
\end{figure}

In the work, similarity of distribution of local data is measured under data space. Local dataset of every node is  represented by using a \textit{sketch} matrix. The similarity of local data distribution is measured based on the distance between sketch matrices.  The similarity network $\Gcal := \{\Ncal, \Ecal\}$ is usually built by using the \textit{K-Nearest Neighbors} (KNN) method. As illustrated in Figure \ref{figure_personalized_representation_similarity_network}, the network $\Gcal$ is generated by given a $K$, where $\Ncal:=\{1,2, ..., N\}$ represents the node set, consisting of $N$ nodes. $\Ecal:=\{e_{i,j} : i\in\Ncal, j {~} \text{is the node $i$'s neighbour}\}$ represents the edge set, consisting of $M$ edges. 

Besides, major notations used in the paper are summarized as follows for easy understanding of mathematical details.
\begin{itemize}
\item Bold and lower letters such as $\a$ represent a vector. Bold and upper letters such as $\X$ represents a matrix.
\item Lower letters such as $f(\cdot)$ and $h(\cdot)$ represent a function. Other letters such as $n$, $N$, $M$ represents a scalar value.
\item $\Ncal$ and $\Ecal$ represent a set, and $\Dcal_n$ represent a data distribution for the $n$-th client.
\item $\odot$ represents Hadamard of two matrices. $\lrnorm{\cdot}_p$ represent the $p$-th norm of a vector.
\item $\nabla$ represents the gradient operator, and $\nabla f(\cdot)$ represent the gradient of $f$.
\item $[\a]_+$ means negative elements of $\a$ is replaced by $0$, and non-negative elements of $\a$ do not make any change.
\end{itemize}

\subsection{\textit{pFedNet}: Formulation of Personalized Federated Learning}
Given the similarity network $\Gcal$, and constant matrices $\M\in\RR^{d\times d_1}$ and $\N\in\RR^{d\times d_2}$, the proposed personalized federated learning, namely \textit{pFedNet}, is finally formulated by
\begin{align}
\nonumber
\min_{\x, \substack{\{\x^{(n)},\z^{(n)}\}_{n=1}^N}} \frac{1}{N}\sum_{n\in\Ncal} f_n\lrincir{\x^{(n)}; \Dcal_n} + \lambda \sum_{\substack{e_{i,j} \in \Ecal,\\ \forall i,j\in\Ncal}} \lrnorm{\z^{(i)} - \z^{(j)}}_p,
\end{align} subject to:
\begin{align}
\nonumber
\x^{(n)} = \M\x + \N \z^{(n)},{~~~~}\forall n\in\Ncal, \x\in\RR^{d_1}, \z^{(n)}\in\RR^{d_2}.
\end{align} Here, $f_n\lrincir{\x^{(n)}; \Dcal_n}$ represents the local loss at the $n$-th client, where $\x^{(n)}$ represents the personalized model, and $\Dcal_n$ represents the local data. For example, it can be instantiated by $f_n\lrincir{\x^{(n)}; \Dcal_n} = \sum_{(\a, y)\sim\Dcal_n} \log\lrincir{\frac{1}{1+e^{-y\a^\top\x^{(n)}}}}$ for the logistic regression task. 

\textit{Note that $\x$ and $\z^{(n)}$ represent the sharing and personalized component of the personalized model $\x^{(n)}$, respectively.} In terms of statistical machine learning models like SVM and logistic regression \cite{shalev2014}, their sharing component may be weights of features like \textit{inflammation}, \textit{diarrhoea}, and \textit{vomiting} etc, and their personalized component may be weights of features like \textit{family genetic history} and \textit{allergen} etc. In terms of deep learning models like dense net \cite{huang2017densely} and u-net \cite{unetmiccai2015}, their sharing component may be weights of layers of feature extraction, and their personalized component may be weights of layers of classifier. 

\begin{figure}
\setlength{\abovecaptionskip}{0pt}
\setlength{\belowcaptionskip}{0pt}
\centering 
\subfigure[\textit{CHD} dataset.]{\includegraphics[width=0.48\columnwidth]{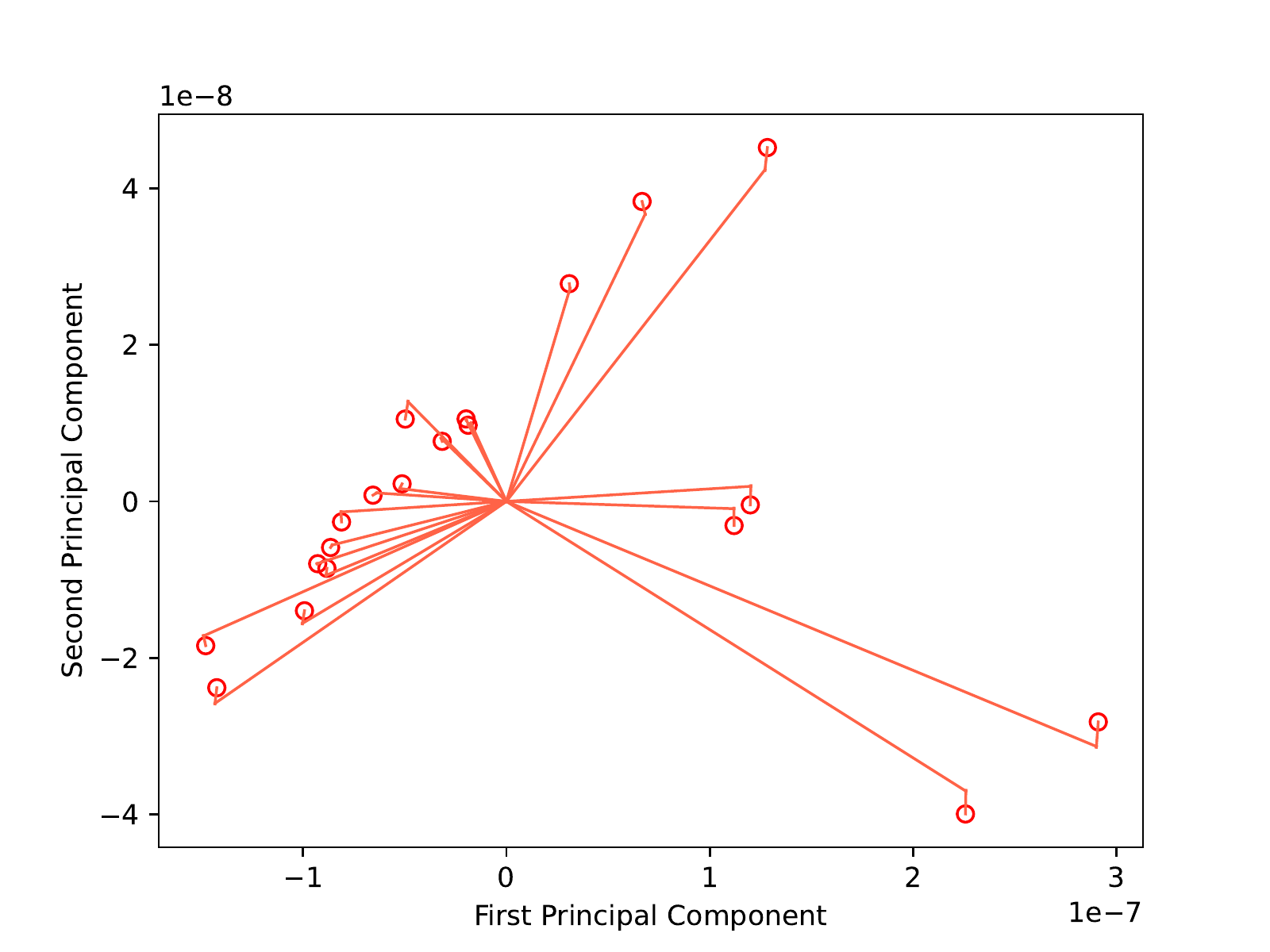}\label{figs_pFedNet_CHDModels}}
\subfigure[\textit{Covid19} dataset.]{\includegraphics[width=0.48\columnwidth]{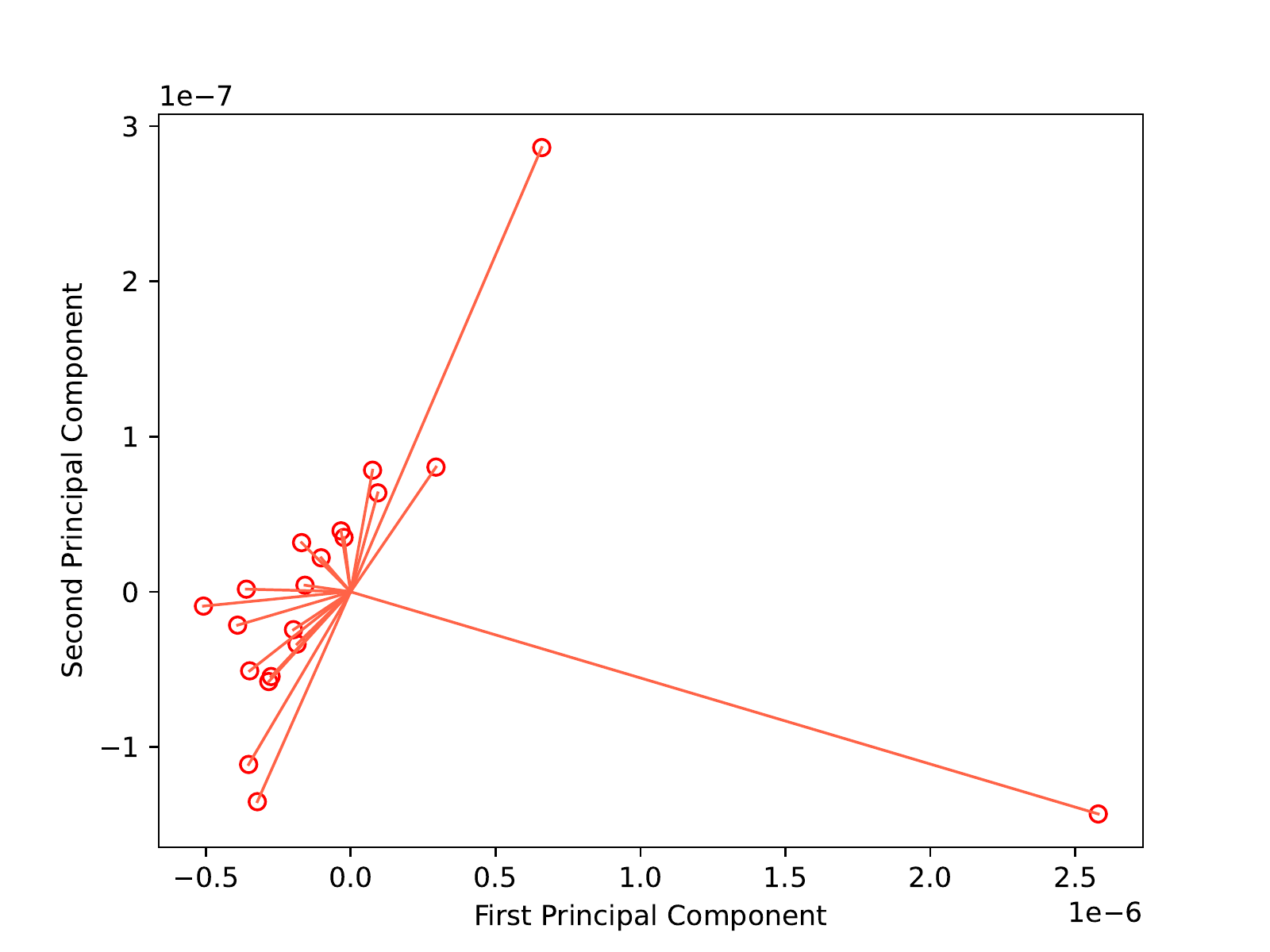}\label{figs_pFedNet_Covid19Models}}
\caption{Personalized models yielded by \textit{pFedNet} appear more similarity with the increase of $\lambda$ for datasets: \textit{CHD} and \textit{Covid19}. Red circle corresponds to a personalized model of a client, and there are $20$ clients. The x-axis and y-axis of figures represents the first and second principal component of the yielded personalized model, respectively.}
\label{figure_pFedNet_CHD_Covid19}
\end{figure}

The proposed formulation has wide application in medical analysis and clinical diagnosis and treatment. Generally, doctors offer diagnosis and treatment service according to patients' medical records and the international clinical guidelines. It is a natural scenario for personalized model with mixture of components to conduct clinical decision. 
\begin{itemize}
\item \textbf{Personalized component}. Since every patient has his/her unique medical record including family genetic history, allergen, medication records etc, the personalized component of model, e.g. $\z^{(n)}$ is necessary to capture characteristics of such data. 
\item \textbf{Sharing component}. The international clinical guideline usually provide a general solution to conduct diagnosis and treatment. For example, blood routine examination is required when inflammation appears. The sharing component of model, e.g. $\x$ is necessary to capture such common characteristics.
\end{itemize} Additionally, some special diseases such as regional disease, and occupational disease also need personalized model with sharing component to conduct clinical decision \cite{chan2011personalized}. Specifically, the treatment of regional and occupational disease needs to consider the location and occupation of patients, respectively, which corresponds to the personalized component of the clinical decision model. Besides, all patients should also be offered some basic treatment such as alleviation of inflammation, which corresponds the sharing component of the model. In a nutshell, the formulation provides a general and flexible framework to conduct personalized federated learning. 
\begin{itemize}
\item \textbf{Generality}. No matter statistical machine learning models such as \textit{ridge regression}, \textit{logistic regression}, \textit{support vector machine} etc \cite{shalev2014} or deep learning models such as \textit{dense net} \cite{huang2017densely} and \textit{u-net} \cite{unetmiccai2015} etc, the formulation can be instantiated by specifying the local loss function $f_n$.   
\item \textbf{Flexibility}. First, it is flexible to select the sharing component of the federated model, namely $\x$ and the personalized component $\z^{(n)}$ according to the specific task. Second, it is flexible to choose a $\lambda$ for making tradeoff between personalized need and global requirement.
\end{itemize}

Additionally, $\lambda$ with $\lambda>0$ is a given hyper-parameter, which controls the personalization of federated model. When $\lambda \rightarrow 0$, more personalization is allowed, that is $\z^{(n)}$ has much difference. The personalization decays with the increase of $\lambda$. Almost all $\z^{(n)}$ with $n\in\{1, 2, \cdots, N\}$ tends to be same for a large $\lambda$. As shown in Figure \ref{figure_pFedNet_CHD_Covid19}, this has been verified by conducting logistic regression on datasets: \textit{CHD} and \textit{Covid19}\footnote{Details of those datasets are shown in Section \ref{subsect_experimental_settings}.}. In the case, there are $20$ clients and $1$ server in the federated network. We observe the similar phenomenon. That is, every personalized model is different with $\lambda=0$ (represented by red circle), and those models begin to gather together with the increase of $\lambda$ (represented by red lines). When $\lambda$ is sufficiently large, all personalized models converge to a point, which means all personalized models indeed become same.

\begin{figure}
\setlength{\abovecaptionskip}{0pt}
\setlength{\belowcaptionskip}{0pt}
\centering 
\subfigure[$\Q$ matrix.]{\includegraphics[width=0.8\columnwidth]{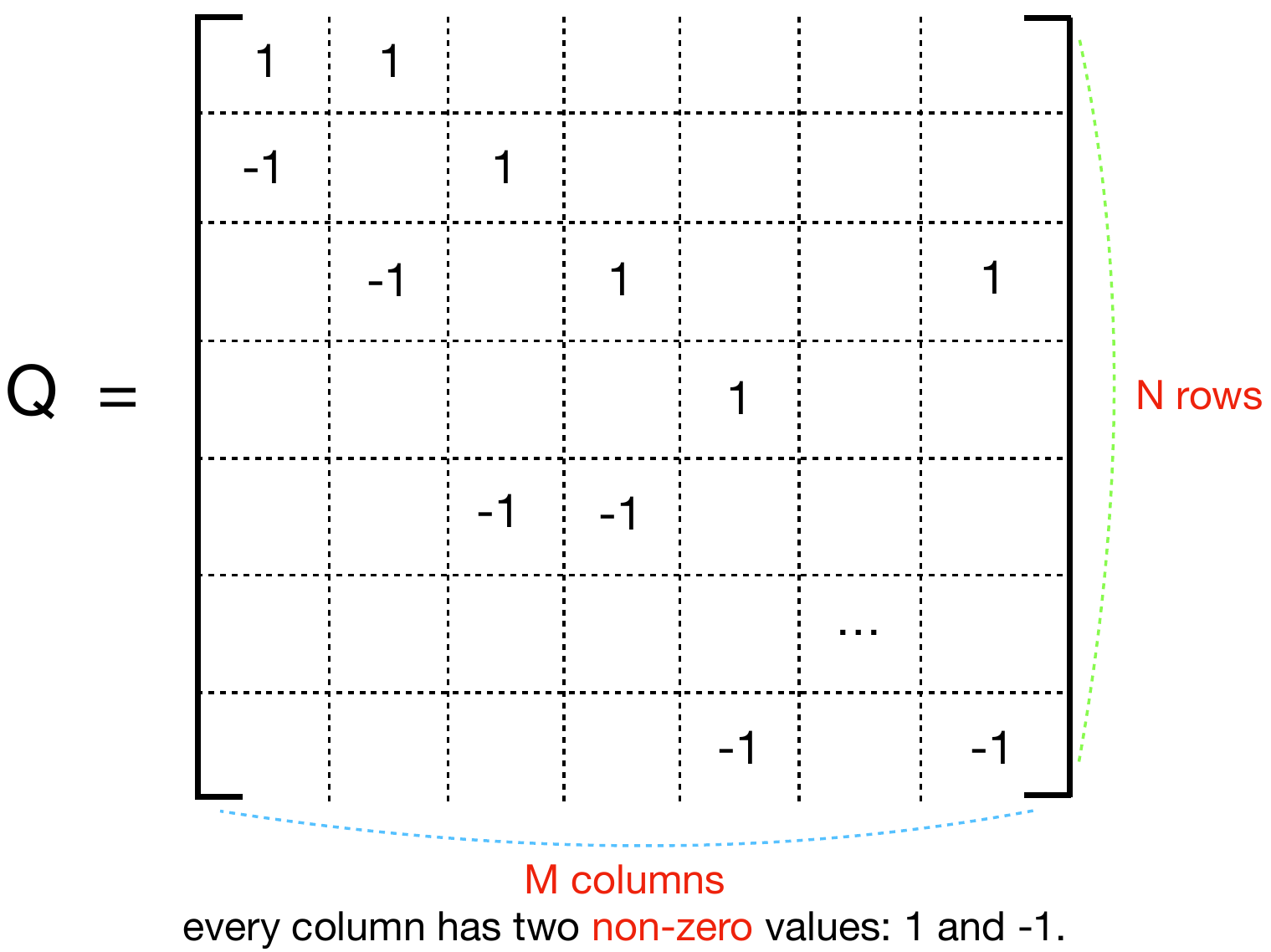}\label{figs_Q_matrix}}
\subfigure[$\M$ and $\N$ matrices.]{\includegraphics[width=0.75\columnwidth]{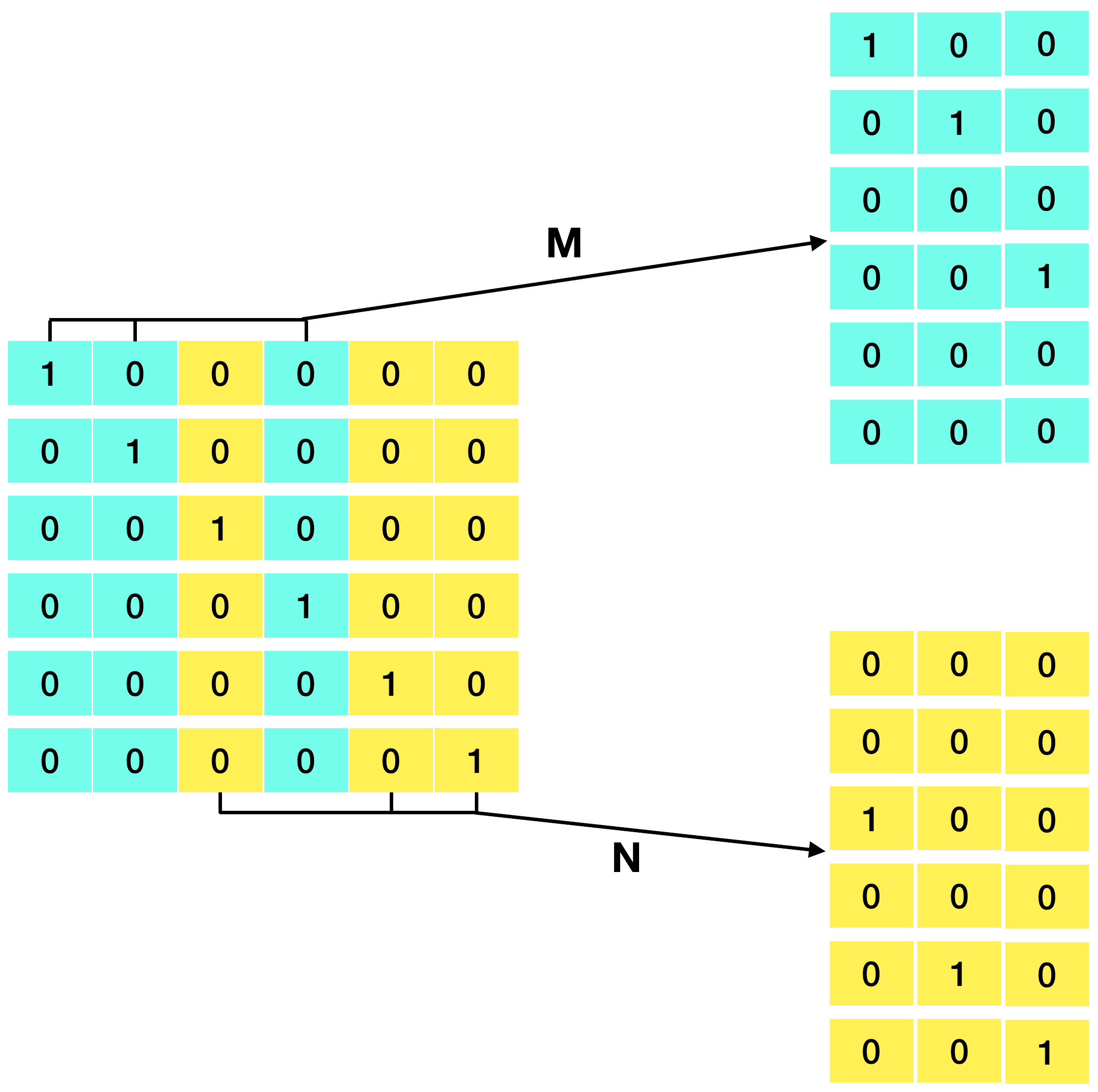}\label{figs_MNmatrix}}
\caption{Illustration of matrices $\Q$, $\M$, and $\N$.}
\label{figure_QMN}
\end{figure}

Note that the formulation can be equally transformed as follows.
\begin{align}
\label{equa_formulation_pFedNet}
\min_{\substack{\{\z^{(n)}\}_{n=1}^N}, \x} \frac{1}{N}\sum_{n=1}^N f_n\lrincir{\x^{(n)}; \Dcal_n} + \lambda \lrnorm{\Z\Q}_{1,p},
\end{align} subject to:
\begin{align}
\nonumber
\x^{(n)} = \M\x + \N \z^{(n)}.
\end{align} Here, $\Z\in\RR^{d_2 \times N}$, $\Q\in\RR^{N\times M}$. $N$ and $M$ represent the total number of nodes and edges in the network $\Gcal$, respectively. $\Z$ represents some a variable matrix, consisting of $N$ variables as columns, that is, $\Z = \left [\z^{(1)}, \z^{(2)}, ..., \z^{(N)} \right ]$. As shown in Figure \ref{figure_QMN}, both $\M$ and $\N$ have special structure, where every row of them has at most one non-zero value, and the non-zero value is $1$. $p\in\{1,2,\infty\}$.  $\Q$ is the given auxiliary matrix, which has $M$ columns and every column has two non-zero values: $1$ and $-1$. Note that $\lrnorm{\cdot}_{1,p}$ is denoted by $\ell_{1,p}$ norm. Given a matrix $\U\in\RR^{d_2 \times M}$, it is defined by 
\begin{align}
\nonumber
\lrnorm{\U}_{1,p} := \sum_{m=1}^M \lrnorm{\U_{:, m}}_p.
\end{align}

\begin{figure*}
\setlength{\abovecaptionskip}{0pt}
\setlength{\belowcaptionskip}{0pt}
\centering 
\includegraphics[width=1.98\columnwidth]{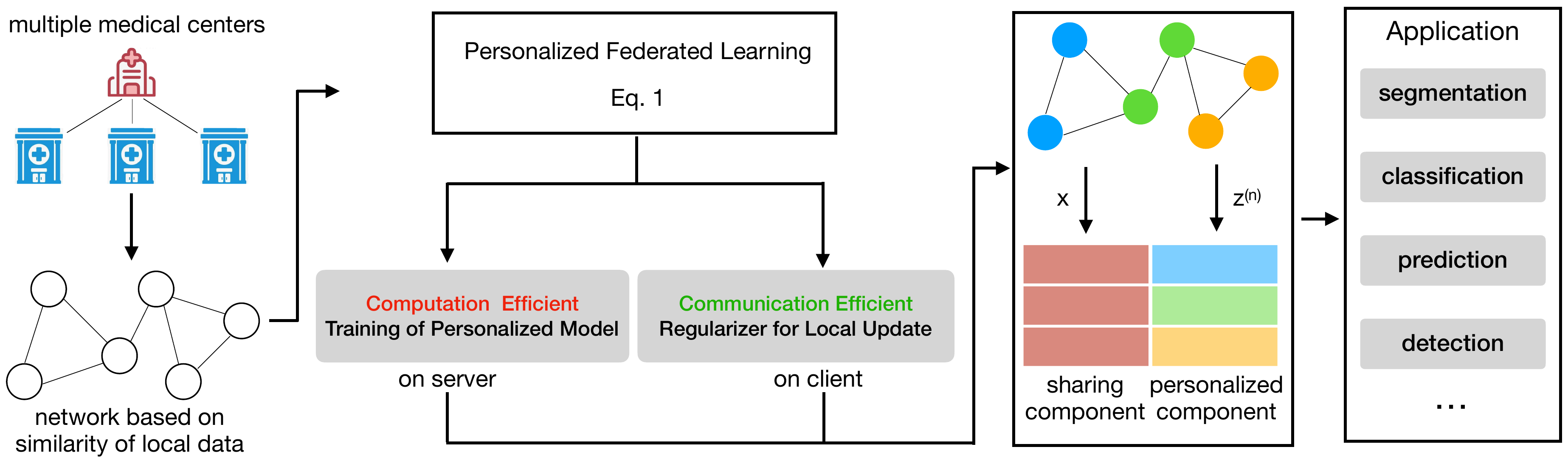}
\caption{Learning framework of the personalized federated model with sharing component.}
\label{figure_workflow}
\end{figure*}

\subsection{Optimization}
The formulation \ref{equa_formulation_pFedNet} is difficult to be solved due to $3$ reasons. First, the optimization variables may be highly non-separable due to $\Q$. As we have shown, every column of $\Q$ corresponds an edge of the similarity network $\Gcal$, which implies that the corresponding personalized component, e.g. $\z^{(i)}$ and $\z^{(j)}$, corresponding to nodes of such edge has dependent relation. Second, the loss function may be highly non-smooth, because the regularizer is sum of norms. Third, the number of optimization variables is large, when the network $\Gcal$ has a large number of nodes and edges.  Generally, the formulation \ref{equa_formulation_pFedNet} is solved by alternative optimization. The variable $\{\x^{(1)}, \x^{(2)}, ..., \x^{(N)}\}$ is obtained by alternatively optimizing $\x$ and $\{\z^{(1)}, \z^{(2)}, ..., \z^{(N)}\}$.

\textbf{Optimizing $\x$ by given $\Z$.} $\x$ is optimized by solving the following problem:
\begin{align}
\nonumber
\min_{\substack{\x\in\RR^{d_1}}} \frac{1}{N}\sum_{n=1}^N f_n\lrincir{\M\x + \N \z^{(n)}; \Dcal_n}.
\end{align} By using the data-driven stochastic optimization method such as SGD \cite{Shalev2014Understanding}, we need to perform the following problem to obtain $\x$ iteratively.
\begin{align}
\label{equa_optimizeX_givenZ}
\min_{\substack{\x\in\RR^{d_1}}} \frac{1}{N}\sum_{n=1}^N \lrangle{\M^\top\g_t^{(n)}, \x} + \frac{1}{2\eta_t}\lrnorm{\x - \x_t}^2,
\end{align} where $\g_t^{(n)}$ is a stochastic gradient of $f_n$ with $\M\x + \N \z_t^{(n)}$ by using data drawn from the local dataset $\Dcal_n$.

\textbf{Optimizing $\Z$ by given $\x$.} $\Z$ is optimized by solving the following problem:
\begin{align}
\nonumber
\min_{\substack{\Z\in\RR^{d_2\times N}}} \frac{1}{N}\sum_{n=1}^N f_n\lrincir{\M\x + \N \z^{(n)}; \Dcal_n} + \lambda \lrnorm{\Z\Q}_{1,p}.
\end{align}  By using the data-driven stochastic optimization method such as SGD \cite{Shalev2014Understanding}, we need to perform the following problem:
\begin{align}
\nonumber
\min_{\substack{\Z\in\RR^{d_2\times N}}} \frac{1}{N}\sum_{n=1}^N \lrangle{\N^\top \g_t^{(n)}, \z^{(n)}} + \lambda \lrnorm{\Z\Q}_{1,p} + \frac{\lrnorm{\Z - \Z_t}_F^2}{2\eta_t}.
\end{align} $\g_t^{(n)}$ is a stochastic gradient of $f_n$ with $\M\x_t + \N \z^{(n)}$ by using stochastic data  drawn from the local dataset $\Dcal_n$. Suppose $\G_t = \left [ \g_t^{(1)}, \g_t^{(2)}, ..., \g_t^{(N)} \right ]$, and $\Z$ is optimized by performing the following problem:
\begin{align}
\label{equa_optimizeZ_givenX}
\min_{\substack{\Z\in\RR^{d_2\times N}}} \frac{\1_d^\top\lrincir{\lrincir{\N^\top\G_t} \odot \Z}\1_N}{N} + \lambda \lrnorm{\Z\Q}_{1,p} + \frac{\lrnorm{\Z - \Z_t}_F^2}{2\eta_t}.
\end{align} Here, $\odot$ means Hadamard product of two matrices.

\begin{algorithm}[!]
    \caption{Compute local stochastic gradient at the $n$-th client for the $t\mathrm{+}1$-th iteration.}
    \label{algo_grad_client}
    \begin{algorithmic}[1]
        %\Require The number of total iterations $T$, and the initial parameter $\x_1$.
        %\algotext{\textbf{On the $n$-th client for the $t+1$-th iteration:}}
        \State Receive the personalized model $\y_t^{(n)} := \M\x_t + \N\z_t^{(n)}$ from the server.
        \State Randomly sample an instance $\a\sim\Dcal_n$, and compute the stochastic gradient $\g^{(n)}_t = \nabla f(\y_t^{(n)};\a)$ with $\a\sim\Dcal_n$.
        \State Send $\g^{(n)}_t$ to the server.
    \end{algorithmic}
\end{algorithm} 

\begin{algorithm}[!]
    \caption{Train personalized models at the server.}
    \label{algo_server}
    \begin{algorithmic}[1]
        \Require The number of total iterations $T$, and the initial model $\x_1$, $\z_1^{(n)}$ with $n\in\{1,2,\cdots,N\}$.
        %\algotext{\textbf{On the server:}}
        \State Deliver the model $\y_1^{(n)} = \M\x_1 + \N\z_1^{(n)}$ to all client $n$ with $n\in\{1,2, ..., N\}$.
        \For {$t=1,2, ..., T$}
            \State Collect stochastic gradient $\G_t = \left [ \g_t^{(1)}, \g_t^{(2)}, ..., \g_t^{(N)} \right ]$ from all client $n$ with $n\in\{1,2, ..., N\}$.
            \State Update the global model $\x$ by solving \ref{equa_optimizeX_givenZ}.
            \State Update the personalized model $\Z$ by solving \ref{equa_optimizeZ_givenX}.
            \State Deliver the parameter $\y_{t+1}^{(n)} = \M\x_{t+1} + \N\z_{t+1}^{(n)}$ to every client.
        \EndFor
        \Return $\x_{T+1}^{(n)} = \M\x_{T+1} + \N\z_{T+1}^{(n)}$ with $n\in\{1,2, ..., N\}$.
      \end{algorithmic}
\end{algorithm}

\textbf{Federated optimization}. According to the above optimization steps, the stochastic gradient $\G_t$ is obtained at client in the scenario of federated learning. Details are illustrated in Algorithm \ref{algo_grad_client}. Moreover, the personalized model $\x^{(n)}$ with $n\in\{1,2,\cdots,N\}$ is optimized at the server, and details are shown in Algorithm \ref{algo_server}. Unfortunately, the federated optimization has two major drawbacks. 
\begin{itemize}
\item \textbf{Heavy workload of communication}. Since every client has to transmit the stochastic gradient, e.g. $\g_t^{(n)}$ to the server, the communication workload will be unbearable for a large $d$. Especially, deep neural network models usually own more than millions of parameters, the transmission of such gradient will lead to high cost of communication.
\item \textbf{High cost of computation}. Since the \textit{sum-of-norms} regularizer leads to high non-separability and non-smoothness of the objective loss, the computation cost is high. The optimization of personalized model is time-consuming and even unbearable.
\end{itemize}

To mitigate those drawbacks, we first develop a communication efficient method for every client to transmit the stochastic gradient. Additionally, we then propose a computation efficient method for the server to update the personalized model. In summary, the learning framework of the personalized federated learning is illustrated in Figure \ref{figure_workflow}.

\begin{figure}
\setlength{\abovecaptionskip}{0pt}
\setlength{\belowcaptionskip}{0pt}
\centering 
\subfigure[$\nabla_{t+1}^{(n)}$  without clustering structures]{\includegraphics[width=0.49\columnwidth]{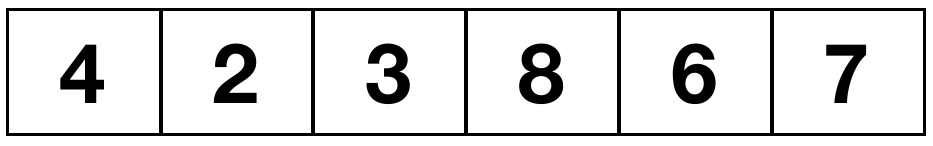}\label{figure_intro_gradient_clustering_without_clustering}}
\subfigure[Difference between elements of $\nabla_{t+1}^{(n)}$ (without \textit{CER}) is dense.]{\includegraphics[width=0.9\columnwidth]{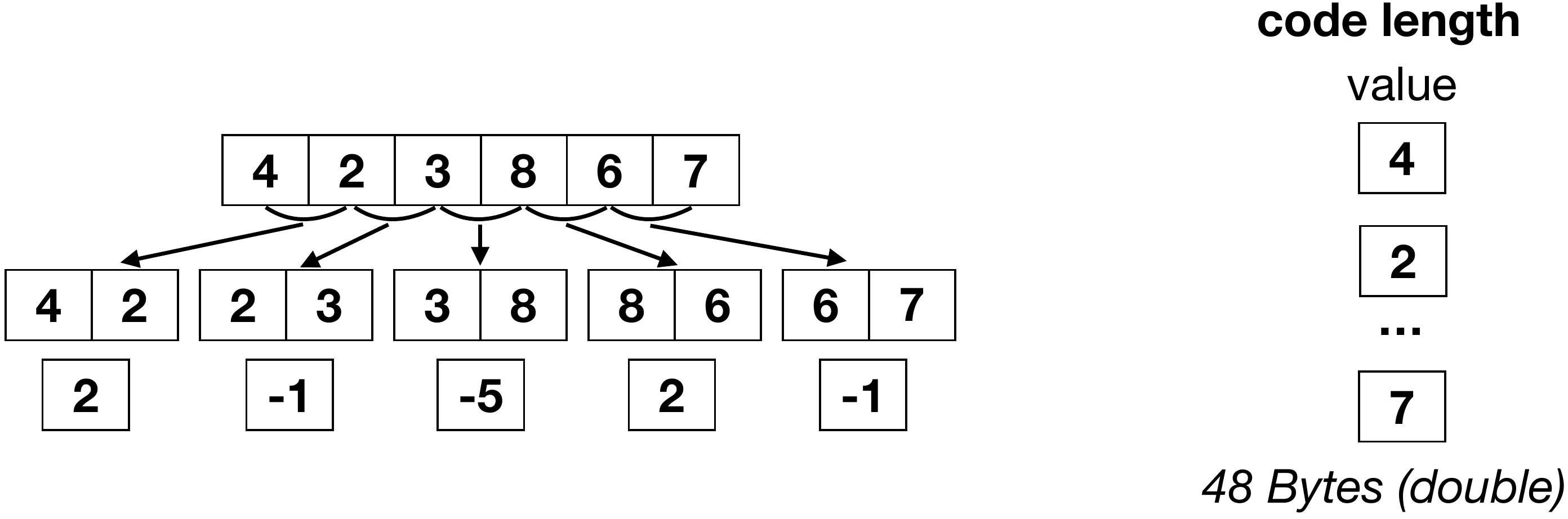}\label{figure_intro_gradient_without_clustering_details}}
\subfigure[$\nabla_{t+1}^{(n)}$ with clustering structures]{\includegraphics[width=0.49\columnwidth]{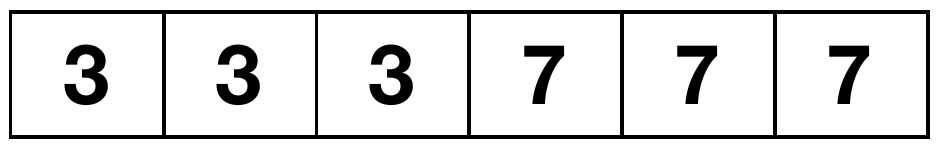}\label{figure_intro_gradient_clustering_with_clustering}}
\subfigure[Difference between elements of $\nabla_{t+1}^{(n)}$  (with \textit{CER}) is \textbf{sparse}.]{\includegraphics[width=0.98\columnwidth]{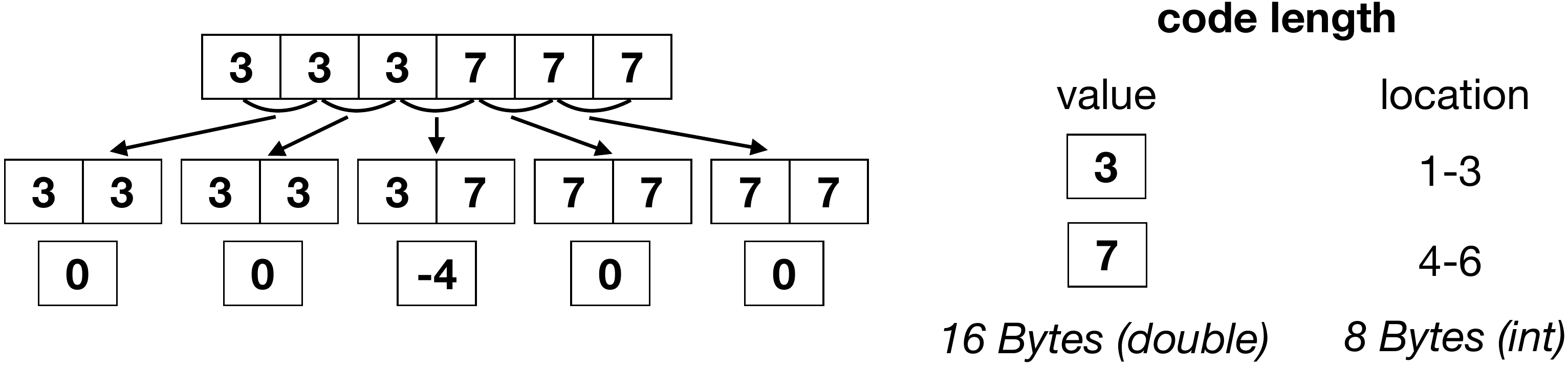}\label{figure_intro_gradient_with_clustering_details}}
\caption{The illustrative example shows that $\nabla_{t+1}^{(n)}$ with clustering structures can be compressed by using fewer bits, and thus the code length is reduced effectively. \textbf{Our basic idea is to make the difference between  elements of $\nabla_{t+1}^{(n)}$ sparse.}}
\label{figure_intro_gradient_clustering} 
\end{figure}

\section{Communication Efficient Update of Model} 
\label{sect_cer}
In the section, we first propose a communication efficient regularizer, which encourages elements of update of local model to own clustering structure, and improves the communication efficiency effectively. Then, we develop an ADMM method \cite{boyd:admm} to conduct the update of local model.   

\subsection{CER: Communication Efficient Regularizer}
In the work, we propose a communication efficient method, which can let $\g_t^{(n)}$ be encoded by using few bits. Since the code length of $\g_t^{(n)}$ is much reduced, the communication efficiency is significantly increased. The basic idea is to induce the clustering structure of elements of $\g_t^{(n)}$ by using differential sparsity regularizer. The regularizer encourages the update of local model $\nabla_{t+1}^{(n)}$ to own clustering structures. Figure \ref{figure_intro_gradient_clustering} presents an illustrative example. According to Figures \ref{figure_intro_gradient_clustering_without_clustering} and \ref{figure_intro_gradient_clustering_with_clustering},  when the elements of $\nabla_{t+1}^{(n)}$ own clustering structures, they can be encoded by using fewer bits. Its code length can be reduced a lot. The update of parameter can be transmitted from clients  and the server efficiently. According to Figures \ref{figure_intro_gradient_without_clustering_details} and \ref{figure_intro_gradient_with_clustering_details}, our basic idea is to let the difference between the elements of $\nabla_{t+1}^{(n)}$ be sparse, which encourages the elements of $\nabla_{t+1}^{(n)}$ to have clustering structures. Comparing with the gradient quantization methods in the previous studies, the proposed method is able to find a good tradeoff between the convergence performance and the communication efficiency.  

To improve the communication efficiency, we propose a new method to conduct the update of the parameter, which is formulated as
\begin{align}
\nonumber
\nabla_{t+1}^{(n)} = \frac{\y_t^{(n)} - \v}{\eta_t}, 
\end{align} where $\v$ is obtained by performing the following problem: 
\begin{align}
\nonumber
& \v \\ \nonumber
= & \argmin_{\y\in\RR^d} \lrangle {\g^{(n)}_t, \y} \mathrm{+} \underbrace{\gamma \lrnorm{\bLambda\lrincir{\y \mathrm{-} \y_t^{(n)}}}_1}_{\text{\scriptsize  communication efficient  regularizer}} \mathrm{+} \frac{\lrnorm{\y \mathrm{-} \y_t^{(n)}}^2}{2\eta_t}. 
\end{align}  Here, $\g^{(n)}_t$ is a stochastic gradient, which is obtained by using the local data at the $n$-th client. The given full rank square matrix $\bLambda\in\mathbb{R}^{d\times d}$ is defined by
\begin{align}
\nonumber
\bLambda := \begin{bmatrix}
 1&  -1&  &  & \\ 
 &  1&  -1&  & \\ 
 &  &   \cdots &  & \\ 
 &  &  & 1 & -1 \\ 
 &  &  &  & 1
\end{bmatrix}.
\end{align} Notice that $\bLambda$ is a full rank square matrix, whose smallest singular value, denoted by $\sigma$,  is positive, that is, $\sigma > 0$. The proposed communication efficient regularizer is an $\ell_1$ norm square. It punishes the difference between elements of $\nabla_{t+1}^{(n)}$, and encourages them  to be small or even zero. Thus, those corresponding elements of $\nabla_{t+1}^{(n)}$ are very similar or even identical. That is, the elements of $\nabla_{t+1}^{(n)}$ own clustering structures. Exploiting the clustering structures, $\x_{t+1}$ can be compressed by using few bits, and thus improves the communication efficiency in the distributed setting. Adjacent elements of $\nabla_{t+1}^{(n)}$ have tiny difference, and appear clustering structures.  

\begin{figure*}
\setlength{\abovecaptionskip}{0pt}
\setlength{\belowcaptionskip}{0pt}
\centering 
\subfigure[$\gamma=50$, $\bLambda \g_t^{(n)}$ v.s. $\bLambda \nabla_{t+1}^{(n)}$ ]{\includegraphics[width=0.49\columnwidth]{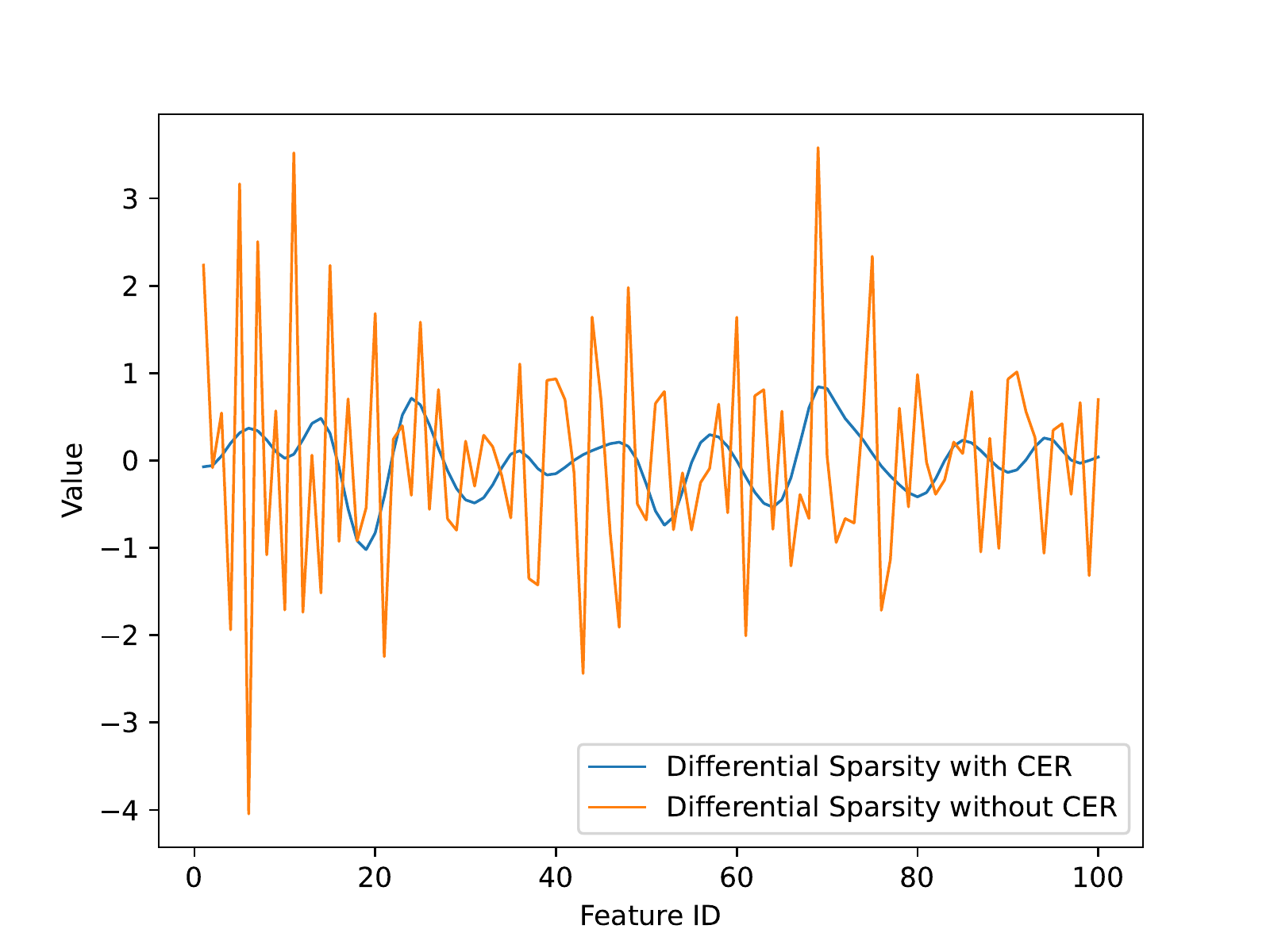}\label{figs_diff_spar_gamma_50}}
\subfigure[$\gamma=100$, $\bLambda \g_t^{(n)}$ v.s. $\bLambda \nabla_{t+1}^{(n)}$]{\includegraphics[width=0.49\columnwidth]{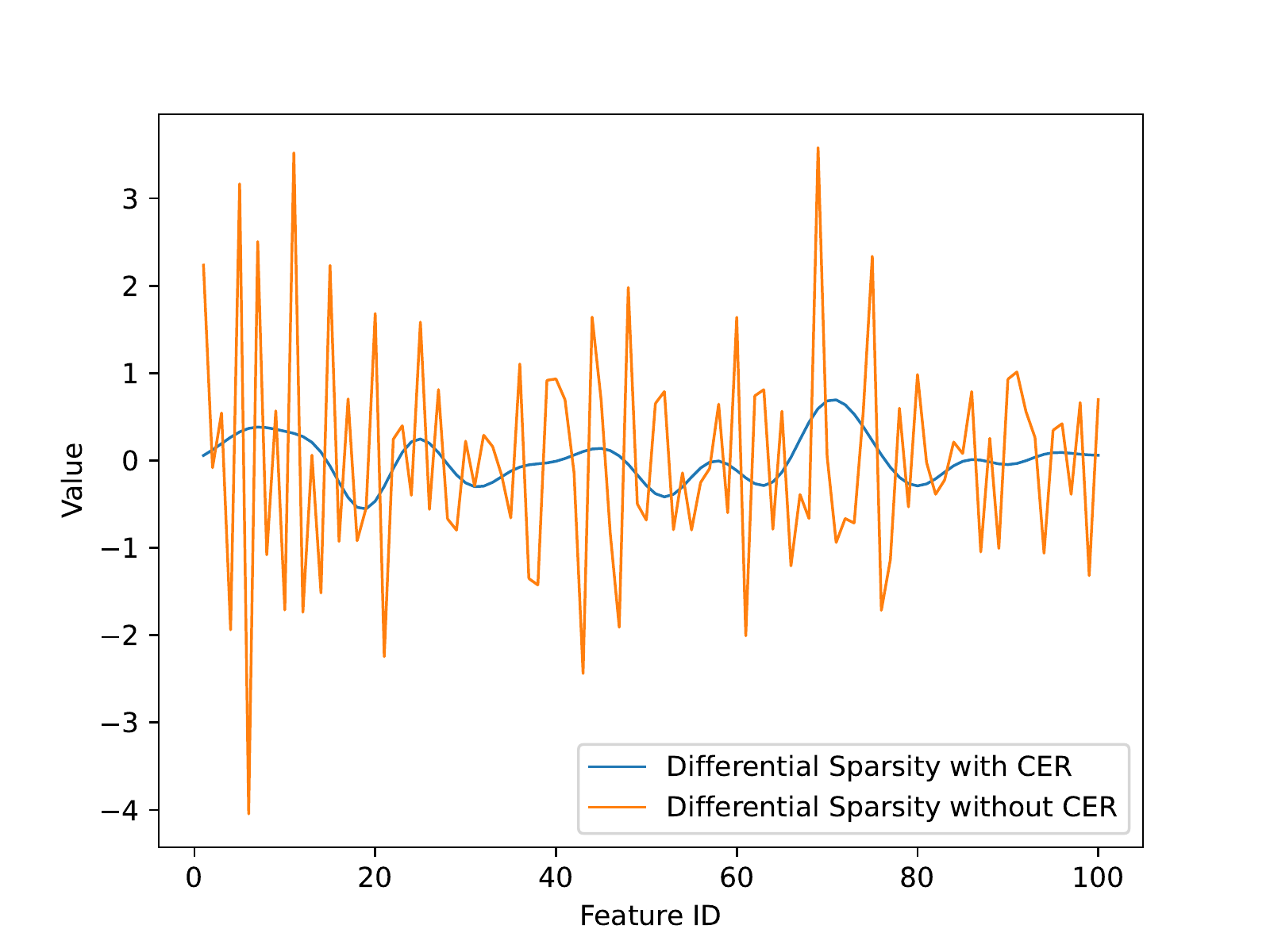}\label{figs_diff_spar_gamma_100}}
\subfigure[$\gamma=500$, $\bLambda \g_t^{(n)}$ v.s. $\bLambda \nabla_{t+1}^{(n)}$]{\includegraphics[width=0.49\columnwidth]{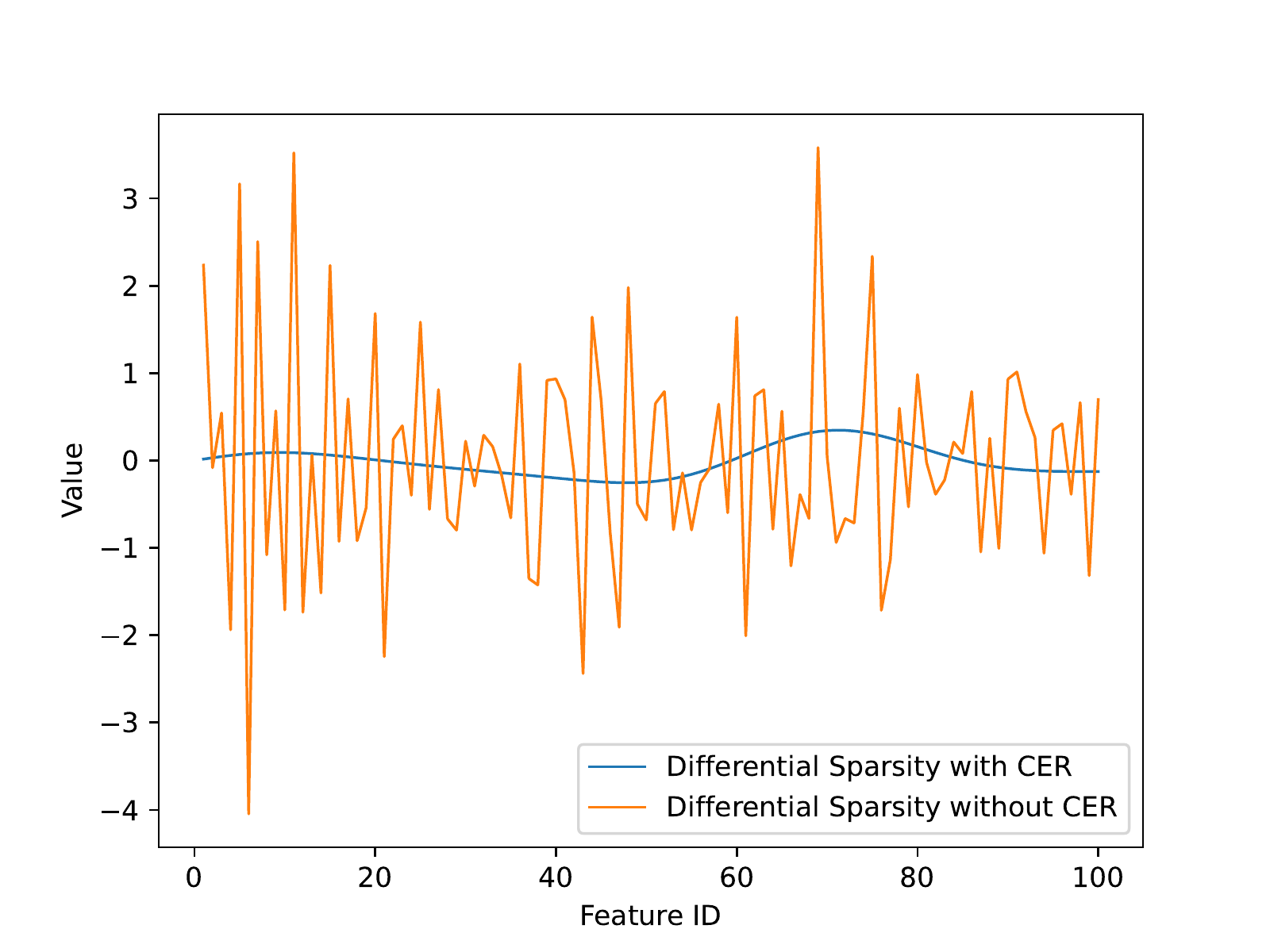}\label{figs_diff_spar_gamma_500}}
\subfigure[$\gamma=1000$, $\bLambda \g_t^{(n)}$ v.s. $\bLambda \nabla_{t+1}^{(n)}$]{\includegraphics[width=0.49\columnwidth]{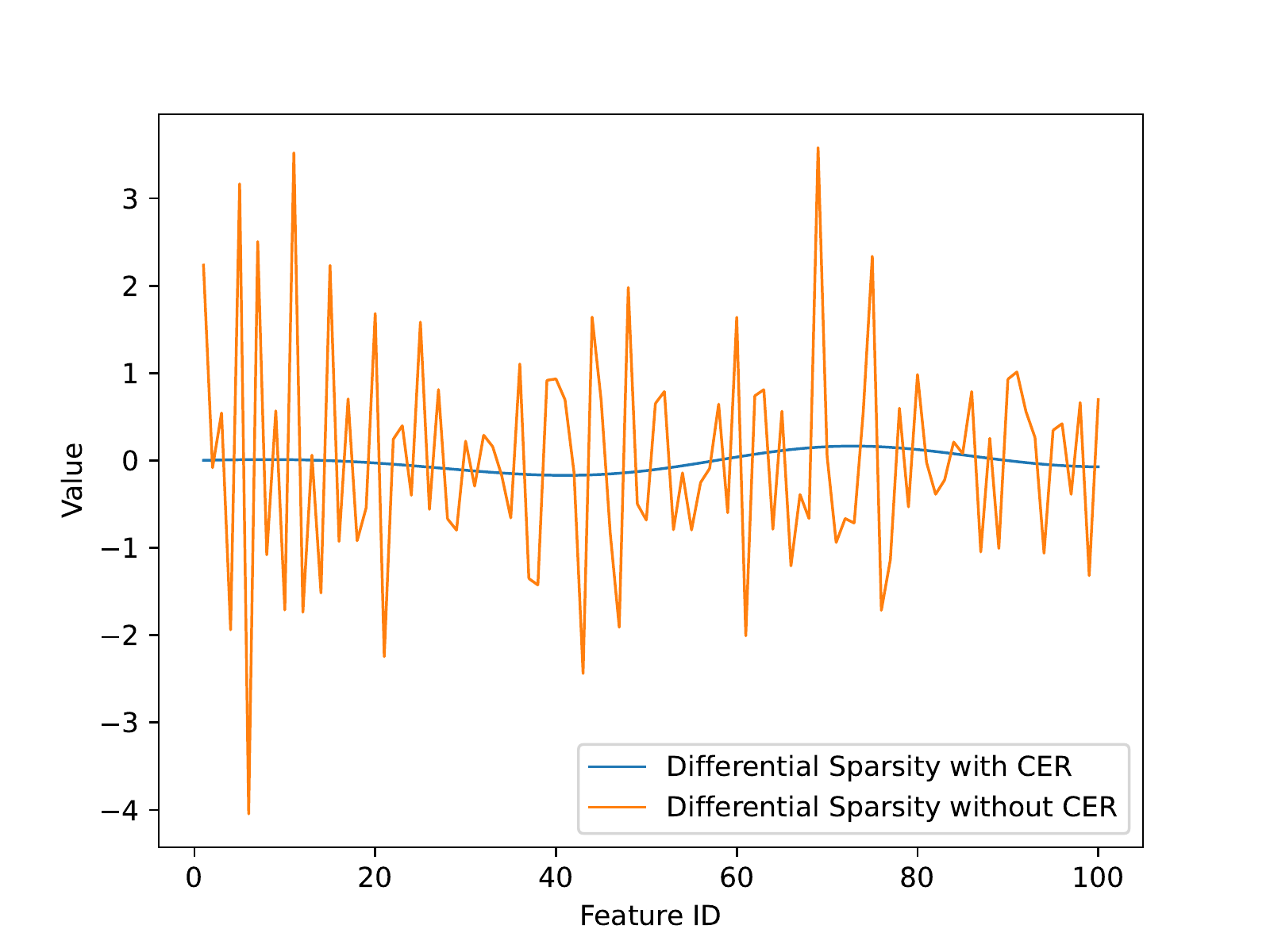}\label{figs_diff_spar_gamma_1000}}
\subfigure[$\gamma=50$, $\g_t^{(n)}$ v.s. $\nabla_{t+1}^{(n)}$]{\includegraphics[width=0.49\columnwidth]{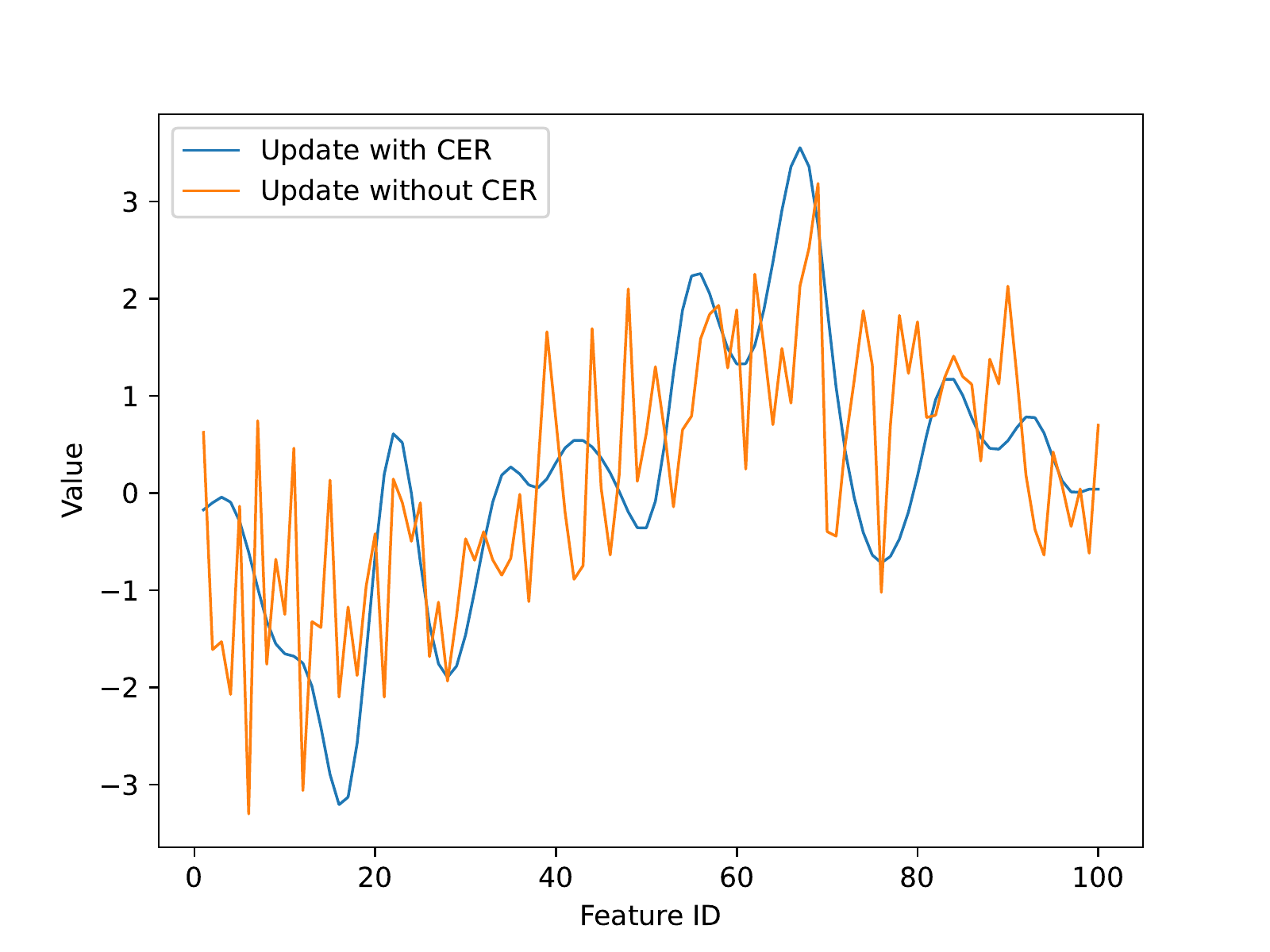}\label{figs_update_gamma_50}}
\subfigure[$\gamma=100$, $\g_t^{(n)}$ v.s. $\nabla_{t+1}^{(n)}$]{\includegraphics[width=0.49\columnwidth]{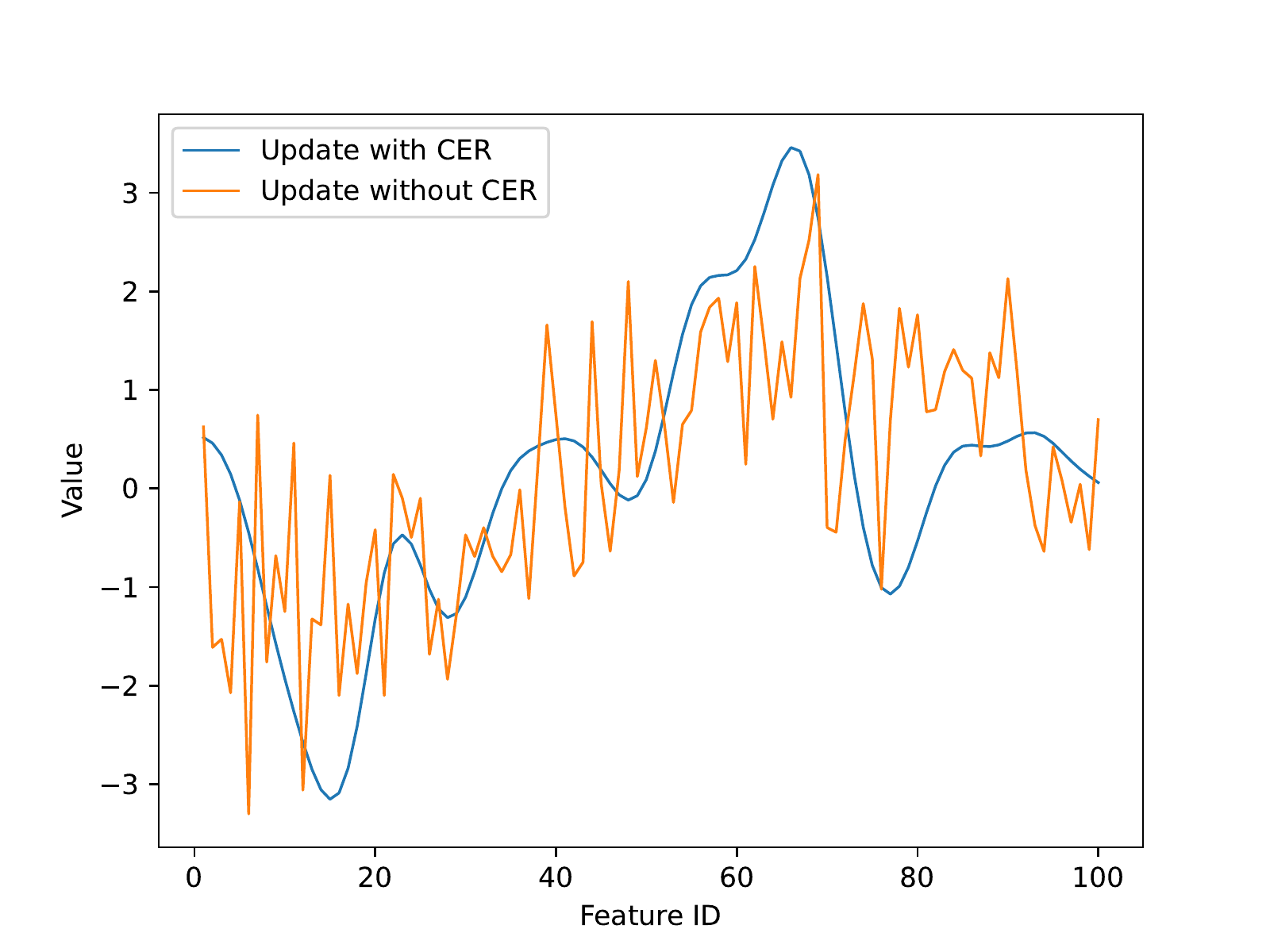}\label{figs_update_gamma_100}}
\subfigure[$\gamma=500$, $\g_t^{(n)}$ v.s. $\nabla_{t+1}^{(n)}$]{\includegraphics[width=0.49\columnwidth]{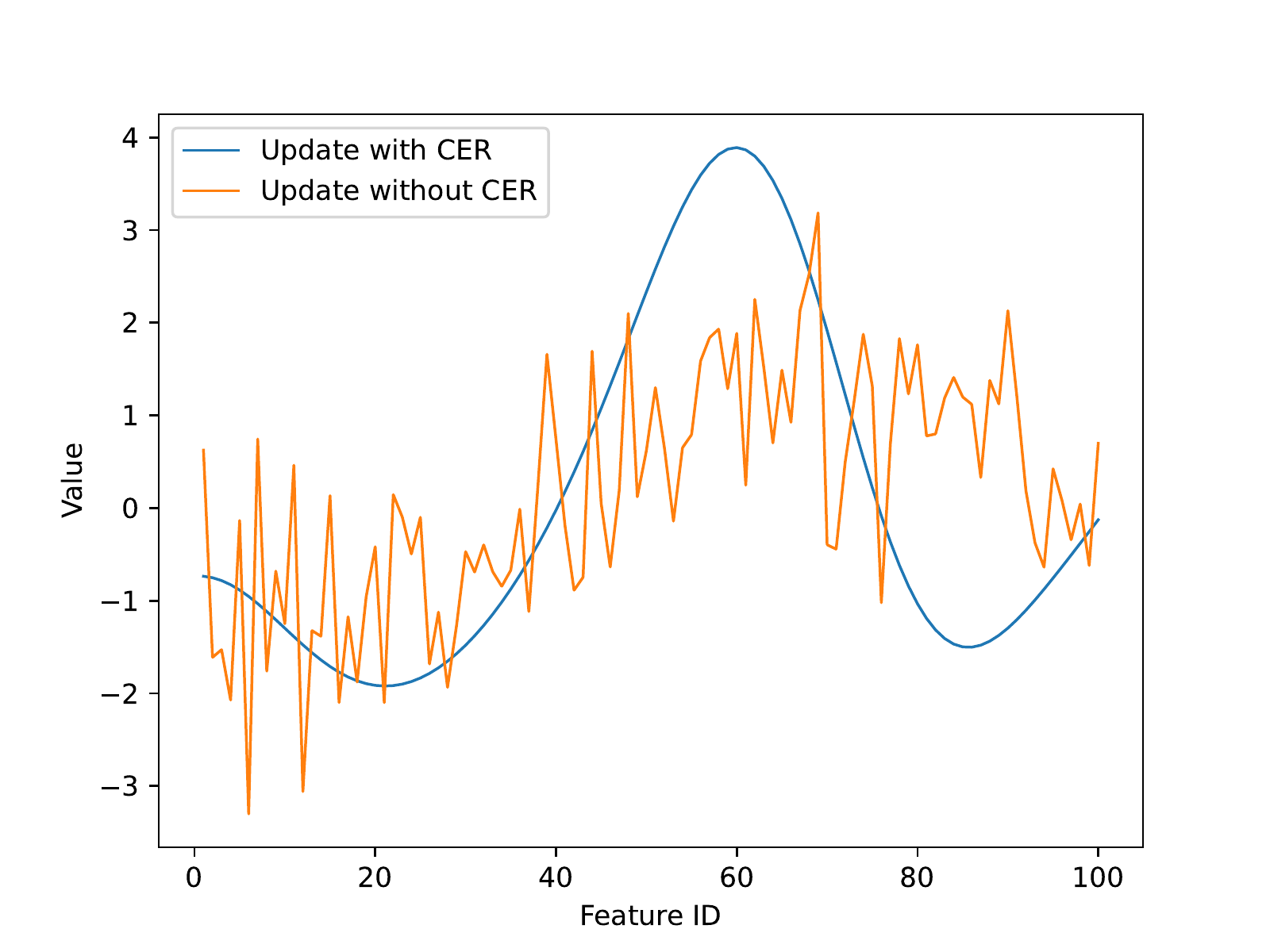}\label{figs_update_gamma_500}}
\subfigure[$\gamma=1000$, $\g_t^{(n)}$ v.s. $\nabla_{t+1}^{(n)}$]{\includegraphics[width=0.49\columnwidth]{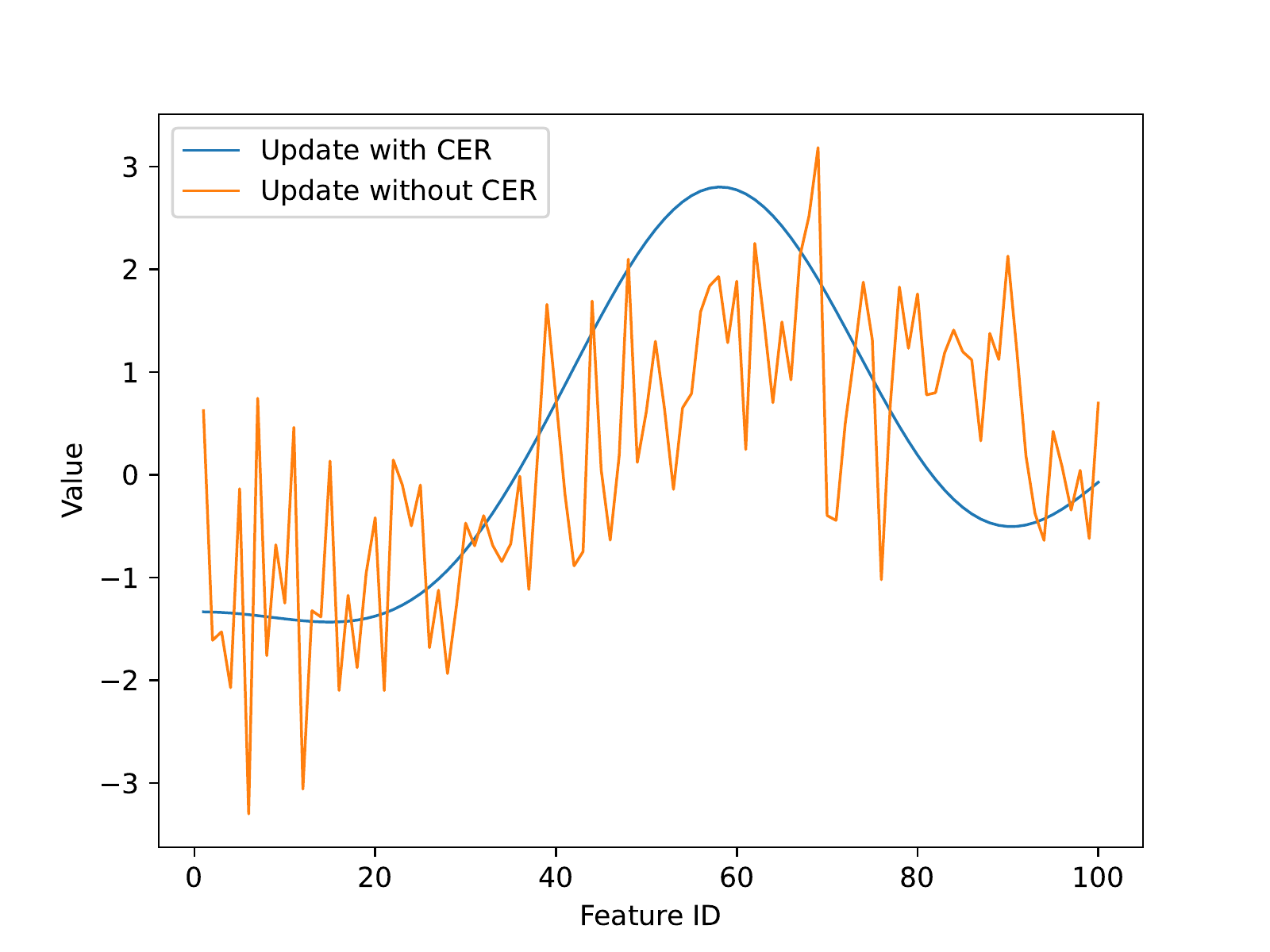}\label{figs_update_gamma_1000}}
\caption{Comparison of \textit{differential sparsity}, namely $\bLambda \g_t^{(n)}$ and $\bLambda \nabla_{t+1}^{(n)}$, and \textit{local update}, namely $\g_t^{(n)}$ and $\nabla_{t+1}^{(n)}$. $\g_t^{(n)}$ is obtained by setting $\gamma=0$, that is local update of model without \textit{CER}.  $\nabla_{t+1}^{(n)}$ is obtained by setting $\gamma>0$, that is local update of model with \textit{CER}.}
\label{figure_illustrative_cer}
\end{figure*}

We present more explanations by taking an example. As illustrated in Figure \ref{figure_illustrative_cer}, we generate local update of the personalized model with $100$ features (orange lines in Figures \ref{figs_update_gamma_50}-\ref{figs_update_gamma_1000}) and difference of its elements (orange lines in Figures \ref{figs_diff_spar_gamma_50}-\ref{figs_diff_spar_gamma_1000}). As we can see,  the differential sparsity, e.g. $\bLambda \nabla_{t+1}^{(n)}$ (blue lines in Figures \ref{figs_diff_spar_gamma_50}-\ref{figs_diff_spar_gamma_1000}) becomes sparse significantly with the increase of $\gamma$ (Figures \ref{figs_diff_spar_gamma_50}-\ref{figs_diff_spar_gamma_1000}). It verifies that the proposed method, namely \textit{CER} successfully encourages difference between elements of local update to be sparse. Meanwhile, we find that $\nabla_{t+1}^{(n)}$ is similar to $\g_t^{(n)}$ for a small $\gamma$ (Figure \ref{figs_update_gamma_50}), and a large $\gamma$ leads to a significant trend (Figures  \ref{figs_update_gamma_50}-\ref{figs_update_gamma_1000}). As illustrated in Figures \ref{figs_diff_spar_gamma_1000} and \ref{figs_update_gamma_1000}, we observe that elements of local update become similar when their difference is sparse, and thus appear clustering structures (peak and bottom of the blue curve). It leads to much easier compression than the original local update. 

Note that there is a trade-off between the accuracy and communication efficiency. When elements of a gradient are partitioned into more clusters, the higher accuracy of the gradient is guaranteed. Meanwhile, the gradient has to be encoded by using more bytes, thus leading to the decrease of the communication efficiency.  

\subsection{Optimizing $\nabla_{t+1}^{(n)}$}
As we have shown, $\nabla_{t+1}^{(n)}$ is obtained by 
\begin{align}
\nonumber
\nabla_{t+1}^{(n)} = \frac{\x_t - \v}{\eta_t}.
\end{align} $\v$ is obtained by performing the following problem.
\begin{align}
\nonumber
& \v \\ \nonumber
= & \argmin_{\y\in\RR^d} \lrangle {\g^{(n)}_t, \y} + \gamma \lrnorm{\bLambda\lrincir{\y - \y_t^{(n)}}}_1 + \frac{\lrnorm{\y - \y_t^{(n)}}^2}{2\eta_t} \\ \nonumber
= & \argmin_{\y\in\RR^d} \lrnorm{\bLambda\lrincir{\y - \y_t^{(n)}}}_1 + \frac{\lrnorm{\y - \lrincir{\y_t^{(n)} - \eta_t\g_t^{(n)}}}^2}{2\eta_t\gamma}.
\end{align} $\v$ can be obtained by performing the following problem.
\begin{align}
\nonumber
\min_{\y\in\RR^d, \r\in\RR^d} \underbrace{\lrnorm{\r}_1 + \frac{\lrnorm{\y - \lrincir{\y_t^{(n)} - \eta_t\g_t^{(n)}}}^2}{2\eta_t\gamma}}_{=: g(\r, \y)},
\end{align} subject to:
\begin{align}
\nonumber
\r = \bLambda \y - \bLambda \y_t^{(n)}.
\end{align} The Lagrangian multiplier of $g(\r, \y)$ is 
\begin{align}
\nonumber
& L(\r, \y, \bomega)  \\ \nonumber
= & g(\r, \y) + \lrangle{\bomega, \r - \bLambda \y + \bLambda\y_t^{(n)}} + \frac{\rho}{2}\lrnorm{\r - \bLambda \y + \bLambda\y_t^{(n)}}^2.
\end{align} ADMM\cite{boyd:admm} is used to solve the above optimization problem, which consists of update of $\r$, $\y$, and $\omega$, iteratively.

\textbf{Update of $\r$.} Given $\y_j$, $\bomega_j$, $\r_{j+1}$ is updated by performing the following problem.
\begin{align}
\nonumber
&\r_{j+1} = \argmin_{\r\in\RR^d} L(\r, \y_j, \bomega_j) \\ \nonumber
= & \argmin_{\r\in\RR^d} \lrnorm{\r}_1 + \lrangle{\bomega_j, \r} + \frac{\rho}{2}\lrnorm{\r - \bLambda \y_j + \bLambda\y_t^{(n)}}^2 \\ \nonumber
= & \argmin_{\r\in\RR^d} \lrnorm{\r}_1 + \frac{\rho}{2}\lrnorm{\r - \lrincir{\bLambda \y_j - \bLambda\y_t^{(n)} - \frac{1}{\rho}\bomega_j}}^2 \\ \nonumber
= & \Prox_{\rho, \lrnorm{\cdot}_1}\lrincir{\bLambda \y_j - \bLambda\y_t^{(n)} - \frac{1}{\rho}\bomega_j} \\ \label{equa_update_r}
\mathrm{=} & \left[\bLambda \lrincir{\y_j - \y_t^{(n)}} \mathrm{-} \frac{\bomega_j}{\rho} \mathrm{-} \rho\right]_\mathrm{+}  \mathrm{-} \left[\bLambda \lrincir{\y_t^{(n)} - \y_j} \mathrm{+} \frac{\bomega_j}{\rho} \mathrm{-} \rho\right]_\mathrm{+}.
\end{align} Here, $\Prox$ represents the \textit{proximal operator}, which is defined by 
\begin{align}
\nonumber
\Prox_{\nu, \phi}(\a) := \argmin_{\b} \phi(\b) + \frac{\nu}{2}\lrnorm{\b - \a}^2.
\end{align} The last equality holds due to
\begin{align}
\nonumber
\Prox_{\nu, \lrnorm{\cdot}_1}(\a) = (\a - \nu)_+ - (-\a - \nu)_+, 
\end{align} where $\b_+$ means that negative elements of $\b$ are set by $0$, and other non-negative elements do not change.

\begin{algorithm}[!]
    \caption{Communication efficient update of local models on the $n$-th client for the $t\mathrm{+}1$ iteration.}
    \label{algo_efficient_update_nabla}
    \begin{algorithmic}[1]
        \Require A positive $\gamma$ to improve communication efficiency. Given $\x_t$, $\P$, and $\bSigma$ such that $\bLambda^\top\bLambda = \P\bSigma\P^{-1}$.
        \State Receive the personalized model $\y_t^{(n)} := \M\x_t + \N\z_t^{(n)}$ from the server.
        \State Randomly sample an instance $\a\sim\Dcal_n$, and compute the stochastic gradient $\g^{(n)}_t = \nabla f(\y_t^{(n)};\a)$ with $\a\sim\Dcal_n$.
           \For {$j=0,1,2, ..., J-1$}
            \State Update $\r_{j+1}$ by performing Eq. \ref{equa_update_r}.
            \State Update $\y_{j+1}$ by performing Eq. \ref{equa_update_y_admm}. 
           \State Update $\bomega_{j+1}$ by performing Eq. \ref{equa_update_omega_admm}.
            \EndFor
        \State Compute $\nabla_{t+1}^{(n)}$ with $\nabla_{t+1}^{(n)} = \frac{\y_t^{(n)} - \y_J}{\eta_t}.$
        \State Send $\nabla_{t+1}^{(n)}$ to the server.
      \end{algorithmic}
\end{algorithm}

\textbf{Update of $\y$.} Given $\r_{j+1}$, $\bomega_j$, $\y_{j+1}$ is updated by performing the following problem:
\begin{align}
\nonumber
&\y_{j+1} = \argmin_{\y\in\RR^d} L(\r_{j+1}, \y, \bomega_j) \\ \nonumber
= & \argmin_{\y\in\RR^d} \frac{\lrnorm{\y - \lrincir{\y_t^{(n)} - \eta_t\g_t^{(n)}}}^2}{2\eta_t\gamma} - \lrangle{\bomega_j, \bLambda \y} \\ \nonumber 
& {~~~~~~~~} + \frac{\rho}{2}\lrnorm{\r_{j+1} - \bLambda \y + \bLambda\y_t^{(n)}}^2 \\ \nonumber
= & \argmin_{\y\in\RR^d} \small \lrnorm{\bLambda \y \mathrm{-} \left[\r_{j+1} \mathrm{+} \bLambda\y_t^{(n)} \mathrm{+} \frac{\bomega_j}{\rho}\right]}^2 \mathrm{+} \frac{\lrnorm{\y \mathrm{-} \y_t^{(n)} \mathrm{+} \eta_t\g_t^{(n)}}^2}{\rho\eta_t\gamma} \\ \nonumber
=&\scriptsize \lrincir{\rho\eta_t\gamma\bLambda^\top\bLambda\mathrm{+}\I}^{-1}\left[\rho\eta_t\gamma\bLambda^\top\left [\r_{j+1}\mathrm{+}\bLambda\y_t^{(n)}\mathrm{+}\frac{\bomega_j}{\rho}\right ]\mathrm{+}\y_t^{(n)}\mathrm{-}\eta_t\g_t^{(n)}\right].
\end{align}  According to the eigen-value decomposition, $\bLambda^\top\bLambda$ can be represented by $\bLambda^\top\bLambda = \P\bSigma\P^{-1}$, where $\bSigma := \diag\lrincir{\lambda_1, \lambda_2, ..., \lambda_d}$, and $\lambda_i$ with $i\in\{1,2, ..., d\}$ are eigen-values of $\bLambda^\top\bLambda$. We have 
\begin{align}
\nonumber
\lrincir{\rho\eta_t\gamma\bLambda^\top\bLambda + \I}^{-1} = \P\lrincir{\rho\eta_t\gamma\bSigma + \I}^{-1}\P^{-1}.
\end{align} Therefore, $\y_{j+1}$ is updated by the following rule:
\begin{align}
\label{equa_update_y_admm}
&\y_{j+1}  \\ \nonumber
= & \scriptsize \P\lrincir{\rho\eta_t\gamma\bSigma \mathrm{+} \I}^{\mathrm{-}1}\P^{\mathrm{-}1} \left[\rho\eta_t\gamma\bLambda^\top\left [\r_{j+1}\mathrm{+}\bLambda\y_t^{(n)}\mathrm{+}\frac{\bomega_j}{\rho}\right ]\mathrm{+}\y_t^{(n)}\mathrm{-}\eta_t\g_t^{(n)}\right].
\end{align} 

\textbf{Update of $\bomega$.} Given $\r_{j+1}$ and $\y_{j+1}$, $\bomega_{j+1}$ is updated by the following rule:
\begin{align}
\label{equa_update_omega_admm}
\bomega_{j+1} = \bomega_j + \rho\lrincir{\r_{j+1} - \bLambda\y_{j+1} + \bLambda\y_t^{(n)}}.
\end{align}
In summary, algorithmic details are illustrated in Algorithm \ref{algo_efficient_update_nabla}.

\section{Computation Efficient Update of Model}
\label{sect_computation_efficient}
In the section, we find optimum of $\x$ and $\Z$ by performing alternative optimization, iteratively.  $\Z$ is optimized by using ADMM, which significantly reduces the computational cost.

\subsection{Efficient update of $\x$} 
When the server collects $\nabla_{t+1}^{(n)}$ from client, the sharing component $\x$ is updated by performing as follows:
\begin{align}
\nonumber
\x_{t+1} = \argmin_{\substack{\x\in\RR^{d_1}}}  \lrangle{\frac{1}{N}\sum_{n=1}^N\M^\top\nabla_{t+1}^{(n)}, \x} + \frac{1}{2\eta_t}\lrnorm{\x - \x_t}^2.
\end{align} Since it is an unconstrained optimization problem, it equals to the following update rule:
\begin{align}
\nonumber
\x_{t+1} = \x_t - \eta_t \lrincir{\frac{1}{N}\sum_{n=1}^N\M^\top\nabla_{t+1}^{(n)}}.
\end{align} That is, the sharing component $\x$ can be updated by  multiplication of matrices, which leads to low computational cost.

\subsection{Efficient update of $\Z$} 
Denote 
\begin{align}
\nonumber
h(\Z) :=  \frac{\1_d^\top\lrincir{\lrincir{\N^\top\nabla_{t+1}} \odot \Z}\1_N}{N} + \frac{1}{2\eta_t}\lrnorm{\Z - \Z_t}_F^2.
\end{align} The update of $\Z$ can be formulated by the following problem:
\begin{align}
\nonumber
\min_{\substack{\Z\in\RR^{d_2\times N}, \W\in\RR^{d_2\times M}}} H(\Z, \W) := h(\Z) + \lambda \lrnorm{\W}_{1,p},
\end{align} subject to:
\begin{align}
\nonumber
\Z\Q - \W = \0.
\end{align} Denote the augmented Lagrangian multiplier of $H(\Z, \W)$ by $L(\Z, \W, \bOmega)$, and we have 
\begin{align}
\nonumber
& L(\Z, \W, \bOmega) \\ \nonumber
:= & h(\Z) \mathrm{+} \lambda \lrnorm{\W}_{1,p} \mathrm{+} \1_{d_2}^\top(\bOmega \odot (\Z\Q\mathrm{-}\W))\1_M \mathrm{+} \frac{\rho}{2}\lrnorm{\W \mathrm{-} \Z\Q}_F^2.
\end{align} Then, $\Z$ is optimized by using ADMM by performing update of $\Z$, $\W$, and $\bOmega$, iteratively.

\begin{algorithm}
    \caption{Computation efficient update of $\Z$.}
    \label{algo_efficient_update_Z}
    \begin{algorithmic}[1]
        \Require The number of total iterations $K$, a positive $\rho$, and the initial model $\Z_t$.
           \For {$k=0,1,2, ..., K-1$}
            \State Update $\Z_{k+1}$ by performing Eq. \ref{equa_update_Z_admm}.
            \State Update the $m$-th column of $\W_{k+1}$, that is $[\W_{k+1}]_{:,m}$ by performing Eq. \ref{equa_update_W_admm}. 
           \State Update $\bOmega_{k+1}$ by performing Eq. \ref{equa_update_Omega_admm}.
            \State \EndFor
        \Return $\Z_K$.
      \end{algorithmic}
\end{algorithm}

\textbf{Update of $\Z$.} Given $\W_k$ and $\bOmega_k$, $\Z_{k+1}$ is obtained by performing the following problem:
\begin{align}
\nonumber
& \Z_{k+1}  = \argmin_{\Z\in\RR^{d_2\times N}} L(\Z, \W_k, \bOmega_k) & \\ \nonumber
= & \argmin_{\Z\in\RR^{d_2\times N}} h(\Z) + \1_{d_2}^\top(\bOmega_k \odot  (\Z\Q))\1_M + \frac{\rho}{2}\lrnorm{\W_k - \Z\Q}_F^2.
\end{align} Since it is unconstrained optimization problem, we can obtain $\Z_{k+1}$ as follows:
\begin{align}
\nonumber
\frac{\N^\top\nabla_{t+1}}{N} \mathrm{+} \frac{\Z_{k+1} \mathrm{-} \Z_t}{\eta_t} \mathrm{+}\bOmega_k\Q^\top \mathrm{+} \rho (\Z_{k+1}\Q \mathrm{-} \W_k)\Q^\top \mathrm{=} \0.
\end{align} That is, we have 
\begin{align}
\label{equa_update_Z_admm}
& \Z_{k+1} \\ \nonumber
= & \small \left [\eta_t\left[\rho\W_k\Q^\top \mathrm{-} \bOmega_k\Q^\top\mathrm{-}\frac{\N^\top\nabla_{t+1}}{N}\right]\mathrm{+}\Z_t \right ]\lrincir{\I_N\mathrm{+}\eta_t\rho\Q\Q^\top}^{\mathrm{-1}}.
\end{align}

\textbf{Update of $\W$.} Given $\Z_{k+1}$ and $\bOmega_k$, $\W_{k+1}$ is obtained by performing the following problem:
\begin{align}
\nonumber
& \W_{k+1} = \argmin_{\W\in\RR^{d_2 \times M}} L(\Z_{k+1}, \W, \bOmega_k) \\ \nonumber
= & \argmin_{\W\in\RR^{d_2 \times M}} \lambda \lrnorm{\W}_{1,p} \mathrm{-} \1_{d_2}^\top(\bOmega_k \odot \W)\1_M \mathrm{+} \frac{\rho\lrnorm{\W \mathrm{-} \Z_{k+1}\Q}_F^2}{2} \\ \nonumber
= & \argmin_{\W\in\RR^{d_2 \times M}} \lambda \lrnorm{\W}_{1,p} + \frac{\rho}{2}\lrnorm{\W - \lrincir{\Z_{k+1}\Q + \frac{1}{\rho}\bOmega_k}}_F^2 \\ \nonumber
= & \Prox_{\frac{\rho}{\lambda}, \lrnorm{\cdot}_{1,p}}\lrincir{\Z_{k+1}\Q + \frac{1}{\rho}\bOmega_k}.
\end{align}  Recall that $\lrnorm{\cdot}_{1,p}$ is the sum of norms, its proximal operator has a closed form \cite{proximal:Parikh}. Specifically, for the $m$-th column with $m\in\{1,2, ..., M\}$ of $\W_{k+1}$ is obtained by performing:
\begin{align}
\nonumber
& [\W_{k+1}]_{:,m} = \left [ \Prox_{\frac{\rho}{\lambda}, \lrnorm{\cdot}_{1,p}}\lrincir{\Z_{k+1}\Q + \frac{1}{\rho}\bOmega_k} \right ]_{:,m} \\ \label{equa_update_W_admm}
=& \left[1\mathrm{-} \frac{\lambda}{\lrnorm{\rho \Z_{k+1}\Q_{:, m} \mathrm{+} [\bOmega_k]{:, m}}_q}\right]_\mathrm{+} \lrincir{\Z_{k+1}\Q_{:, m} \mathrm{+} \frac{[\bOmega_k]_{:, m}}{\rho}},
\end{align} where $[\A]_{:,m}$ represents the $m$-th column of $\A$, and $\lrnorm{\cdot}_q$ is the dual norm of $\lrnorm{\cdot}_p$ such that $\frac{1}{p}+\frac{1}{q} = 1$.

\textbf{Update of $\bOmega$.} Given $\Z_{k+1}$ and $\W_{k+1}$, $\bOmega_{k+1}$ is obtained by performing the following rule:
\begin{align}
\label{equa_update_Omega_admm}
\bOmega_{k+1}  = \bOmega_k + \rho \lrincir{ \Z_{k+1}\Q - \W_{k+1}}.
\end{align} Algorithmic details are shown in Algorithm \ref{algo_efficient_update_Z}.

In summary, the federated model with personalized and sharing compenents is optimized by performing update of $\x$ and $\Z$ iteratively. Algorithmic details are illustrated in Algorithm \ref{algo_computaion_efficient_server}.

\begin{algorithm}
    \caption{Computation efficient training of personalized models at the server.}
    \label{algo_computaion_efficient_server}
    \begin{algorithmic}[1]
        \Require The number of total iterations $T$, and the initial model $\x_1$, and $\z_1^{(n)}$ with $n\in\{1,2,\cdots, N\}$.
        \State Deliver the model $\y_1^{(n)} = \M\x_1 + \N\z_1^{(n)}$ to all client $n$ with $n\in\{1,2, ..., N\}$.
        \For {$t=1,2, ..., T$}
            \For {$i=0,1,2,\cdots, I-1$}
                 \State Collect update of local model $\nabla_{i} = \left [ \nabla_t^{(1)}, \nabla_t^{(2)}, ..., \nabla_t^{(N)} \right ]$ from all client $n$ with $n\in\{1,2, ..., N\}$.
            	\State Update the global model $\x_{t+1}$ by performing:
            $$ \x_{i+1} = \x_i - \eta_i \lrincir{\frac{1}{N}\sum_{n=1}^N\M^\top\nabla_i^{(n)}}. $$
	         \State Deliver the model $\y_{i+1}^{(n)} = \M\x_{i+1} + \N\z_t^{(n)}$ to every client.
            \EndFor
           \For {$j=0,1,2,\cdots, J-1$}
           \State Collect update of local model $\nabla_j = \left [\nabla_j^{(1)}, \nabla_j^{(2)}, ..., \nabla_j^{(N)} \right ]$ from all client $n$ with $n\in\{1,2, ..., N\}$.
           \State Update the personalized model $\Z_{j+1}$ according to Algorithm \ref{algo_efficient_update_Z}.
           \State Deliver the parameter $\y_{j+1}^{(n)} = \M\x_I + \N\z_j^{(n)}$ to every client.
           \EndFor
        \EndFor
        \Return $\x_{T+1}^{(n)} = \M\x_I + \N\z_J^{(n)}$ with $n\in\{1,2, ..., N\}$.
      \end{algorithmic}
\end{algorithm}

% Preview source code for paragraph 0
\begin{table}[!h]
\centering
\caption{Summary of experimental settings.}
\begin{tabular}{c|c|c|c|c}
\hline 
Datasets & Models     & Tasks            & Data types    & Metrics  \tabularnewline
\hline 
\hline 
Luna16   & D-Net     & classification   & CT images     & test accuracy \tabularnewline
\hline 
BraTS2017 & U-Net & segmentation & MRI images     &  IoU \tabularnewline
\hline 
CHD        & LR       & classification     & structured data & test accuracy \tabularnewline
\hline 
Diabetes  & LR       & classification    & structured data & test accuracy  \tabularnewline
\hline 
Covid19  & LR        & classification    & structured data & test accuracy \tabularnewline
\hline
\end{tabular}
\label{table_experimental_settings}
\end{table}

% Preview source code for paragraph 0
\begin{table}[!h]
\centering
\caption{Evaluate accuracy for \textit{luna} by varying $\delta$.}
\begin{tabular}{c|c|c|c|c}
\hline 
Algo. & $\delta=1$ &  $\delta=2$ &  $\delta=4$ &  $\delta=7$ \tabularnewline
\hline 
\hline 
Ditto & 77.42$\pm$0.14 & 73.55$\pm$0.64 & 70.54$\pm$2.95 & 69.25$\pm$12.17 \tabularnewline
\hline 
FedAMP & 72.92$\pm$0.29 & 71.17$\pm$0.51 & 73.74$\pm$0.91 & 80.50$\pm$1.09 \tabularnewline
\hline 
FedAvg & 80.50$\pm$0 & 81.12$\pm$0 & 77.27$\pm$0 & 63.92$\pm$8.08 \tabularnewline
\hline 
L2GD & 73.33$\pm$0.14 & 69.98$\pm$0.15 & 70.20$\pm$0.44 & 81.08$\pm$1.04 \tabularnewline
\hline 
FedPer & 62.33$\pm$0.14 & 66.50$\pm$0.15 & 77.27$\pm$0 & 83.92$\pm$0.14 \tabularnewline
\hline 
FedProx & 82.50$\pm$1.50 & 77.89$\pm$1.26 & 72.47$\pm$1.54 & 75.75$\pm$1.56 \tabularnewline
\hline 
FedRoD & 79.50$\pm$0.25 & 79.25$\pm$0.53 & 83.42$\pm$1.17 & 86.00$\pm$1.09 \tabularnewline
\hline 
FPFC & 73.50$\pm$0.25 & 74.91$\pm$0.15 & 78.96$\pm$0.15 & 85.00$\pm$0 \tabularnewline
\hline 
IFCA & 80.17$\pm$0.29 & 80.61$\pm$0 & 73.74$\pm$0 & 63.33$\pm$0.52 \tabularnewline
\hline 
pFedMe & 71.08$\pm$0.14 & 67.43$\pm$0.90 & 71.21$\pm$4.59 & 82.00$\pm$0.90 \tabularnewline
\hline 
SuPerFed & 61.17$\pm$0.38 & 65.31$\pm$0 & 76.77$\pm$0 & 84.08$\pm$0.14 \tabularnewline
\hline 
FedRep    & 77.50$\pm$0.50 & 79.68$\pm$0.39 & 79.97$\pm$0.89 & 83.75$\pm$0 \tabularnewline
\hline 
\bf{pFedNet} & \bf{86.25$\pm$0} & \bf{82.74$\pm$0.15} & \bf{86.45$\pm$0.29} & \bf{86.42$\pm$0.14} \tabularnewline
\hline 
\bf{rank} & \bf{top 1} & \bf{top 1} & \bf{top 1} & \bf{top 1} \tabularnewline
\hline 
\end{tabular}
\label{table_classification_lung_nodules}
\end{table}

% Preview source code for paragraph 0
\begin{table}[!h]
\centering
\caption{Evaluate \textit{IoU} on \textit{BraTS2017}.}
\begin{tabular}{c|c|c|c|c}
\hline 
Algo. & lack \#$0$ &  lack \#$1$ &  lack \#$2$ &  lack \#$3$ \tabularnewline
\hline 
\hline 
Ditto & 69.80$\pm$0.24       & 69.68$\pm$0.25 & 67.92$\pm$0.42 & 69.07$\pm$0.38 \tabularnewline
\hline 
FedAMP & 67.08$\pm$0.12 & 65.97$\pm$0.29 & 65.42$\pm$0.10 & 67.33$\pm$0.61 \tabularnewline
\hline 
FedAvg & 70.13$\pm$0.29  & \bf{70.84$\pm$0.18} & 67.82$\pm$0.09 & 67.29$\pm$0.35 \tabularnewline
\hline 
L2GD & 66.40$\pm$0.21     & 65.77$\pm$0.52       & 65.28$\pm$0.25 & 67.75$\pm$0.65 \tabularnewline
\hline 
FedPer & 61.05$\pm$0.90  & 63.53$\pm$0.78 & 62.22$\pm$0.55 & 65.55$\pm$0.17 \tabularnewline
\hline 
FedProx & 50.65$\pm$0.02 & 52.49$\pm$0         & 51.42$\pm$0.03 & 56.77$\pm$0.25 \tabularnewline
\hline 
FedRoD & 68.19$\pm$0.31 & 68.78$\pm$0.23 & 69.04$\pm$0.04 & 69.79$\pm$0.23 \tabularnewline
\hline 
FPFC & 61.98$\pm$0.27     & 61.64$\pm$0.36 & 63.90$\pm$0.26 & 65.81$\pm$0.45 \tabularnewline
\hline 
IFCA & 68.95$\pm$0.11      & 69.72$\pm$0.09 & 67.58$\pm$0.14 & 66.09$\pm$0.09 \tabularnewline
\hline 
pFedMe & 68.78$\pm$0.11 & 67.38$\pm$0.45 & 66.98$\pm$0.22 & 69.14$\pm$0.49 \tabularnewline
\hline 
SuPerFed & 62.67$\pm$0.35 & 63.25$\pm$0.32 & 62.82$\pm$0.40   & 66.78$\pm$0.19 \tabularnewline
\hline 
FedRep & 70.39$\pm$0.08  & 66.80$\pm$0.43     & 70.42$\pm$0.25 & 69.62$\pm$0.75 \tabularnewline
\hline 
\bf{pFedNet} & \bf{70.73$\pm$0.26} & 70.40$\pm$0.04 & \bf{71.56$\pm$0.21} & \bf{70.63$\pm$0.33} \tabularnewline
\hline 
\bf{rank} & \bf{top 1} & \bf{top 2} & \bf{top 1} & \bf{top 1} \tabularnewline
\hline 
\end{tabular}
\label{table_brats_iou}
\end{table}

% Preview source code for paragraph 0
\begin{table}[!h]
\centering
\caption{Evaluate \textit{label IoU} on \textit{BraTS2017}.}
\begin{tabular}{c|c|c|c|c}
\hline 
Algo. & lack \#$0$ &  lack \#$1$ &  lack \#$2$ &  lack \#$3$ \tabularnewline
\hline 
\hline 
Ditto & 66.25$\pm$0.35       & 65.56$\pm$0.10 & 64.48$\pm$0.76 & 65.73$\pm$0.11 \tabularnewline
\hline 
FedAMP & 63.45$\pm$0.30 & 62.23$\pm$0.26 & 61.71$\pm$0.35 & 63.18$\pm$0.24 \tabularnewline
\hline 
FedAvg & 67.68$\pm$0.24  & \bf{67.89$\pm$0.13} & 64.67$\pm$0.23 & 63.72$\pm$0.02 \tabularnewline
\hline 
L2GD & 63.20$\pm$0.23     & 62.27$\pm$0.88 & 61.95$\pm$0.59 & 63.89$\pm$0.32 \tabularnewline
\hline 
FedPer & 56.47$\pm$0.89  & 59.46$\pm$0.26 & 57.30$\pm$0.58 & 58.22$\pm$0.26 \tabularnewline
\hline 
FedProx & 39.16$\pm$0.03 & 42.16$\pm$0.14 & 38.24$\pm$0.16 & 46.07$\pm$1.01 \tabularnewline
\hline 
FedRoD & 64.91$\pm$0.32 & 65.82$\pm$0.10 & 65.46$\pm$0.23 & 66.21$\pm$0.10 \tabularnewline
\hline 
FPFC & 59.12$\pm$0.09     & 59.42$\pm$0.08 & 61.10$\pm$0.22 & 62.63$\pm$0.26 \tabularnewline
\hline 
IFCA & 67.13$\pm$0.03      & 67.19$\pm$0.05 & 64.49$\pm$0.33 & 63.44$\pm$0.07 \tabularnewline
\hline 
pFedMe & 65.16$\pm$0.23 & 64.29$\pm$0.22 & 63.56$\pm$0.47 & 65.05$\pm$0.21 \tabularnewline
\hline 
SuPerFed & 59.38$\pm$0.14 & 59.11$\pm$0.66 & 60.19$\pm$0.28 & 63.07$\pm$0.08 \tabularnewline
\hline 
FedRep & 68.05$\pm$0.22 & 65.10$\pm$0.21 & 68.03$\pm$0.21 & 67.32$\pm$0.54 \tabularnewline
\hline 
\bf{pFedNet} & \bf{71.25$\pm$0.19} & 67.62$\pm$0.26 & \bf{69.43$\pm$0.10} & \bf{68.61$\pm$0.38} \tabularnewline
\hline 
\bf{rank} & \bf{top 1} & \bf{top 2} & \bf{top 1} & \bf{top 1} \tabularnewline
\hline 
\end{tabular}
\label{table_brats_label_iou}
\end{table}

\section{Empirical Studies}
\label{sect_empirical_studies}
This section presents performance of the proposed method on model effectiveness, communication efficiency and so on by conducting extensive empirical studies.

\subsection{Experimental Settings}
\label{subsect_experimental_settings}
\textbf{Datasets and tasks.} We conduct classification and segmentation tasks on $2$ public medical datasets: \textit{Luna16}, \textit{BraTS2017}, and $3$ private medical datasets collecting from multiple medical centers of hosiptal: \textit{CHD}, \textit{Diabetes}, and \textit{Covid19}. Those datasets own different modalities. Specifically, \textit{Luna16}\footnote{\url{https://luna16.grand-challenge.org/Data/}} and \textit{BraTS2017}\footnote{\url{https://www.med.upenn.edu/sbia/brats2017/data.html}} are lung CT and brain tumor MRI images, respectively.  \textit{CHD}, \textit{Diabetes}, and \textit{Covid19} are structural medical data. Details of datasets are presented as follows.

\begin{itemize}
\item \textbf{Luna16}.  It is a public dataset to evaluate the algorithmic performance of lung nodule detection. The dataset consists of $888$ patients' CT scans, and every scan is sliced into $64$ pieces. More than $551,065$ candidates of lung nodules are recognized by tools automatically, while only $1186$ true nodules are identified by real doctors. In the experiment, we extract every candidate of lung nodules by using a $32\times32$ patch. 
\item \textbf{BraTS2017}. It is a public dataset, and is usually used to segmentation of glioma sub-regions of brain. The dataset consists of $484$ patients' MRI scans, and every scan owns $4$ channels. In the experiment, we extract every candidate of brain tumor by using a $64\times64$ patch.
\item \textbf{CHD}. The dataset is built from the first medical center of the PLA general hospital of China. It is used to conduct prediction of bleeding risk in elderly patients with coronary heart disease combined with intestinal malignant tumors. The dataset consists of $716$ patients' medical records, and every record owns $58$ features. Logistic regression model is used to predict whether the event of bleeding appears.
\item \textbf{Diabetes}. The dataset is built from the first medical center of the PLA general hospital of China, and is used to conduct risk prediction of type $2$ diabetes retinopathy. The dataset consists of $31,476$ patients' medical records, and every record owns $63$ features. Logistic regression model is used to predict whether the event of diabetes retinopathy.  
\item \textbf{Covid19}. The dataset is built from three medical centers (the first/fifth/sixth medical center) of the PLA general hospital of China, and is used to predict event of Covid-19 infection. The dataset consists of $2402$ patients' medical records, and every record owns $77$ features. Logistic regression model is used to predict whether the infection event.  
\end{itemize}

Additionally, we conduct $3$ medical analysis tasks, including lung nodule classification, brain tumor segmentation, and clinical risk prediction. 
\begin{itemize}
\item \textbf{Lung nodule classification}. \textit{Dense net} \cite{huang2017densely} (D-Net) model is chosen to detect real lung nodules from all candidates. We choose parameters of the fully connecting layer as the personalized component, and others as the sharing component. 
\item \textbf{Brain tumor segmentation}. \textit{U-net} \cite{unetmiccai2015} (U-Net) model is picked to conduct segmentation of brain tumors. Parameters of down-sampling layers are chosen as the sharing component, and up-sampling layers' parameters are chosen as the personalized component.
\item \textbf{Clinical risk prediction}. We use \textit{Logistic Regression} (LR) model \cite{shalev2014} to predict whether clinical risks (bleeding, and infection etc.) appears. All features are chosen as the personalized component.
\end{itemize}
In the experiment, we first fill all missing values by using zeros, and normalize values between $-1$ and $1$. Experimental settings are shown in Table \ref{table_experimental_settings} briefly.

% Preview source code for paragraph 0
\begin{table}[!h]
\centering
\caption{Evaluate accuracy (\%) for \textit{CHD} by varying $\delta$.}
\begin{tabular}{c|c|c|c|c}
\hline 
Algo. & $\delta=1$ &  $\delta=2$ &  $\delta=3$ &  $\delta=4$ \tabularnewline
\hline 
\hline 
Ditto        & 92.86$\pm$0     & 69.04$\pm$1.09     & 67.37$\pm$4.06 & 67.13$\pm$4.72 \tabularnewline
\hline 
FedAMP & 87.62$\pm$2.06 & 70.00$\pm$0         & 69.72$\pm$0      & 69.50$\pm$0 \tabularnewline
\hline 
FedAvg  & 92.62$\pm$0.42 & 69.76$\pm$0.42     & 69.95$\pm$0.40 & 69.50$\pm$0 \tabularnewline
\hline 
L2GD     & 91.19$\pm$2.89 & 70.00$\pm$0          & 69.72$\pm$0      & 69.50$\pm$0 \tabularnewline
\hline 
FedPer   & 92.62$\pm$0.42 & 70.47$\pm$1.49     & 70.42$\pm$1.21 & 69.50$\pm$0 \tabularnewline
\hline 
FedProx & 92.86$\pm$0      & 70.00$\pm$0          & 69.72$\pm$0       & 69.50$\pm$0 \tabularnewline
\hline 
APFL      & 92.86$\pm$0      & 69.76$\pm$1.09     & 69.96$\pm$0.41  & 69.50$\pm$0 \tabularnewline
\hline 
FPFC     & 92.86$\pm$0       &\bf{71.67$\pm$2.89}& 69.72$\pm$0      & 65.48$\pm$6.96 \tabularnewline
\hline 
IFCA      & 91.19$\pm$2.88  & 70.00$\pm$0.72      & 69.72$\pm$0      & 69.75$\pm$0.43 \tabularnewline
\hline 
pFedMe & 91.19$\pm$2.89  & 70.00$\pm$0           & 69.72$\pm$0      & 69.50$\pm$0 \tabularnewline
\hline 
SuPerFed & 74.76$\pm$10.69 & 63.09$\pm$5.27  & 69.48$\pm$0     & 69.03$\pm$1.47 \tabularnewline
\hline 
FedRep & 79.52$\pm$11.57 & 69.29$\pm$2.86 & 69.95$\pm$1.77 & 68.79$\pm$1.23 \tabularnewline
\hline 
\bf{pFedNet} & \bf{92.86$\pm$0} & 70.00$\pm$0 & \bf{71.36$\pm$2.85} & \bf{70.45$\pm$0} \tabularnewline
\hline 
\bf{rank} & \bf{top 1} & \bf{top 3} & \bf{top 1} & \bf{top 1} \tabularnewline
\hline 
\end{tabular}
\label{table_acc_chd}
\end{table}

% Preview source code for paragraph 0
\begin{table}[!h]
\centering
\caption{Evaluate accuracy (\%) for \textit{Diabetes} by varying $\delta$.}
\begin{tabular}{c|c|c|c|c}
\hline 
Algo. & $\delta=1$ &  $\delta=2$ &  $\delta=3$ &  $\delta=4$ \tabularnewline
\hline 
\hline 
Ditto        & 90.42$\pm$2.84  & 66.43$\pm$1.99     & 68.07$\pm$0.41 & 67.43$\pm$1.64 \tabularnewline
\hline 
FedAMP & 93.63$\pm$0       & 67.76$\pm$0          & 67.74$\pm$0      & 68.72$\pm$0 \tabularnewline
\hline 
FedAvg  & 92.90$\pm$1.26   & 67.76$\pm$0          & 67.77$\pm$0.05 & 68.71$\pm$0 \tabularnewline
\hline 
L2GD     & 93.63$\pm$0        & 67.76$\pm$0          & 67.74$\pm$0      & 68.72$\pm$0 \tabularnewline
\hline 
FedPer   & 91.66$\pm$3.36  & 67.75$\pm$0.02     & 66.66$\pm$1.89 & 68.77$\pm$0.11 \tabularnewline
\hline 
FedProx & 93.63$\pm$0      & 67.76$\pm$0           & 67.74$\pm$0       & 68.71$\pm$0 \tabularnewline
\hline 
APFL      & 92.90$\pm$0      & 67.75$\pm$0           & 67.77$\pm$0.05  & 68.71$\pm$0 \tabularnewline
\hline 
FPFC     & 93.62$\pm$0.33  & 67.78$\pm$0.03     & 67.39$\pm$0.75   & 68.64$\pm$0.42 \tabularnewline
\hline 
IFCA      & 93.62$\pm$0.02  & 67.77$\pm$0.02      & 67.70$\pm$0.07   & 68.72$\pm$0 \tabularnewline
\hline 
pFedMe & 68.39$\pm$0       & 67.87$\pm$0           & 67.75$\pm$0      & \bf{69.05$\pm$0} \tabularnewline
\hline 
SuPerFed & 93.01$\pm$1.01 & 66.66$\pm$2.13    & 65.74$\pm$3.43     & 68.98$\pm$0.87 \tabularnewline
\hline 
FedRep & 91.26$\pm$3.23    & 67.36$\pm$0.68     & 66.44$\pm$1.22     & 66.30$\pm$2.46 \tabularnewline
\hline 
\bf{pFedNet} & \bf{93.63$\pm$0} & \bf{67.96$\pm$0.41} & \bf{68.29$\pm$0.68} & 68.72$\pm$0 \tabularnewline
\hline 
\bf{rank} & \bf{top 1} & \bf{top 1} & \bf{top 1} & \bf{top 3} \tabularnewline
\hline 
\end{tabular}
\label{table_acc_diabetes}
\end{table}

% Preview source code for paragraph 0
\begin{table}[!h]
\centering
\caption{Evaluate accuracy (\%) for \textit{Covid19} by varying $\delta$.}
\begin{tabular}{c|c|c|c|c}
\hline 
Algo. & $\delta=1$ &  $\delta=2$ &  $\delta=3$ &  $\delta=4$ \tabularnewline
\hline 
\hline 
Ditto        & 91.66$\pm$2.87  & 79.10$\pm$0.12     & 79.03$\pm$0.24 &  79.17$\pm$0 \tabularnewline
\hline 
FedAMP & 90.55$\pm$1.56   & 79.17$\pm$0          & 79.17$\pm$0      & 79.17$\pm$0 \tabularnewline
\hline 
FedAvg  & 95.35$\pm$0.84   & 78.95$\pm$0.54      & 78.96$\pm$0.56 & 77.01$\pm$4.10 \tabularnewline
\hline 
L2GD     & \bf{95.83$\pm$0}  & 79.17$\pm$0          & 79.31$\pm$0.12  & 79.17$\pm$0 \tabularnewline
\hline 
FedPer   & 90.49$\pm$4.77  & 78.61$\pm$0.49     & 76.46$\pm$4.88  & \bf{80.00$\pm$1.08} \tabularnewline
\hline 
FedProx & 95.83$\pm$0       & 79.17$\pm$0           & 79.17$\pm$0       & 79.17$\pm$0 \tabularnewline
\hline 
APFL      & 94.37$\pm$2.52  & 78.95$\pm$0.54      & 78.96$\pm$0.56  & 77.01$\pm$4.10 \tabularnewline
\hline 
FPFC     & 95.28$\pm$0.64  & 73.13$\pm$10.46     & 74.72$\pm$5.92   & 77.78$\pm$2.80 \tabularnewline
\hline 
IFCA      & 94.51$\pm$1.26  & 79.17$\pm$0            & 79.03$\pm$0.32   & 78.89$\pm$0.48 \tabularnewline
\hline 
pFedMe & 37.08$\pm$0       & 63.96$\pm$0           & 47.43$\pm$0.12    & 53.96$\pm$0 \tabularnewline
\hline 
SuPerFed & 92.28$\pm$1.64 & 79.10$\pm$0.12    & 79.10$\pm$0.12    & 79.17$\pm$0 \tabularnewline
\hline 
FedRep  &93.20$\pm$3.32 & 78.26$\pm$1.82       & 77.78$\pm$2.13     & 77.64$\pm$1.36 \tabularnewline
\hline 
\bf{pFedNet} & 95.56$\pm$0.24 & \bf{79.24$\pm$0.12} & \bf{79.40$\pm$0.40} & 79.17$\pm$0 \tabularnewline
\hline 
\bf{rank} & \bf{top 3} & \bf{top 1} & \bf{top 1} & \bf{top 2} \tabularnewline
\hline 
\end{tabular}
\label{table_acc_covid19}
\end{table}

\begin{figure}
\setlength{\abovecaptionskip}{0pt}
\setlength{\belowcaptionskip}{0pt}
\centering 
\includegraphics[width=0.99\columnwidth]{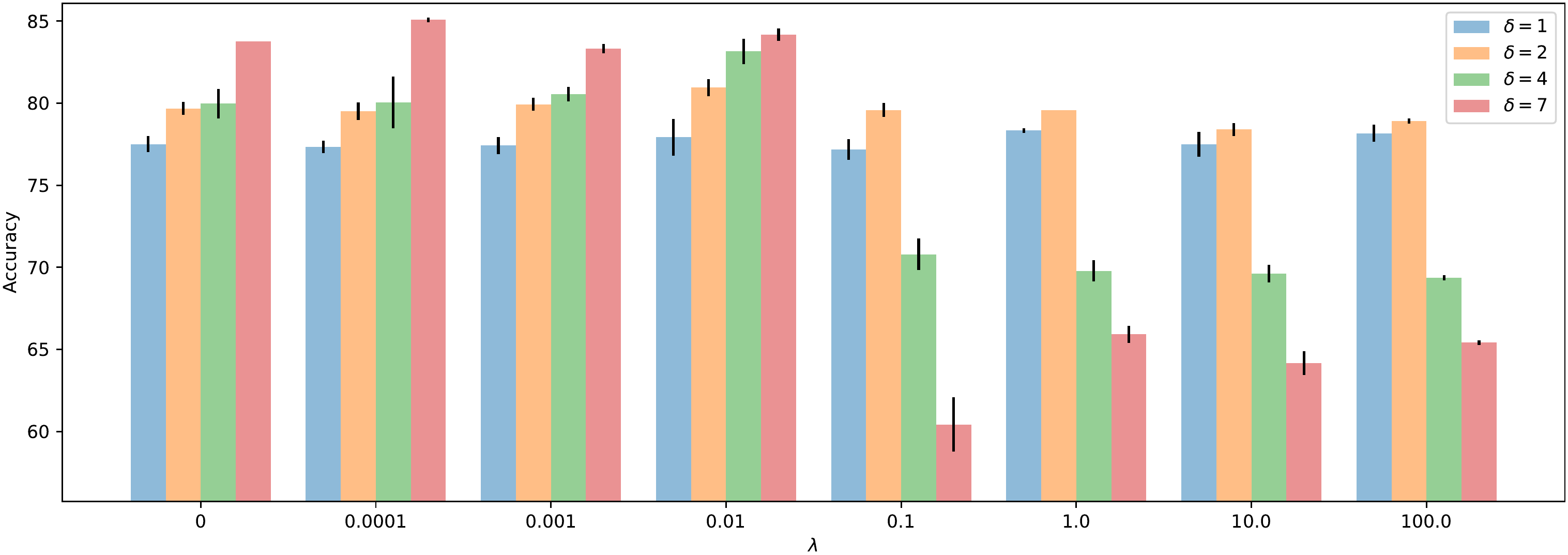}
\caption{Illustrative results of test accuracy w.r.t. $\lambda$.}
\label{figs_lambda_luna}
\end{figure}

\textbf{Methods and metrics.} The proposed method, that is \textit{pFedNet}, is evaluated by comparing $13$ existing methods. Those methods include \textit{Ditto} \cite{Li2020DittoFA}, \textit{FedAMP} \cite{huang2021personalized}, \textit{FedAvg} \cite{li2019convergence}, \textit{L2GD} \cite{hanzely2020federated}, \textit{FedPer} \cite{arivazhagan2019federated}, \textit{FedProx} \cite{li2020federated}, \textit{FedRoD} \cite{Chen2021OnBG}, \textit{APFL} \cite{deng2020adaptive}, \textit{FPFC} \cite{yu2022clustered}, \textit{IFCA} \cite{ghosh2022efficient}, \textit{pFedMe} \cite{dinh2020}, \textit{SuPerFed} \cite{hahn2022}, and \textit{FedRep} \cite{collins2021exploiting}.  \textit{FedAvg} and \textit{FedProx} are general optimization methods for federated learning, while others are recently proposed personalized federated learning methods. Additionally, the performance of all classification model is measured by \textit{test accuracy}, and the performance of the segmentation model is measured by \textit{Intersection over Union (IoU)}. These metrics are widely used in previous work \cite{chen2022personalized,wang2022personalizing,li2020federated,hahn2022}. The communication efficiency is measured by the model size.

\textbf{Federated setting.} In the experiment, there are $5$ clients and $1$ server. That is, the similarity network consists of $5$ nodes, and its edges are generated by using KNN with $k=3$. Every dataset is partitioned and allocated to all clients, $80\%$ of them is used to training model, and $20\%$ of them is used to conduct model test. Specifically, data federation for classification is built based on the setting of label unbalance, which is measured by $\delta := n_{\text{negative}}/n_{\text{positive}}$. Here, $n_{\text{negative}}$ and $n_{\text{positive}}$ represent the number of negative and positive labels, respectively. Data federation for segmentation is built based on setting of channel unbalance, which is measured by the id of the missing channel (e.g. \textit{lack \#0}, and \textit{lack \#1} etc). All methods are implemented by using PyTorch, and run on a machine with $24$ GB memory, $2$TB SSD, and Nvidia Tesla 3090. 

\begin{figure}
\setlength{\abovecaptionskip}{0pt}
\setlength{\belowcaptionskip}{0pt}
\centering 
\subfigure[Model size reduction, \textit{CER} v.s. \textit{STC}]{\includegraphics[width=0.49\columnwidth]{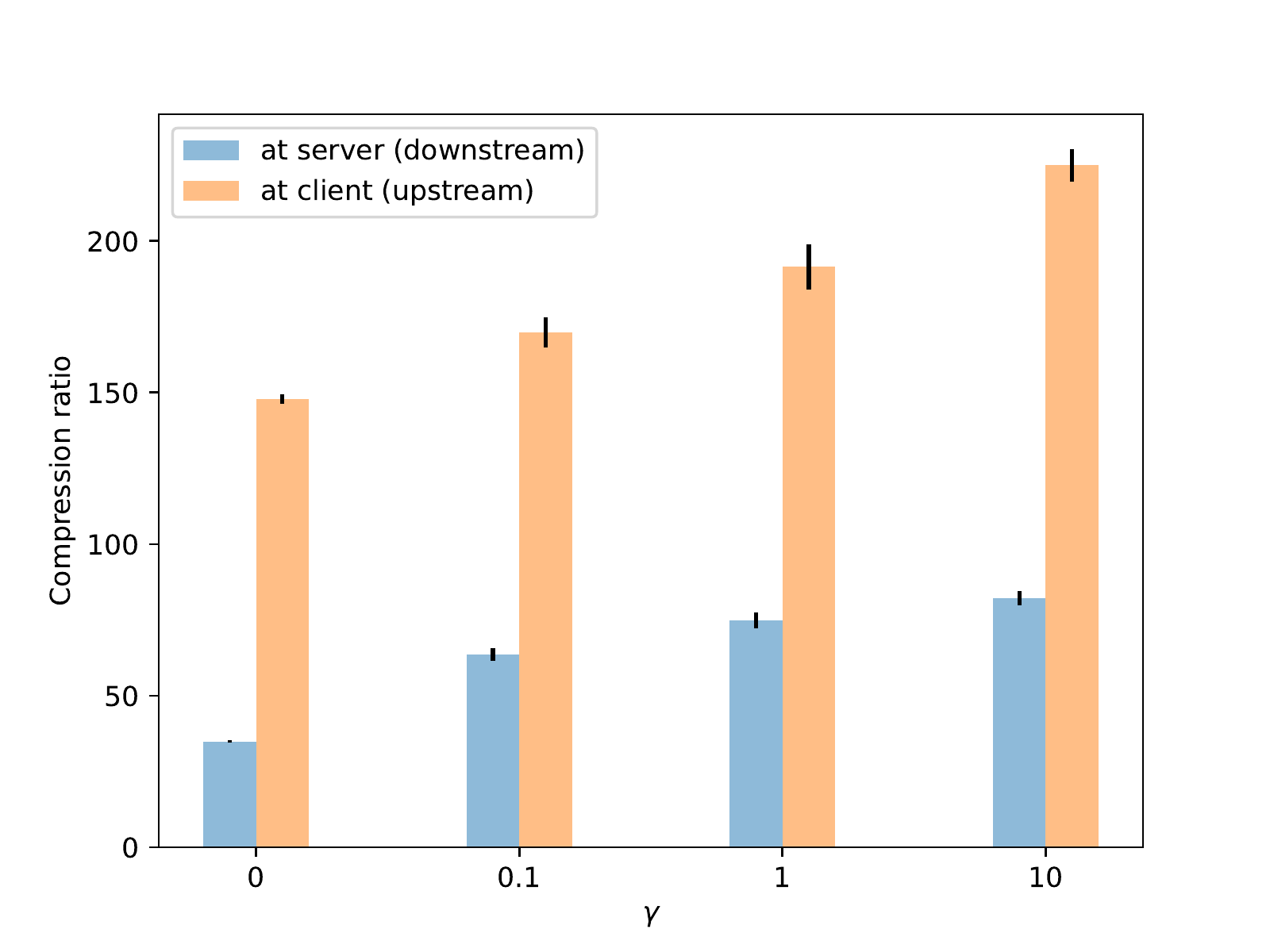}\label{figs_gamma_luna}}
\subfigure[Test accuracy with \textit{CER}, \textit{CER} v.s. \textit{STC}]{\includegraphics[width=0.49\columnwidth]{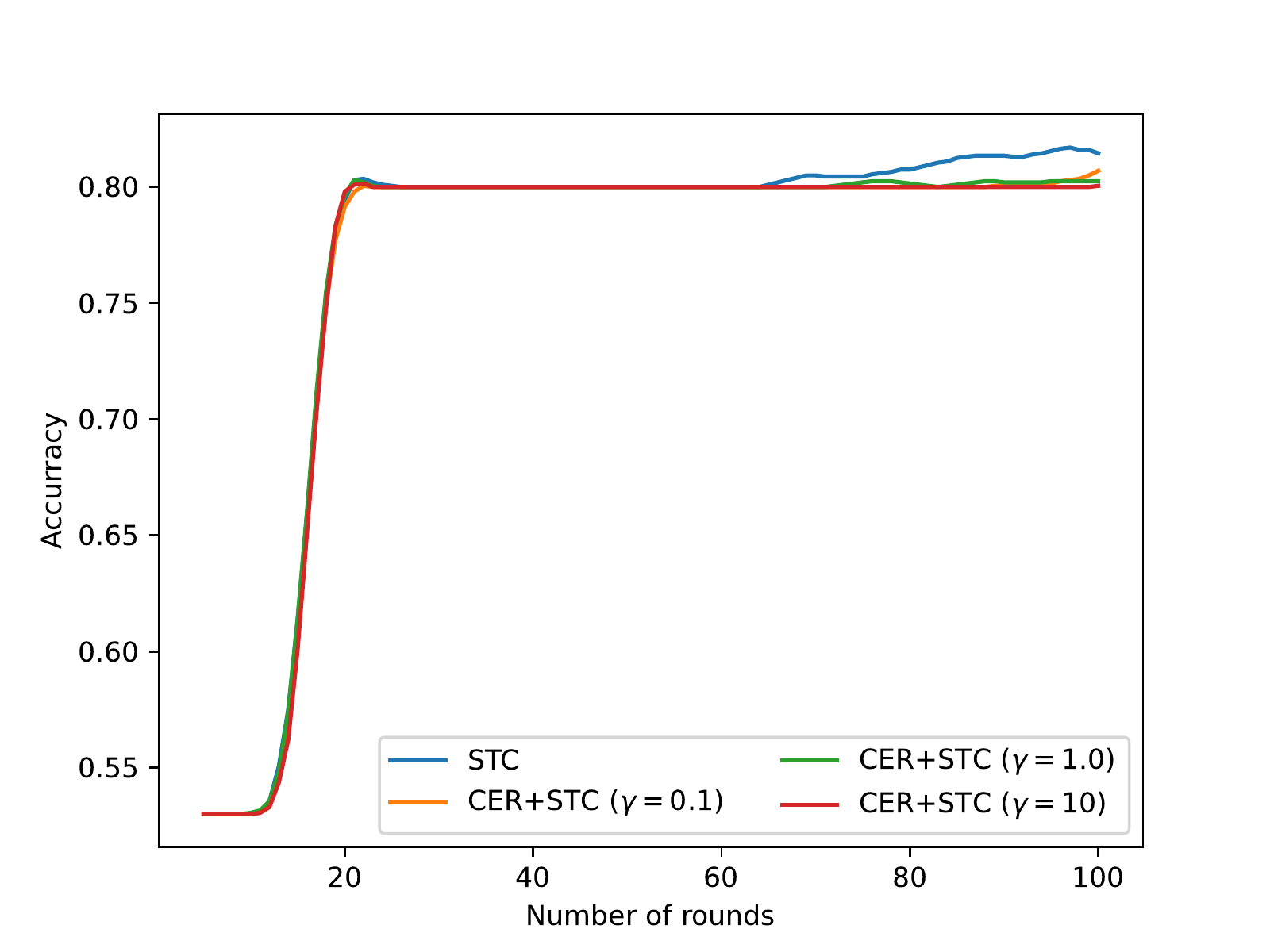}\label{figs_densenet_communicate_acc_line}}
\subfigure[Model size reduction at client, \textit{CER} v.s. \textit{STC}]{\includegraphics[width=0.49\columnwidth]{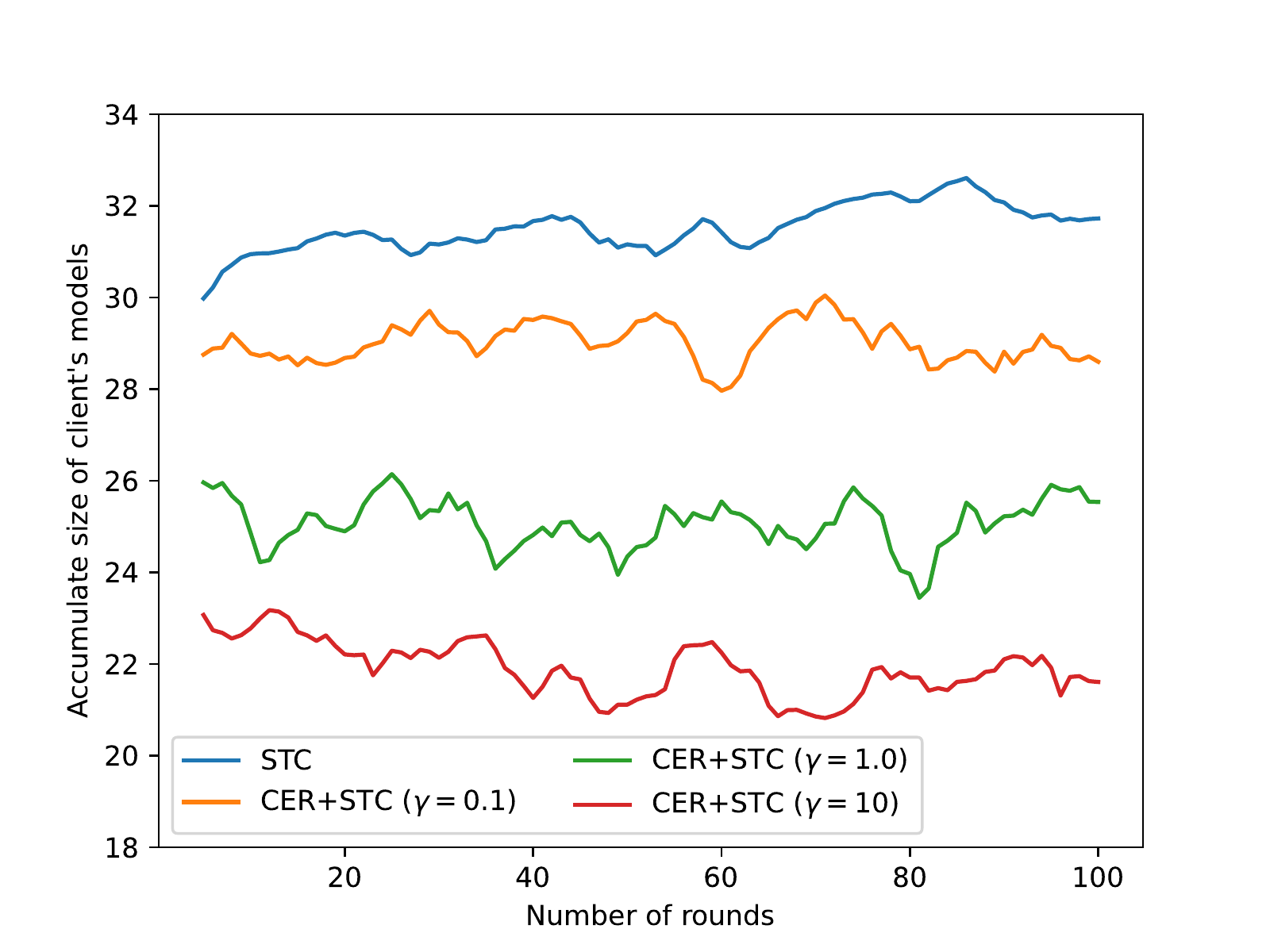}\label{figs_densenet_communicate_client_line}}
\subfigure[Statistics of model size reduction at client, \textit{CER} v.s. \textit{STC}]{\includegraphics[width=0.49\columnwidth]{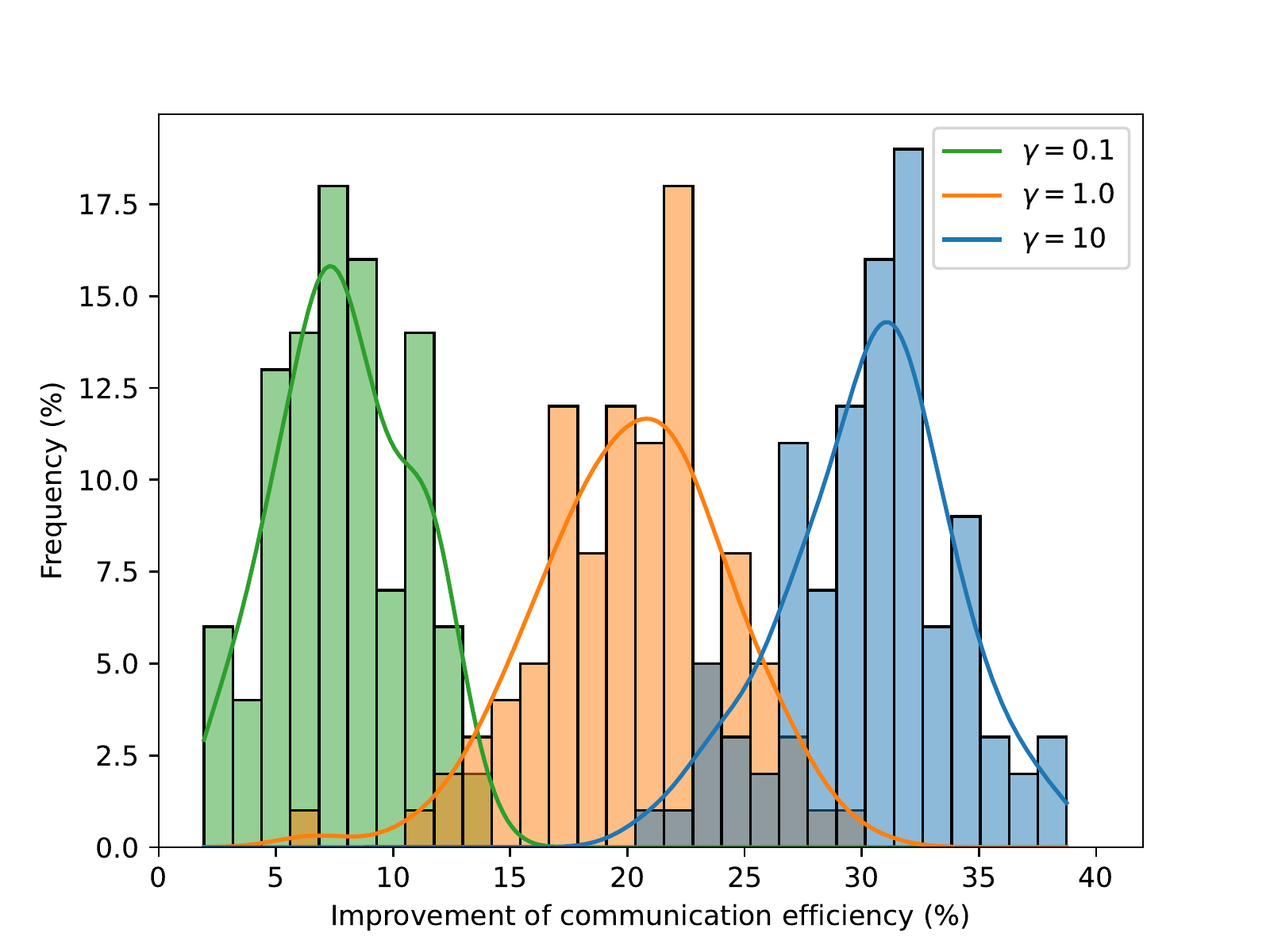}\label{figs_densenet_communicate_client}}
\subfigure[Model size reduction at server, \textit{CER} v.s. \textit{STC}]{\includegraphics[width=0.49\columnwidth]{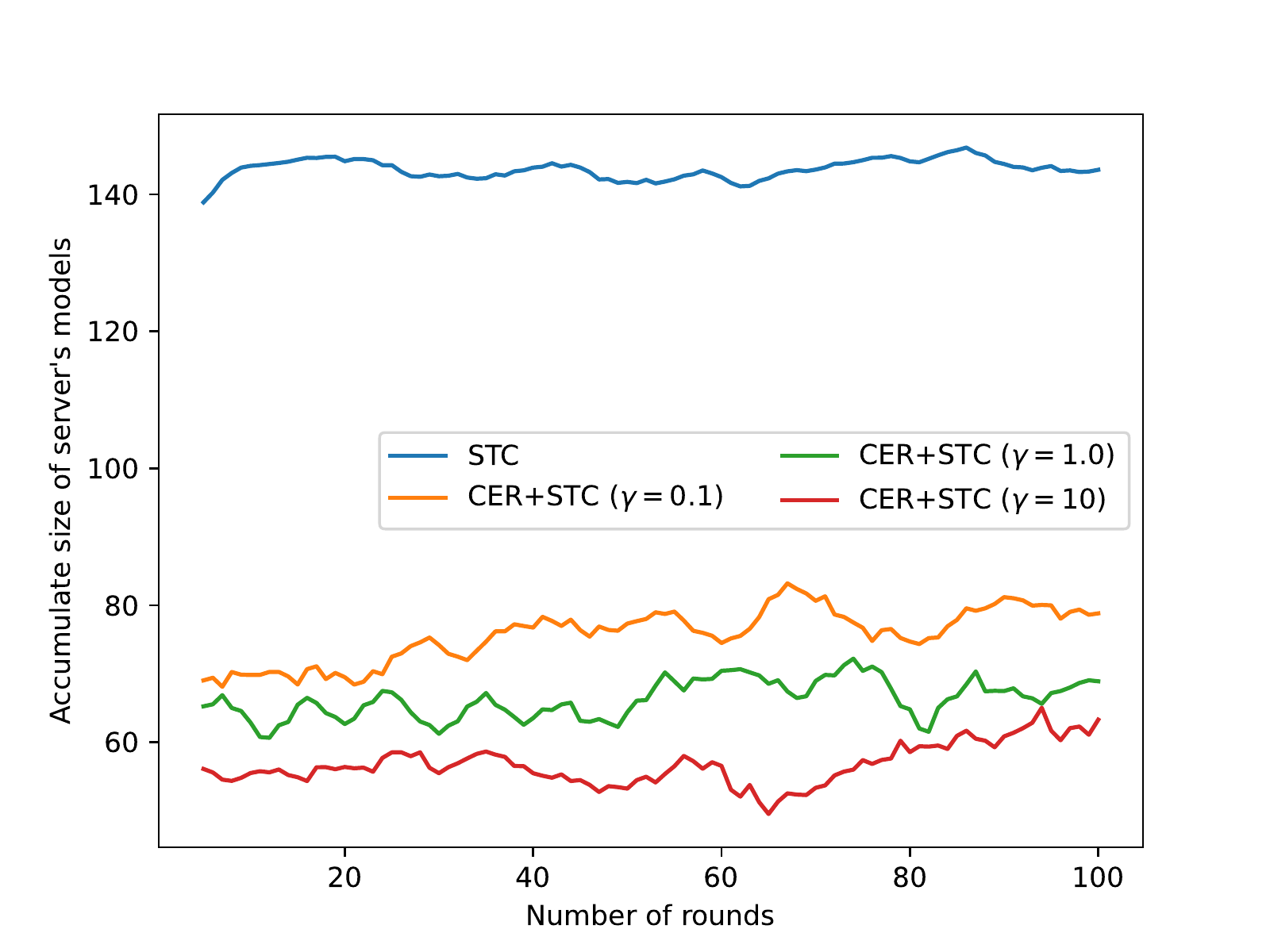}\label{figs_densenet_communicate_server_line}}
\subfigure[Statistics of model size reduction at server, \textit{CER} v.s. \textit{STC}]{\includegraphics[width=0.49\columnwidth]{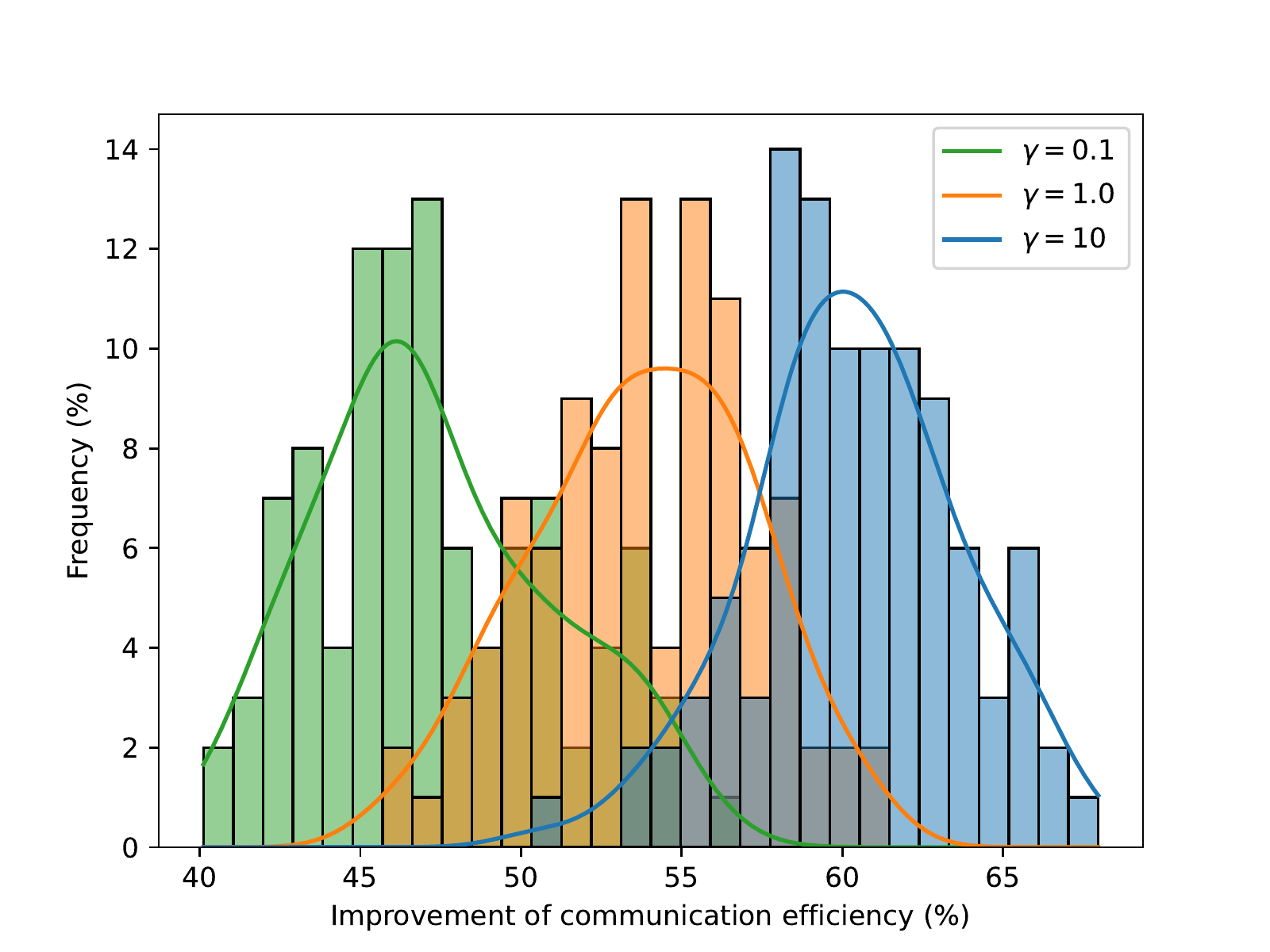}\label{figs_densenet_communicate_server}}
\caption{\textit{pFedNet} significantly reduces model size with \textit{CER}.}
\label{figure_luna_cer}
\end{figure}

\subsection{Numerical Results on Public Datasets}
\subsubsection{Classification of Lung Nodules}
First, we evaluate the model performance of methods, and find \textit{pFedNet} successfully beat other existing methods. We test personalized model at client, collect all local test accuracy, and then compute the average as the final test accuracy. As illustrated in Table \ref{table_classification_lung_nodules}, \textit{pFedNet} achieves the best performance, and enjoys more than $3\%$ gains of test accuracy higher than other methods in most case. Additionally, we vary $\lambda$ to generate different personalized models. As shown in Figure \ref{figs_lambda_luna}, a small $\lambda$ tends to yield a more personalized model, which could be adaptive to unbalance data, and obtains higher test accuracy. However, a tiny $\lambda$ with $\lambda < 10^{-3}$ may falsely view some noise of data as the personalized component, which leads to over-personalized model, and decrease the model performance. It seems that $\lambda = 0.01$ is a good choice since most of unbalanced data achieves best performance. Moreover, we find that test accuracy is sensitive to $\lambda$. The more unbalanced data, the more sensitive it is. Specifically, comparing with the unbalanced data with different settings of $\delta$, we find the test accuracy decreases much more significantly  with the increase of $\lambda$ for a larger $\delta$.

Second, the proposed method, namely \textit{CER}, successfully improves communication efficiency by reducing model size effectively. Figure \ref{figure_luna_cer} shows the superiority of communication efficiency. It is a good complement for existing methods, and can promote their performance effectively. We choose one of widely used model compression methods, that is \textit{STC} \cite{sattler2019robust}, to show benefits of the proposed method. As illustrated in Figure \ref{figs_gamma_luna}. \textit{STC} without \textit{CER}, that is $\gamma=0$, achieves up to $146\times$ compression ratio at client, and $34\times$ compression ratio at server, respectively. After equipping with \textit{CER} ($\gamma=0.1$), that is \textit{CER}+\textit{STC}, successfully achieves up to $172\times$ compression ratio at client and $61\times$ compression ratio at server. The advantage becomes more significant with the increase of $\gamma$. As we have claimed, the communication efficiency may be achieved with sacrifice of model performance. Figure \ref{figs_densenet_communicate_acc_line} demonstrates the test accuracy climbs up   fast, and the gap caused by \textit{CER} is insignificant. That is,  the superiority on the communication efficiency can be achieved without significant harm to the test accuracy.  Specifically, as shown in Figure \ref{figs_densenet_communicate_client_line}, \textit{CER} reduces size of client's model effectively, which becomes more and more significant with the increase of $\gamma$. Figure \ref{figs_densenet_communicate_client} shows that \textit{CER} can  improve the communication efficiency at client by reducing $7\%\sim32\%$ model size more than \textit{STC}. The benefit becomes more significant when delivering personalized model to every client. Figure \ref{figs_densenet_communicate_server_line} shows that \textit{CER} can promote the performance of \textit{STC} prominently at server, and obtains much more noticeable advantages on the communication efficiency than that at client. Similarly, Figure \ref{figs_densenet_communicate_server} shows that the communication efficiency at server can be improved by reducing $45\%\sim60\%$ model size more than \textit{STC}. Therefore, those numerical results validate that \textit{CER} makes a good tradeoff between accuracy and communication efficiency.

\subsubsection{Segmentation of Brain Tumor}

\begin{figure}
\setlength{\abovecaptionskip}{0pt}
\setlength{\belowcaptionskip}{0pt}
\centering 
\includegraphics[width=0.99\columnwidth]{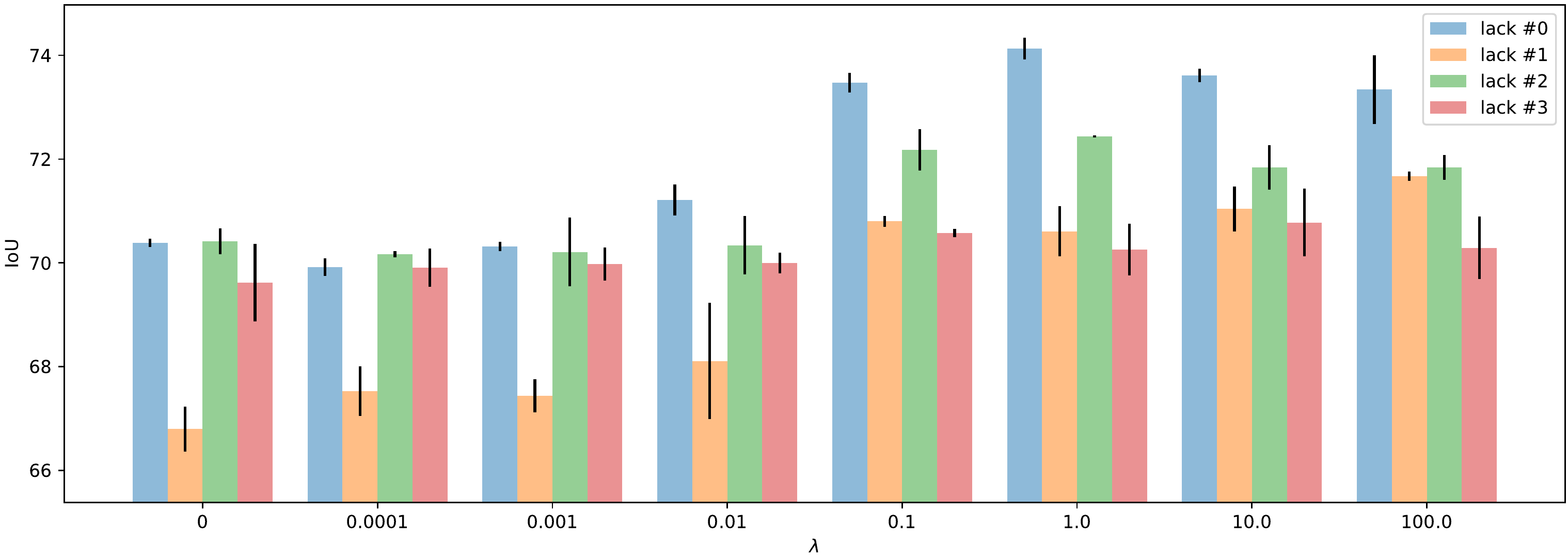}
\caption{Illustrative result of test accuracy w.r.t. $\lambda$.}
\label{figs_lambda_braTS2017}
\end{figure}

\begin{figure}
\setlength{\abovecaptionskip}{0pt}
\setlength{\belowcaptionskip}{0pt}
\centering 
\subfigure[Model size reduction, \textit{CER} v.s. \textit{STC}]{\includegraphics[width=0.49\columnwidth]{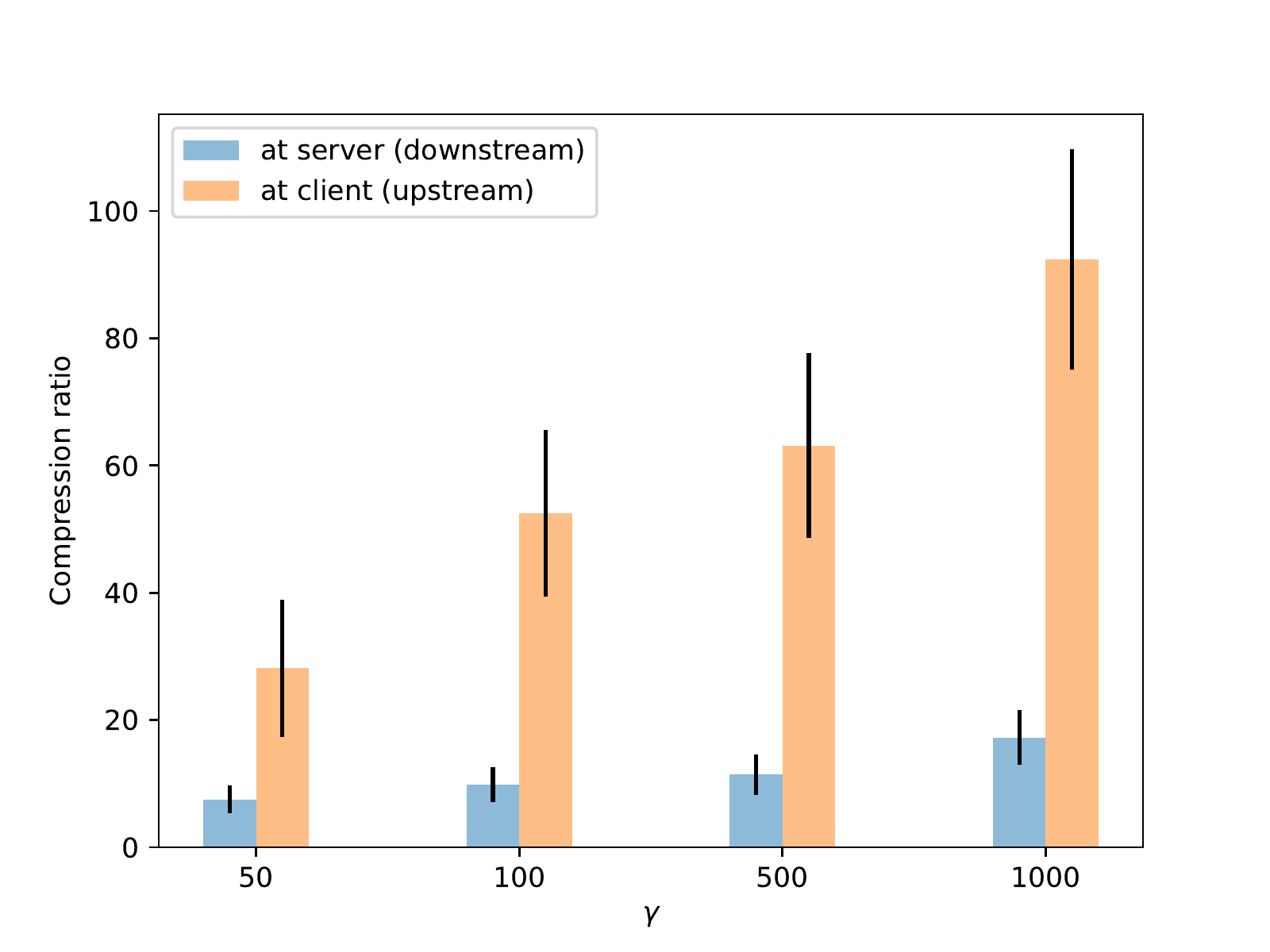}\label{figs_gamma_braTS2017}}
\subfigure[Test IoU with \textit{CER}, \textit{CER} v.s. \textit{STC}]{\includegraphics[width=0.49\columnwidth]{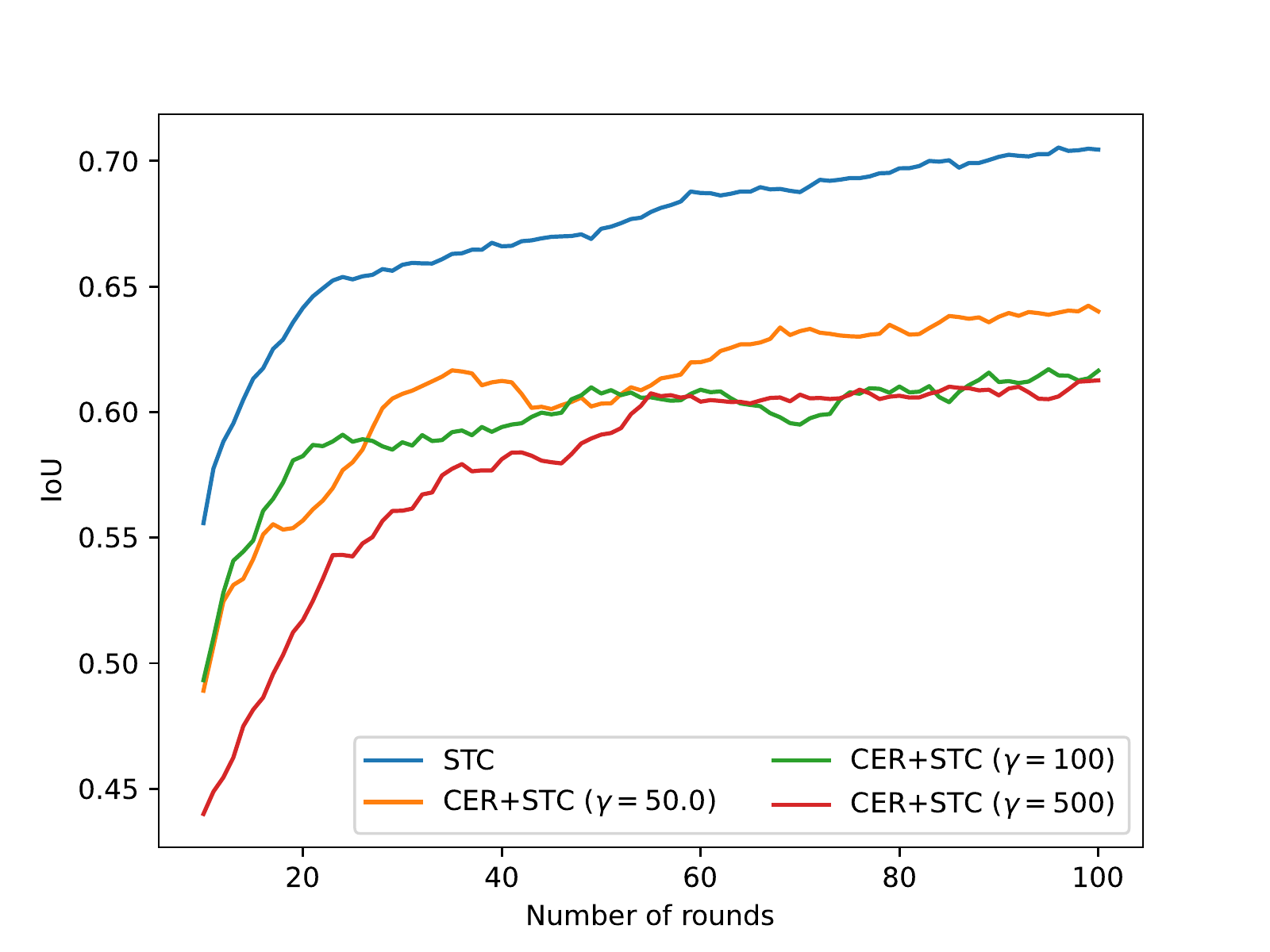}\label{figs_unet_communicate_iou_line}}
\subfigure[Model size reduction at client, \textit{CER} v.s. \textit{STC}]{\includegraphics[width=0.49\columnwidth]{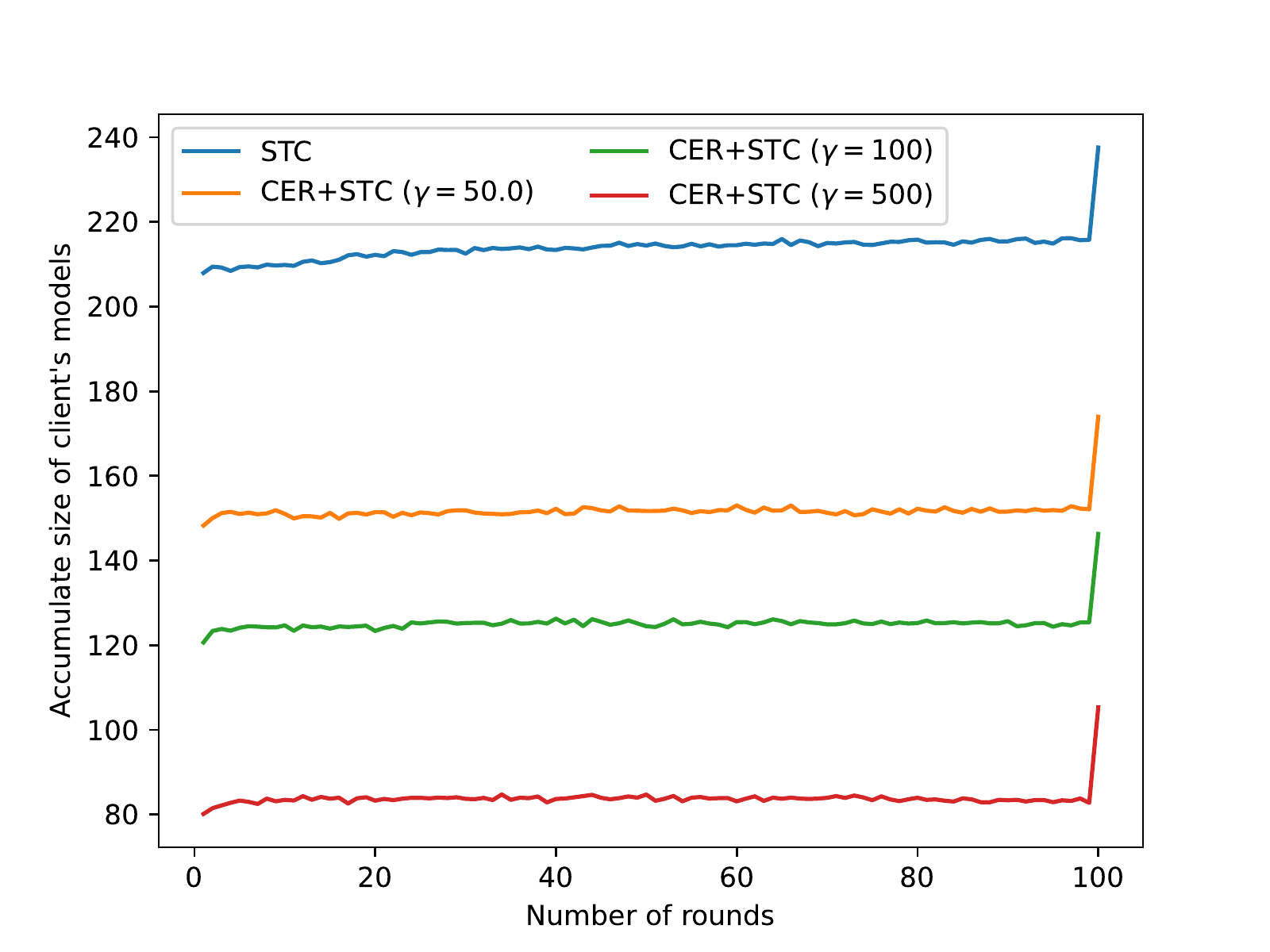}\label{figs_unet_communicate_client_line}}
\subfigure[Statistics of model size reduction at client, \textit{CER} v.s. \textit{STC}]{\includegraphics[width=0.49\columnwidth]{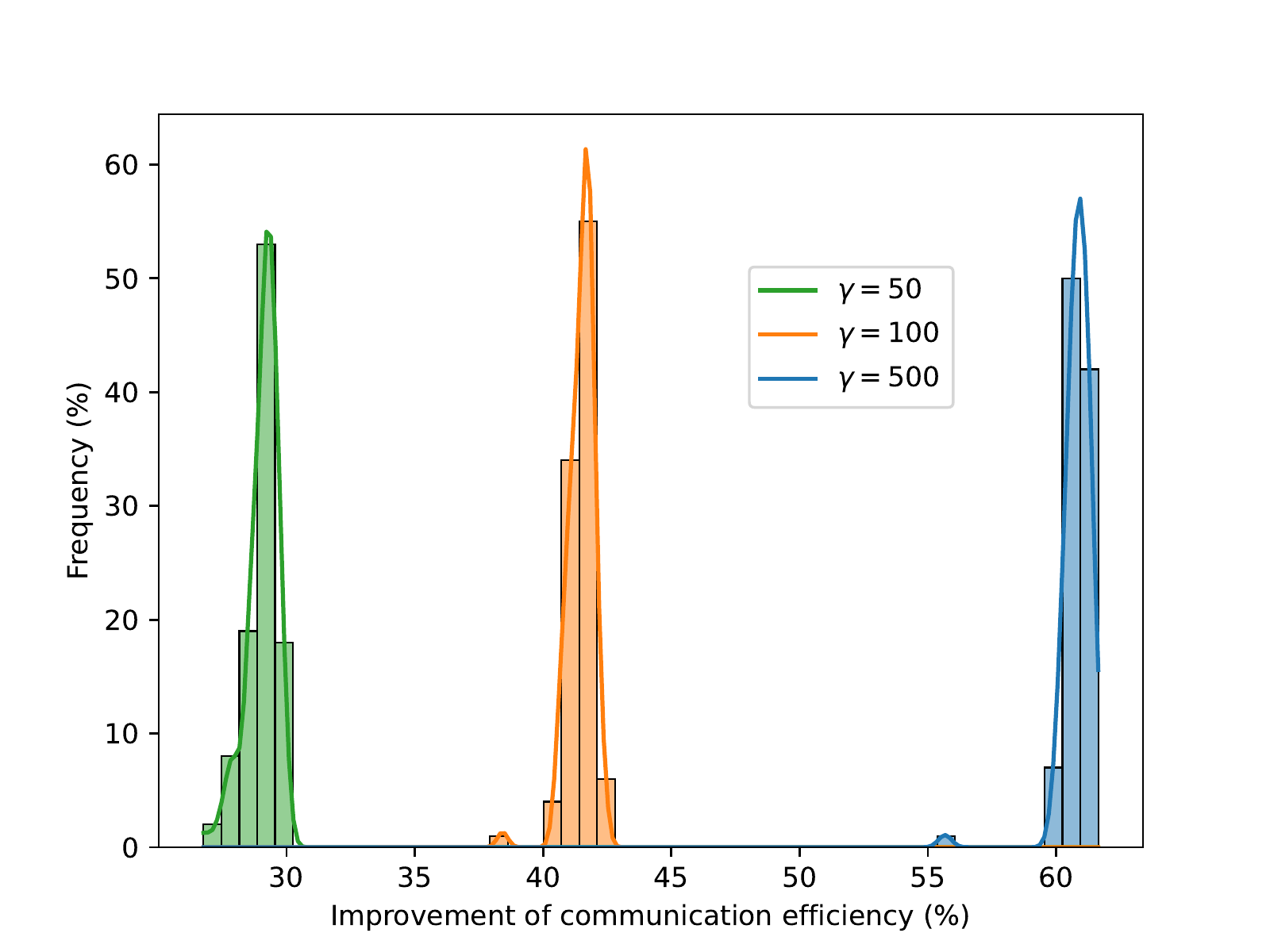}\label{figs_unet_communicate_client}}
\subfigure[Model size reduction at server, \textit{CER} v.s. \textit{STC}]{\includegraphics[width=0.49\columnwidth]{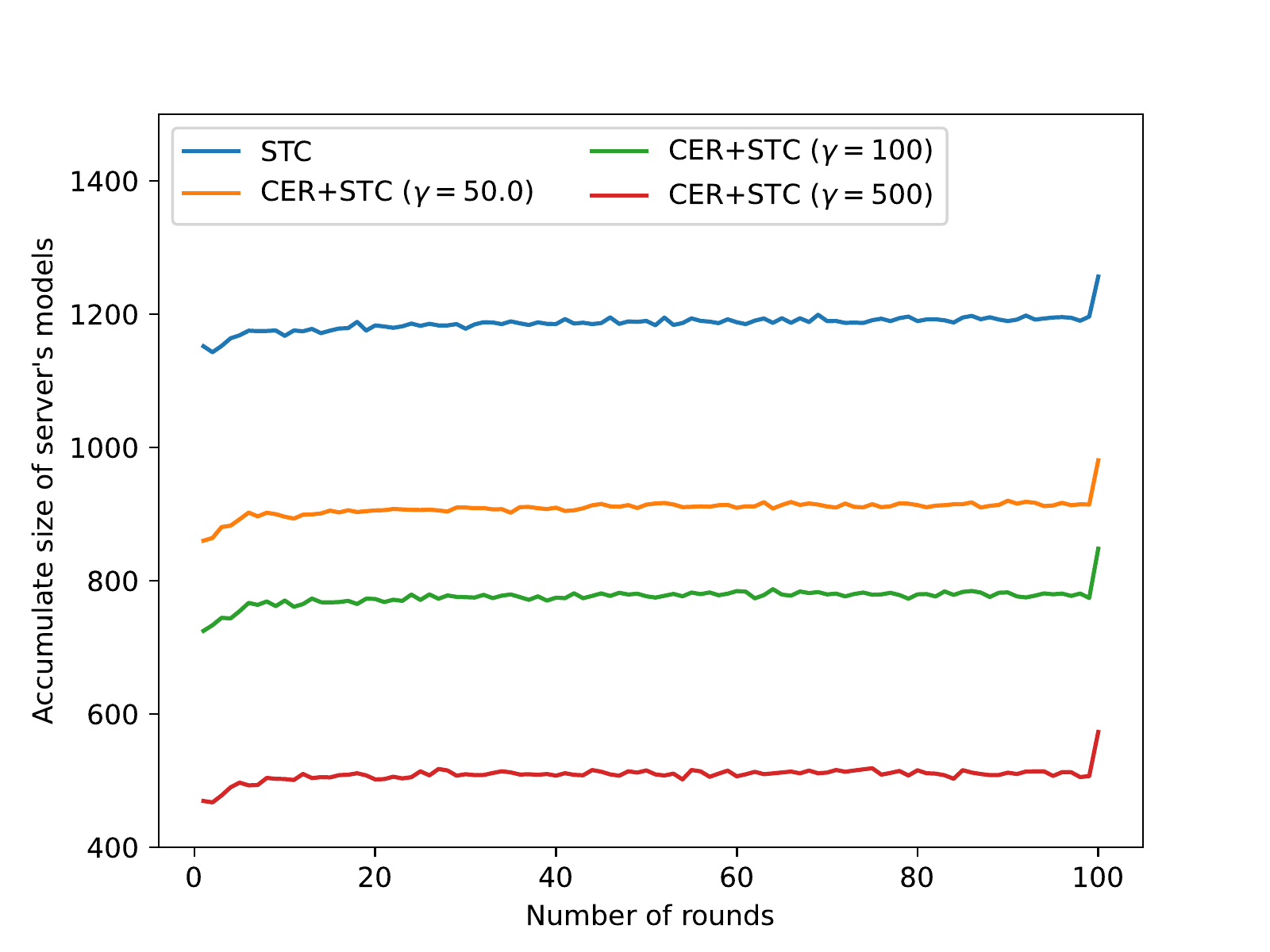}\label{figs_unet_communicate_server_line}}
\subfigure[Statistics of model size reduction at server, \textit{CER} v.s. \textit{STC}]{\includegraphics[width=0.49\columnwidth]{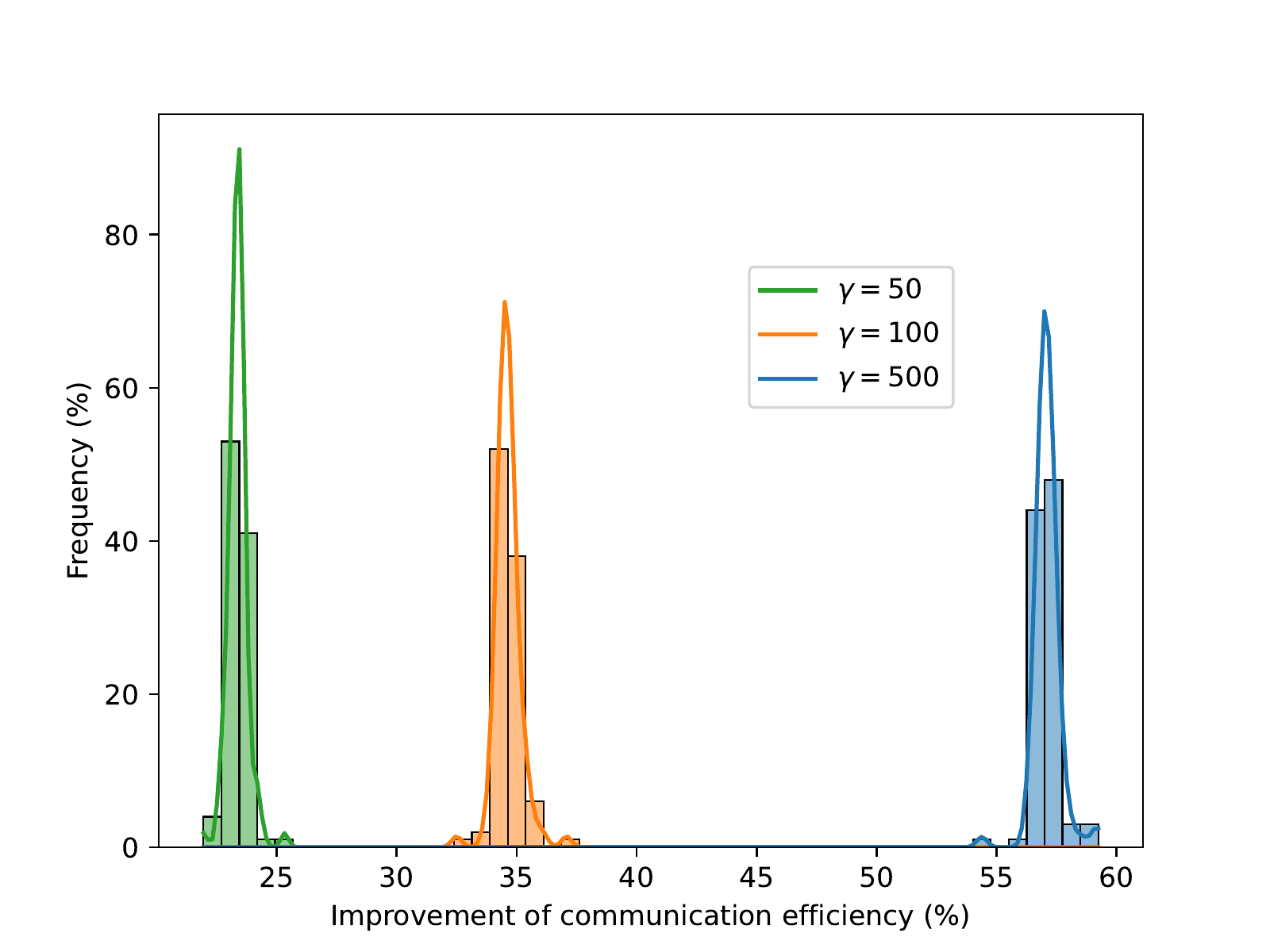}\label{figs_unet_communicate_server}}
\caption{\textit{pFedNet} effectively reduces model size with \textit{CER}.}
\label{figure_brats_cer}
\end{figure}

First, we evaluate \textit{IoU} and \textit{label IoU} for all methods on the dataset \textit{BraTS2017} by varying the number of missing channels of MRI images. Here, \textit{IoU} is an evaluation metric used to measure the accuracy of an object detector on a particular dataset. Similarly, \textit{label IoU} measures the accuracy of the foreground of the target object. As shown in Tables \ref{table_brats_iou} and \ref{table_brats_label_iou}, the proposed method, namely \textit{pFedNet}, achieves better performance than most of existing methods. Although \textit{pFedNet} does not achieve the best performance when missing either $\#1$ or $\#2$ channel, the gap is not significant (less than $0.5\%$), and still outperforms other methods. Additionally, Figure \ref{figs_lambda_braTS2017} shows that  \textit{IoU} increases significantly with the increase of $\lambda$, but may decrease with a too large value of $\lambda$. We find that $\lambda=1.0$ seems to be a good choice for different settings of data federation.

%Moreover, we observe that \textit{IoU} is much more sensitive to $\lambda$ for the .  

Second, we evaluate the communication efficiency of the proposed method, that is \textit{CER}. As illustrated in Figure \ref{figs_gamma_braTS2017}, \textit{STC} without \textit{CER}, that is the case of $\gamma=0$ enjoys more than $7\times$ compression ratio of model size at server, and $28\times$ compression ration at client. Although it is effective to compress model, the proposed method \textit{CER} successfully achieves more than $9\times$ and $52\times$ compression ratio for the server and client when choosing $\gamma=50$, respectively. Its advantage becomes significant with the increase of $\gamma$, and achieves more than $17\times$ and $92\times$ compression ratio for the server and client. The reason is that \textit{CER} encourages local update of model to own clustering structure, which is more suitable to conduct compression by using existing methods. According to Figure \ref{figs_unet_communicate_iou_line}, we observe that \textit{CER} with a large 
$\gamma$ is indeed harmful to the model training, which means the model may need more time to achieve the optimum by using large $\gamma$. It validates that \textit{CER} makes tradeoff between communication efficiency and model performance. We suggest to adopt the dynamic strategy to choosing $\gamma$ during model learning to obtain more gains of communication efficiency without much sacrifice of model performance.   Figure \ref{figs_unet_communicate_client_line} shows that \textit{CER} can be used  together with \textit{STC}, and achieves much more significant compression. The superiority becomes significant with increase of $\gamma$. According to Figure \ref{figs_unet_communicate_client}, we observe that \textit{CER} gains more than $28\%\sim60\%$ improvement of communication efficiency at client. Similarly, we find that more than $23\%\sim57\%$ improvement of communication efficiency at server according to Figures \ref{figs_unet_communicate_server_line} and \ref{figs_unet_communicate_server}. It validates that \textit{CER} can successfully find a good tradeoff between model performance and communication efficiency once more.

\begin{figure*}
\setlength{\abovecaptionskip}{0pt}
\setlength{\belowcaptionskip}{0pt}
\centering 
\includegraphics[width=1.99\columnwidth]{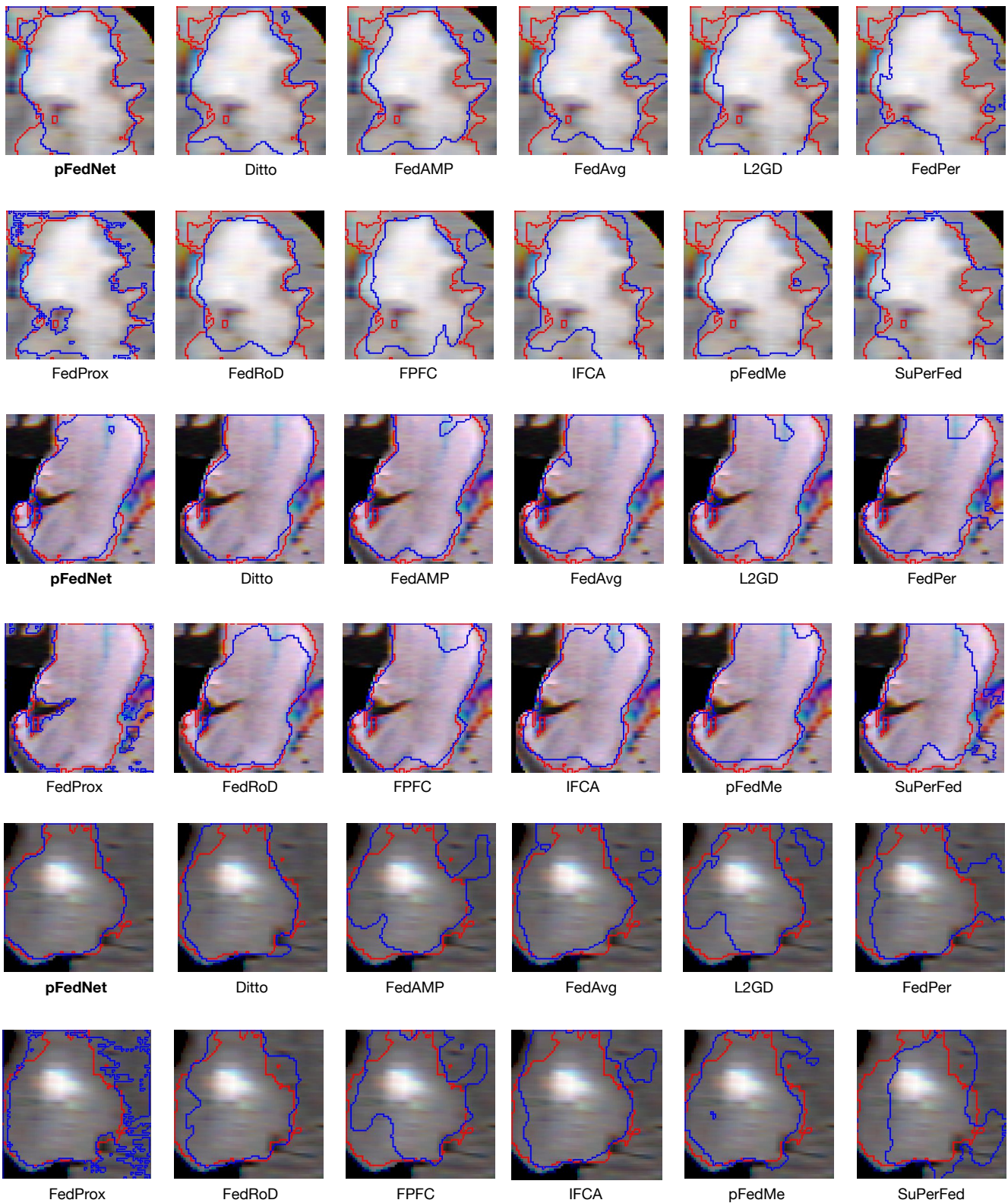}
\caption{Illustrative examples of segmentation of brain tumor by using \textit{pFedNet} and other existing methods. \textcolor{red}{Red} line represents the true region, and \textcolor{blue}{Blue} line represents the segmentation region.}
\label{figs_segmenation_demo}
\end{figure*}

Finally, Figure \ref{figs_segmenation_demo} illustrates the true region of target object (the red line), and some examples of the segmentation region (the blue line) yielded by \textit{pFedNet} and other methods for the setting of `lack \#0'. As we can see, \textit{pFedNet} captures details of interested region more accurately than others. 

\subsection{Numerical Results on Private Datasets}
\subsubsection{Prediction of Clinical Risk}
\begin{figure}
\setlength{\abovecaptionskip}{0pt}
\setlength{\belowcaptionskip}{0pt}
\centering 
\subfigure[Accuracy w.r.t. $\lambda$ for \textit{CHD}]{\includegraphics[width=0.99\columnwidth]{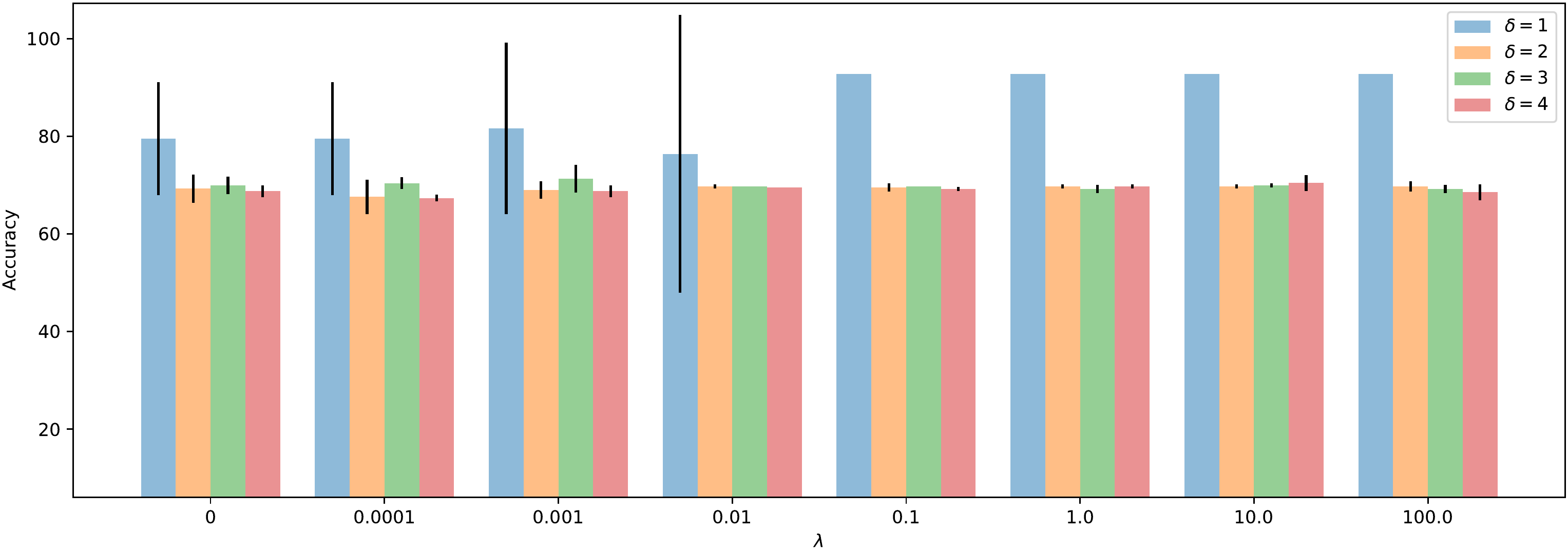}\label{figs_lambda_Lngxbhbcdexzlhzcxfxyc}}
\subfigure[Accuracy w.r.t. $\lambda$ for \textit{Diabetes}]{\includegraphics[width=0.99\columnwidth]{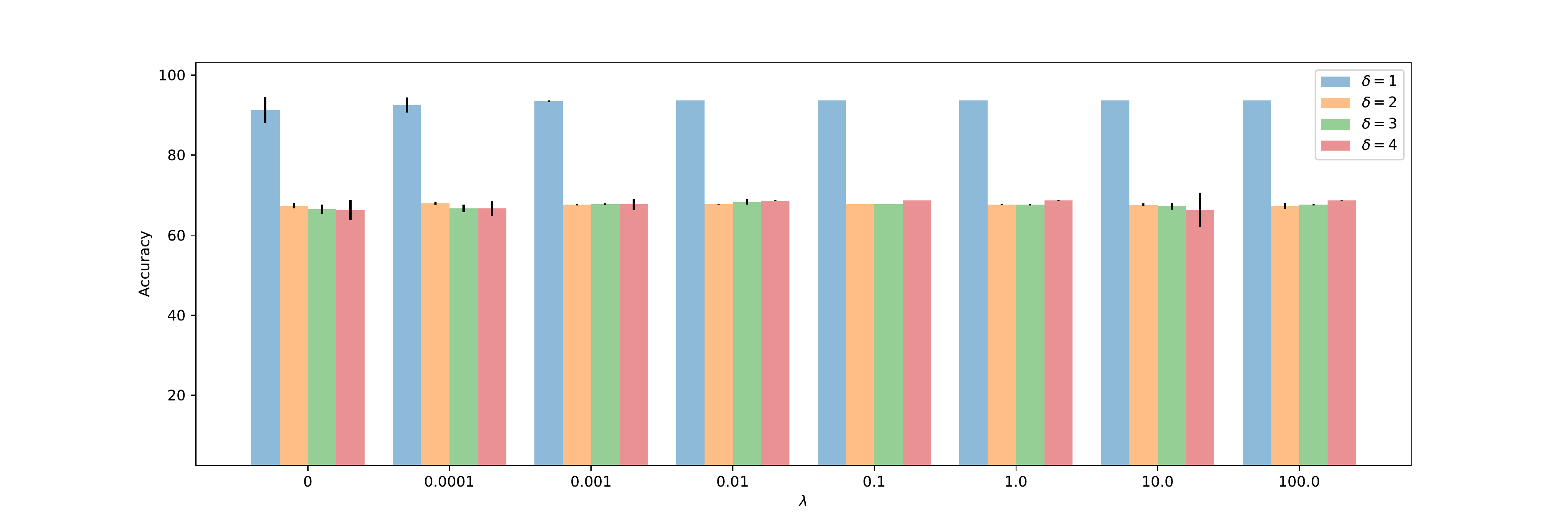}\label{figs_lambda_IITnbswmbbfxyc}}
\subfigure[Accuracy w.r.t. $\lambda$ for \textit{Covid19}]{\includegraphics[width=0.99\columnwidth]{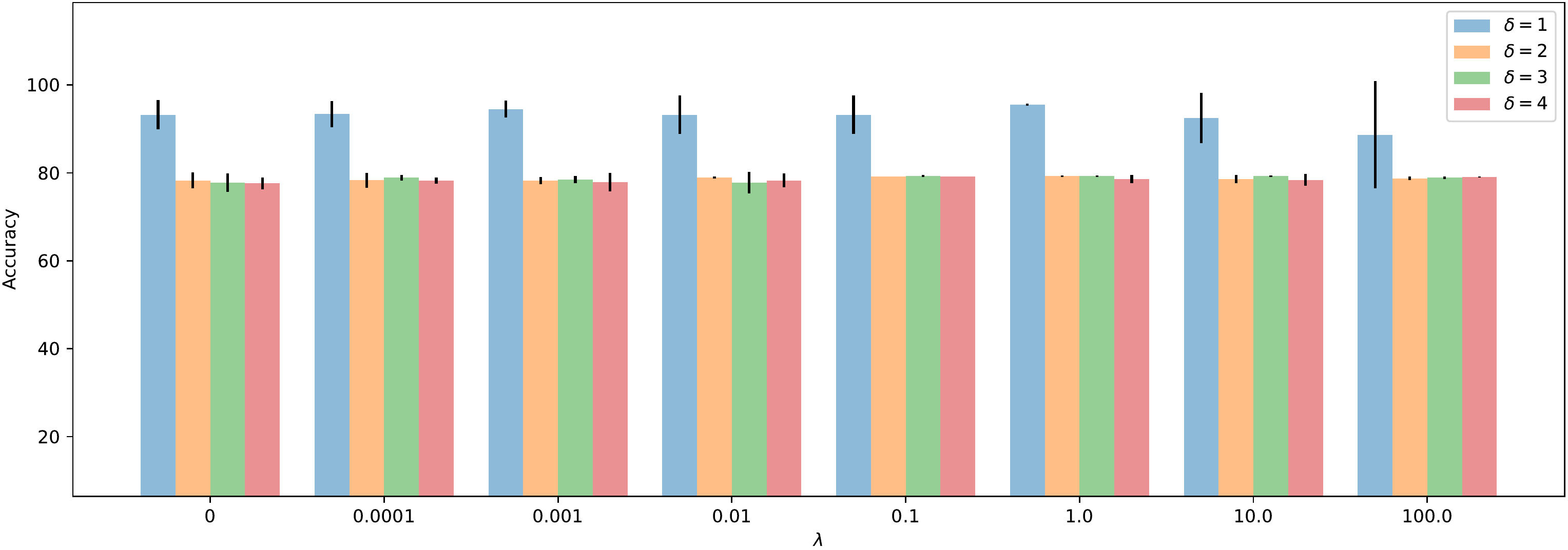}\label{figs_lambda_covid19}}
\caption{Illustrative results of accuracy w.r.t $\lambda$.}
\label{figure_acc_lambda_chd_iit_covid19}
\end{figure}

We evaluate \textit{pFedNet} by using LR model on three structural medical datasets. As illustrated in Tables \ref{table_acc_chd}-\ref{table_acc_covid19}, \textit{pFedNet} outperforms other methods on the test accuracy in most settings of $\delta$. It shows that the proposed model \textit{pFedNet} works well for tabular data, and validates the  superiority on the model performance again. Additionally, we evaluate the test accuracy by varying $\lambda$. According to Figure \ref{figure_acc_lambda_chd_iit_covid19}, we observe that \textit{pFedNet} achieves significantly high accuracy in the balanced setting of the data federation, that is $\delta=1$. Specifically, the accuracy seems to increase slightly with a large $\lambda$, but may decrease when $\lambda$ is too large. The reason is that $\lambda$ controls the the tradeoff between personalized model and general model, and it should not be either too large or small. Specifically, $\lambda=0.1$ is recommended for the task since it gains best performance in most settings of $\delta$. Since every LR model in the task only has $58\sim77$ features, which leads to tiny workload of data transmission, the communication efficiency is not evaluated here. 

\section{Conclusion}
\label{section_conclusion}
We propose a new formulation of personalized federated learning, which has good adaption to heterogenous medical data, and achieves better performance than existing methods. To improve the communication efficiency, we further develop a communicate efficient regularizer , which can decrease workload of communication effectively. Additionally, we propose a new optimization framework to update personalized models, which reduces computation cost significantly. Extensive empirical studies have been conducted to verify the effectiveness of the proposed method. In the future, we explore and analyze the dynamics of medical data, and try to develop the adaptive version of the proposed model to capture such dynamics.

\section*{Acknowledgment}
This work was supported by the funding Grants No. 22-TQ23-14-ZD-01-001, and No. 145BQ090003000X03. Zongren Li and Qin Zhong provide help on collecting medical datasets and related papers. Hebin Che provide help on processing of medical data. Mingming Jiang gives advices about English writing. Thanks a lot for their kind heart and help.

\balance

\bibliographystyle{IEEEtran}  
\bibliography{reference}

\begin{IEEEbiography}[{\includegraphics[width=1in,height=1.25in,clip,keepaspectratio]{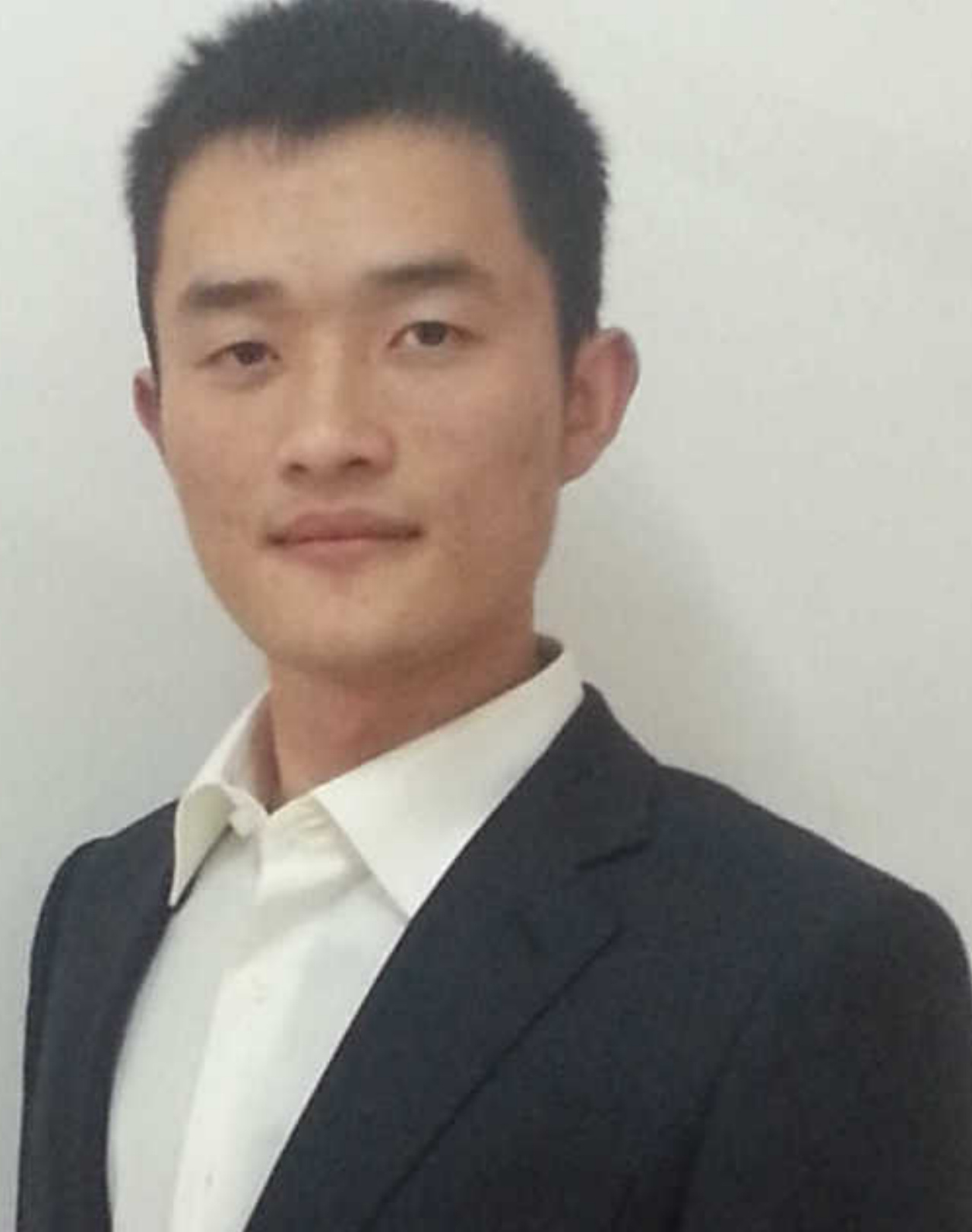}}]{Yawei Zhao} is now working at Medical Big Data Research Center of Chinese PLA General Hospital \& National Engineering Research Center for the Application Technology of Medical Big Data, Beijing, 100853, China. He received his Ph.D, B.E., and M.S. degree in Computer Science from the National University of Defense Technology, China, in 2013, 2015, and 2020, respectively. His research interests include federated learning, and medical artificial intelligence. Dr. Zhao has published 10+ peer-reviewed papers in highly regarded journals and conferences such as IEEE T-KDE, IEEE T-NNLS, AAAI, etc. 
\end{IEEEbiography}

\begin{IEEEbiography}[{\includegraphics[width=1in,height=1.25in,clip,keepaspectratio]{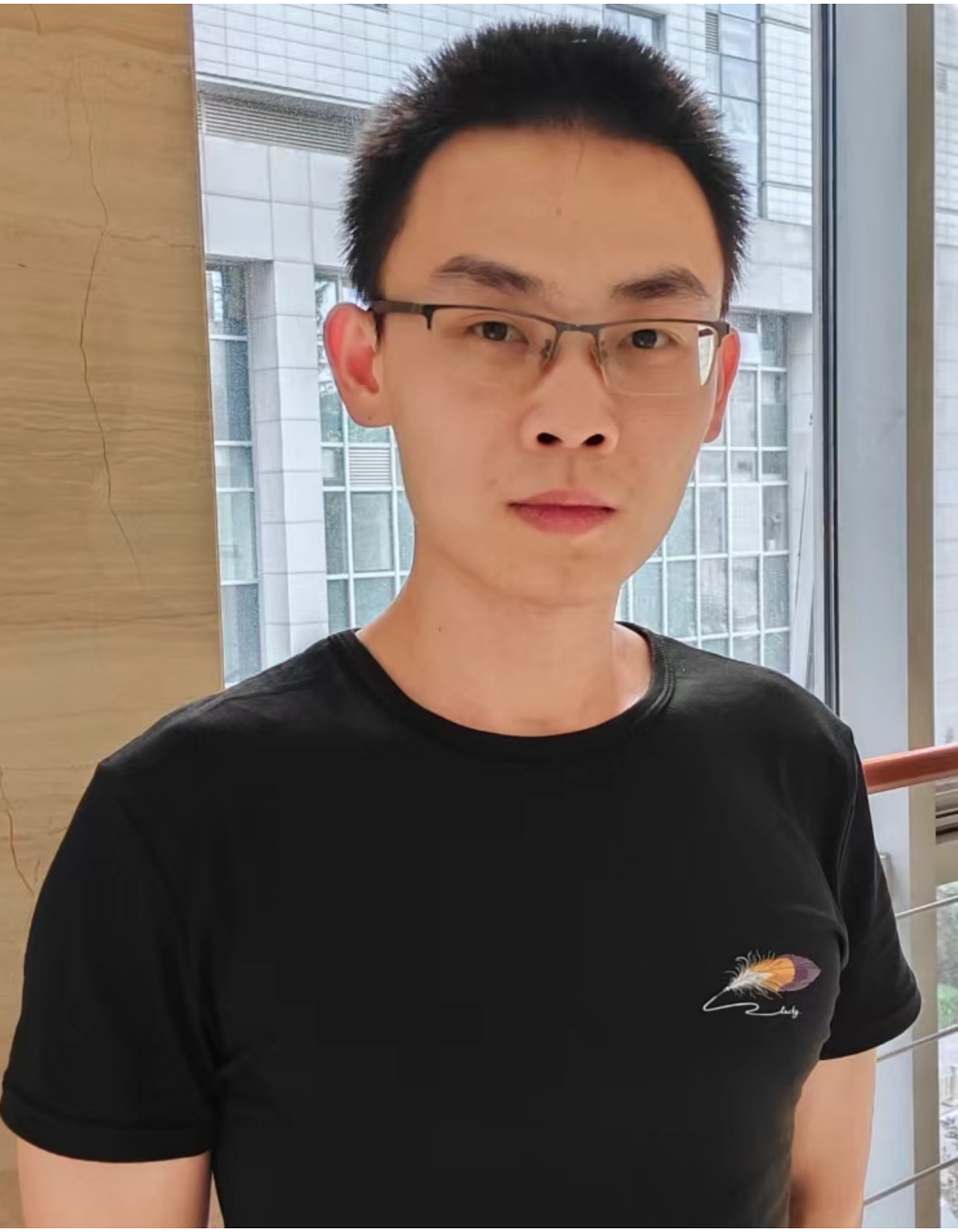}}]{Qinghe Liu} is now pursuing his M.S. degree at Chinese PLA General Hospital \& National Engineering Research Center for the Application Technology of Medical Big Data, Beijing, 100853, China. His research interests include federated learning and medical image analysis.
\end{IEEEbiography}

\begin{IEEEbiography}
[{\includegraphics[width=1in,height=1.25in,clip,keepaspectratio]{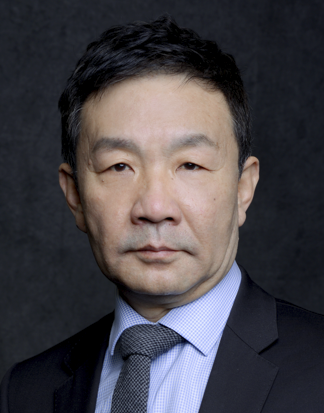}}]
{Kunlun He} received his M.D. degree from The 3rd Military Medical University, Chongqing, China in 1988, and PhD degree in Cardiology from Chinese PLA Medical school, Beijing, China in 1999. He worked as a postdoctoral research fellow at Division of circulatory physiology of Columbia University from 1999 to 2003. He is director and professor of Medical Big Data Research Center, Chinese PLA General Hospital, Beijing, China. His research interests include big data and artificial intelligence of cardiovascular disease.
\end{IEEEbiography}

\begin{IEEEbiography}
[{\includegraphics[width=1in,height=1.25in,clip,keepaspectratio]{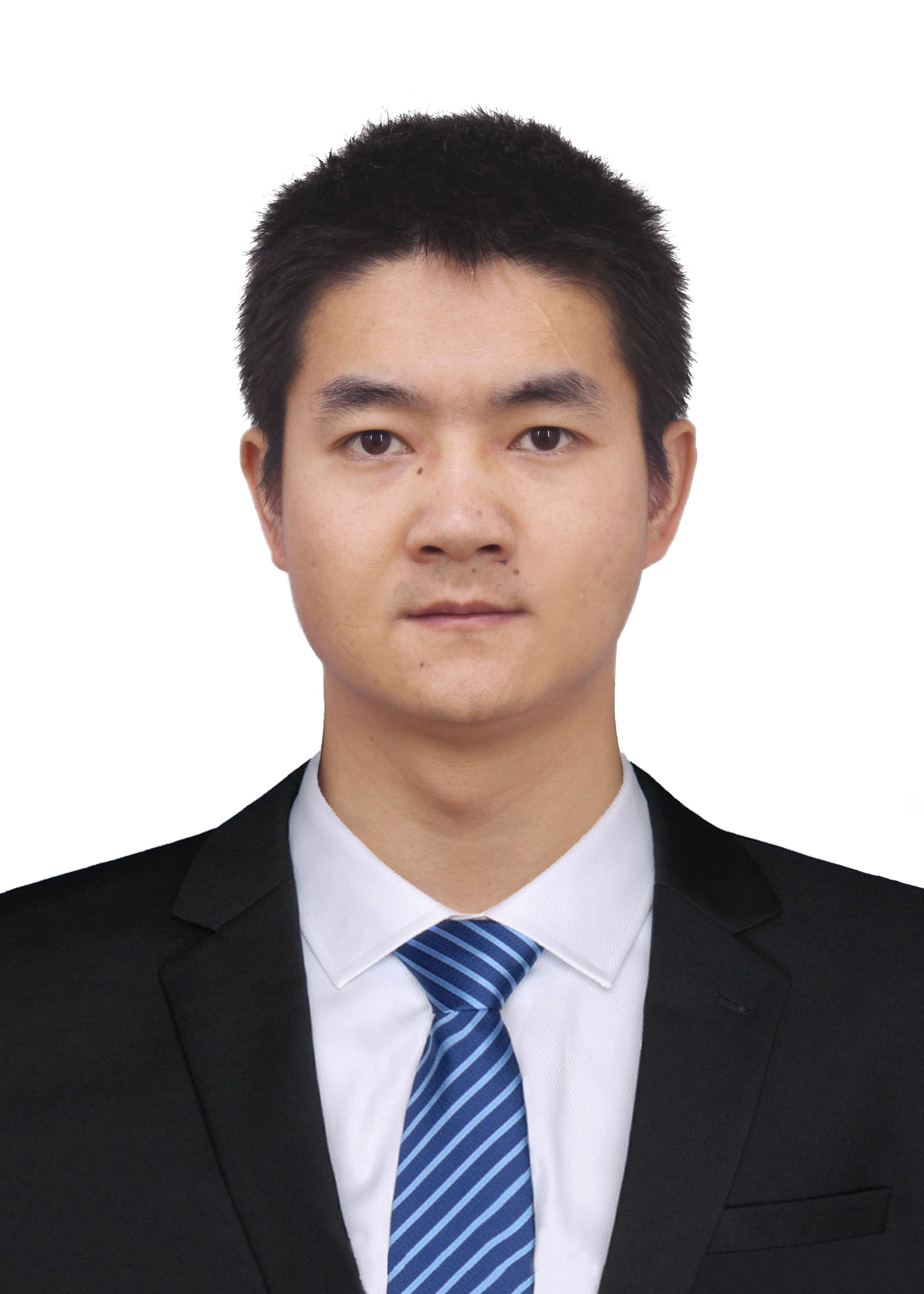}}]
{Xinwang Liu} received his Ph.D. degree from the National University of Defense Technology (NUDT), China. He is now a full professor at the College of Computer, NUDT. His current research interests include kernel learning and unsupervised feature learning. Dr. Liu has published 100+ peer-reviewed papers, including those in highly regarded journals and conferences such as IEEE T-PAMI, IEEE T-KDE, IEEE T-IP, IEEE T-NNLS, IEEE T-MM, IEEE T-IFS, ICML, NeurIPS, ICCV, CVPR, AAAI, IJCAI, etc. He is a senior member of IEEE. He serves as the associated editor of the Information Fusion Journal, IEEE T-NNLS Journal, and IEEE T-CYB Journal. More information can be found at \url{https://xinwangliu.github.io}.
\end{IEEEbiography}

\end{document}